\documentclass[10pt,twocolumn,letterpaper]{article}
\usepackage{cvpr}
\def\confName{CVPR}
\def\confYear{2023}
\usepackage{graphicx}
\usepackage{amsmath,amssymb,amsfonts,mathabx,mathbbol}
\usepackage{algorithmic}
\usepackage[linesnumbered,ruled,vlined]{algorithm2e}
\usepackage{acronym}
\usepackage{enumitem}
\usepackage[pagebackref,breaklinks,colorlinks,citecolor=gray]{hyperref}
\usepackage{balance}
\usepackage{xspace}
\usepackage{setspace}
\usepackage[skip=3pt,font=small]{subcaption}
\usepackage[skip=3pt,font=small]{caption}
\usepackage[dvipsnames]{xcolor}
\usepackage[capitalise]{cleveref}
\usepackage{booktabs,tabularx,colortbl,multirow,array,makecell}
\usepackage{overpic,wrapfig}
\usepackage[misc]{ifsym}
\usepackage{nicefrac}
\usepackage{stfloats}
\usepackage[export]{adjustbox} %

\makeatletter
\DeclareRobustCommand\onedot{\futurelet\@let@token\@onedot}
\def\@onedot{\ifx\@let@token.\else.\null\fi\xspace}
\def\eg{\emph{e.g}\onedot} 

\def\ie{\emph{i.e}\onedot}

\def\etal{\emph{et al}\onedot}
\makeatother

\DeclareMathOperator*{\argmax}{arg\,max}

\newcommand{\ba}{\mathbf{a}}
\newcommand{\bc}{\mathbf{c}}
\newcommand{\bs}{\mathbf{s}}

\newcommand{\bI}{\mathbf{I}}
\newcommand{\bg}{\mathbf{g}}
\newcommand{\btau}{\boldsymbol\tau}
\newcommand{\bmu}{\boldsymbol\mu}
\newcommand{\bepsilon}{\boldsymbol\epsilon}
\newcommand{\bSigma}{\boldsymbol\Sigma}
\newcommand{\bbR}{\mathbb{R}}
\newcommand{\bbE}{\mathbb{E}}
\newcommand{\cN}{\mathcal{N}}
\newcommand{\cS}{\mathcal{S}}
\newcommand{\cG}{\mathcal{G}}
\newcommand{\cJ}{\mathcal{J}}
\newcommand{\cO}{\mathcal{O}}
\newcommand{\obj}{\varphi}

\newcommand{\Eb}[2]{\bbE_{#1}\!\left[\|#2\|^2\right]}

\makeatletter
\renewcommand{\paragraph}{%
  \@startsection{paragraph}{4}%
  {\z@}{0ex \@plus 0ex \@minus 0ex}{-1em}%
  {\hskip\parindent\normalfont\normalsize\bfseries}%
}
\makeatother

\crefname{algorithm}{Alg.}{Algs.}
\Crefname{algocf}{Alg.}{Algs.}
\crefname{section}{Sec.}{Secs.}
\Crefname{section}{Section}{Sections}
\crefname{table}{Tab.}{Tabs.}
\Crefname{table}{Table}{Tables}
\crefname{figure}{Fig.}{Fig.}
\Crefname{figure}{Figure}{Figure}

\graphicspath{{figures/}}

\newcommand{\model}{\text{SceneDiffuser}\xspace}

\acrodef{cvae}[cVAE]{conditional Variational Autoencoder}
\acrodef{cgan}[cGAN]{conditional generative adversarial networks}
\acrodef{tamp}[TAMP]{task and motion planning}
\acrodef{rl}[RL]{reinforcement learning}
\acrodef{mdp}[MDP]{Markov decision process}
\acrodef{bc}[BC]{Behavior Cloning}

\SetCommentSty{mycommfont}

\definecolor{pose_gen}{RGB}{72,179,179}
\definecolor{motion_gen}{RGB}{159,71,179}
\definecolor{grasp_gen}{RGB}{143,179,0}
\definecolor{navigation}{RGB}{59,79,179}
\definecolor{goal}{RGB}{179,0,0}
\definecolor{motion_plan}{RGB}{0,179,83}

\newcommand\blfootnote[1]{%
  \begingroup
  \renewcommand\thefootnote{}\footnote{#1}%
  \addtocounter{footnote}{-1}%
  \endgroup
}
\begin{document}

\title{Diffusion-based Generation, Optimization, and Planning in 3D Scenes}
\author{%
    Siyuan Huang$^{1*\,\textrm{\Letter}}$, Zan Wang$^{1,2*}$, Puhao Li$^{1,3}$, Baoxiong Jia$^{1}$\\
    Tengyu Liu$^1$, Yixin Zhu$^{4}$, Wei Liang$^{2\,\textrm{\Letter}}$, Song-Chun Zhu$^{1,3,4}$
    \vspace{6pt}\\
    \small $^1$ National Key Laboratory of General Artificial Intelligence, BIGAI\\
    \small $^2$ School of Computer Science \& Technology, Beijing Institute of Technology\\
    \small $^3$ Dept. of Automation, Tsinghua University\quad{}
    \small $^4$ Institute for AI, Peking University\quad{}
    \vspace{6pt}\\
    \href{https://scenediffuser.github.io}{https://scenediffuser.github.io}
}

\twocolumn[{%
\renewcommand\twocolumn[1][]{#1}%
\maketitle
\begin{center}
    \centering
    \captionsetup{type=figure}
        \includegraphics[width=\linewidth]{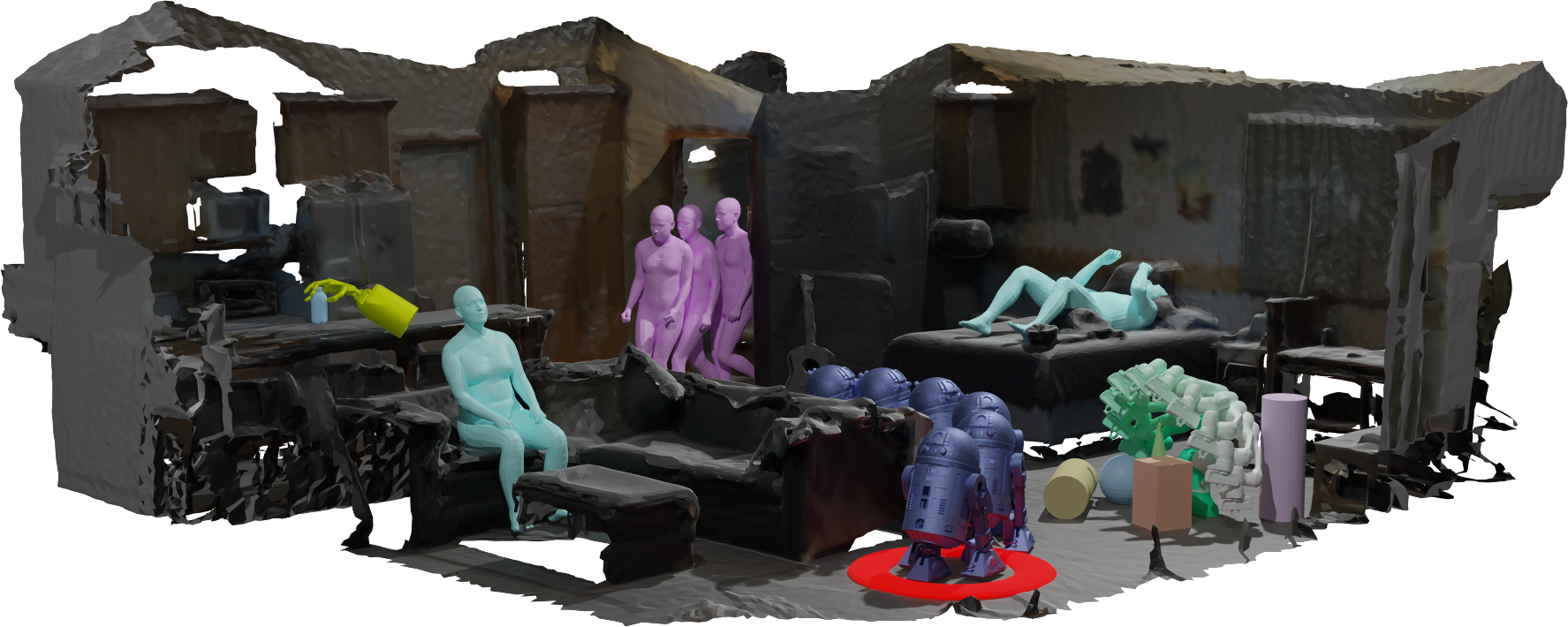}
        \captionof{figure}{\textbf{Illustration of the \model}, applicable to various scene-conditioned 3D tasks: (a) \textcolor{pose_gen}{human pose generation}, (b) \textcolor{motion_gen}{human motion generation}, (c) \textcolor{grasp_gen}{dexterous grasp generation}, (d) \textcolor{navigation}{path planning for 3D navigation} with \textcolor{goal}{goals}, and (e) \textcolor{motion_plan}{motion planning for robot arms}.%
        }
    \label{fig:illustration}
\end{center}
}]
\blfootnote{$^*$ These authors contributed equally to this work.}
\blfootnote{$^\textrm{\Letter}$ Corresponding authors: Siyuan Huang (\texttt{syhuang@bigai.ai}) and Wei Liang (\texttt{liangwei@bit.edu.cn}).}

\begin{abstract}
We introduce \model, a conditional generative model for 3D scene understanding. \model provides a unified model for solving scene-conditioned generation, optimization, and planning. In contrast to prior work, \model is intrinsically scene-aware, physics-based, and goal-oriented. With an iterative sampling strategy, \model jointly formulates the scene-aware generation, physics-based optimization, and goal-oriented planning via a diffusion-based denoising process in a fully differentiable fashion. Such a design alleviates the discrepancies among different modules and the posterior collapse of previous scene-conditioned generative models. We evaluate \model on various 3D scene understanding tasks, including human pose and motion generation, dexterous grasp generation, path planning for 3D navigation, and motion planning for robot arms. The results show significant improvements compared with previous models, demonstrating the tremendous potential of \model for the broad community of 3D scene understanding.
\end{abstract}

\section{Introduction}

The ability to generate, optimize, and plan in 3D scenes is a long-standing goal for multiple research domains across computer vision, graphics, and robotics. Various tasks have been devised to achieve these goals, fostering downstream applications in motion generation~\cite{li2019putting,zhangyan2020generating,wang2021synthesizing,wang2022humanise}, motion planning~\cite{peng2016terrain,starke2019neural,peng2021amp,won2022physics}, grasp generation~\cite{jiang2021hand,liu2021synthesizing,li2022gendexgrasp}, navigation~\cite{zhu2017target,anderson2018vision}, embodied perception and manipulation~\cite{mu2021maniskill,szot2021habitat,li2021igibson}, and autonomous driving~\cite{sadat2019jointly,casas2021mp3}.

Despite rich applications and great successes, existing models designed for these tasks exhibit two fundamental limitations for real-world 3D scene understanding.

First, most prior work~\cite{li2019putting,wang2022humanise,rempe2021humor,wang2021synthesizing,hassan2021stochastic,wang2021scene,starke2019neural,cui2021lookout,wang2022towards} leverages the \ac{cvae} for the conditional generation in 3D scenes. \ac{cvae} model utilizes an encoder-decoder structure to learn the posterior distribution and relies on the learned latent variables to sample. Although \ac{cvae} is easy to train and sample due to its simple architecture and one-step sampling procedure, it suffers from the \textbf{posterior collapse problem}~\cite{zhao2017infovae,tolstikhin2018wasserstein,kim2018semi,he2018lagging,fu2019cyclical,yuan2020dlow,wang2021synthesizing}; the learned latent variable is ignored by a strong decoder, leading to limited generation diversity from these collapsed modes. Such collapse is further magnified in 3D tasks with stronger 3D decoders and more complex and noisy input conditions, \eg, the natural 3D scans~\cite{dai2017scannet}.

Second, despite the close relations among generation, optimization, and planning in 3D scenes, there \textbf{lacks a unified framework} that could address existing discrepancies among these models. Previous work~\cite{hassan2019resolving,liu2021synthesizing,wang2021synthesizing} applies off-the-shelf physics-based post-optimization methods over outputs of generative models and often produces inconsistent and implausible generations, especially when transferring to novel scenes. Similarly, planners are usually standalone modules over results of generative model~\cite{hassan2021stochastic,cui2021lookout} for trajectory planning or learned separately with the \ac{rl}~\cite{zhu2017target}, leading to gaps between planning and other modules (\eg, generation) during inference, especially in novel scenes where explorations are limited.

To tackle these limitations, we introduce \model, a conditional generative model based on the diffusion process. \model eliminates the discrepancies and provides a single home for scene-conditioned generation, optimization, and planning. Specifically, with a denoising process, it learns a diffusion model for scene-conditioned generation while training. During inference, \model jointly solves the scene-aware generation, physics-based optimization, and goal-oriented planning through a unified iterative guided-sampling framework. Such a design equips \model with the following superiority:

\paragraph{Generation}

Building upon the diffusion model, \model significantly alleviates the posterior collapse problem of scene-conditioned generative models. Since the forward diffusion process can be treated as data augmentation in 3D scenes, it helps traverse sufficient scene-conditioned distribution modes.

\paragraph{Optimization}

\model integrates the physics-based objective into each step of the sampling process as conditional guidance, enabling the differentiable physics-based optimization during both the learning and sampling process. This design facilitates the physically-plausible generation, which is critical for tasks in 3D scenes.

\paragraph{Planning}

Based on the scene-conditioned trajectory-level generator, \model possesses a global trajectory planner with physics and goal awareness, making the learned planner generalize better to long-horizon trajectories and novel 3D scenes.

As illustrated in \cref{fig:illustration}, we evaluate \model on diverse 3D scene understanding tasks. The results on human pose, motion, and dexterous grasp generation significantly improve, demonstrating plausible and diverse generations with 3D scene and object conditions. The results on path planning for 3D navigation and motion planning for robot arms reveal the generalizable and long-horizon planning capability of \model.

\section{Related Work}

\paragraph{Conditional Generation in 3D Scenes}

Generating diverse contents and rich interactions in 3D scenes is essential for understanding the 3D scene affordances. Recently, we have witnessed several applications on conditional scene generation~\cite{jiang2018configurable,qi2018human,zhang2019survey,wang2018deep}, human pose~\cite{li2019putting,zhangyan2020generating,zhangsiwei2020generating,hassan2021populating,zhao2022compositional} and motion generation~\cite{li2019putting,wang2022humanise,rempe2021humor,wang2021synthesizing,hassan2021stochastic,wang2021scene,starke2019neural,wang2022towards} in furnished 3D indoor scenes, and object-conditioned grasp pose generation~\cite{li2022gendexgrasp,liu2021synthesizing,jiang2021hand,taheri2020grab,wu2021saga}.
However, most previous methods~\cite{taheri2020grab,wang2022humanise,li2022gendexgrasp,jiang2021hand,wang2022towards,hassan2021populating,hassan2021stochastic,chen2022scept,won2022physics} rely on \ac{cvae} and suffer from the posterior collapse problem~\cite{zhao2017infovae,tolstikhin2018wasserstein,kim2018semi,he2018lagging,fu2019cyclical,yuan2020dlow,wang2021synthesizing}, especially when the 3D scene is natural and complex. In this work, \model addresses the posterior collapse with the diffusion-based denoising process.

\paragraph{Physics-based Optimization in 3D Scenes}

Producing physically plausible generations compatible with 3D scenes is one of the challenges in the scene-conditioned generation. Previous work uses physics-based post-optimization~\cite{hassan2019resolving,liu2021synthesizing,wang2021synthesizing} or differentiable objective~\cite{zhangyan2020generating,wang2022humanise,jiang2021hand} to integrate collision and contact constraints into the generation framework. However, post-optimization approaches~\cite{hassan2019resolving,liu2021synthesizing,wang2021synthesizing} are oftentimes inefficient and cannot be learned jointly with the generative models, yielding inconsistent generation results. Similarly, differentiable approaches~\cite{zhangyan2020generating,wang2022humanise,jiang2021hand} post constraints on the final objective, thus cannot optimize the physical interactions during the sampling, producing implausible generations, especially when adapting to novel scenes. In this work, \model eliminates such inconsistency with the differentiable physics-based optimization integrated into each step of the sampling process.

\paragraph{Planning in 3D Scenes}

The ability to act and plan in 3D scenes is critical for an intelligent agent and has led to the recent culmination of embodied AI research ~\cite{kolve2017ai2,savva2019habitat,shridhar2020alfred,li2021igibson}. Among all tasks, visual navigation has been most studied in the vision and robotics community~\cite{zhu2017target,wortsman2019learning,narasimhan2020seeing,chaplot2020learning,hao2020towards,hong2021vln,zhu2020vision}. However, existing works rely heavily on model-based planning with the single-step dynamic model~\cite{yang2018visual,ramakrishnan2021exploration,wijmans2019dd,du2020learning,chaplot2020object,wahid2021learning}, lacking a trajectory-level optimization for long-horizon planning. Further, the physical interactions are not explicitly modeled into the planning. This deficiency makes it challenging to generalize to natural scenes, where exploration is limited, and fast learning and adaptation are required. In comparison, with the global trajectory planner based on a trajectory-level generator, \model demonstrates better generalization in long-horizon plans and novel 3D scenes.

\paragraph{Diffusion-based Models}

Diffusion model~\cite{sohl2015deep,ho2020denoising,song2019generative,hyvarinen2005estimation} has come forth into a promising class of generative model for learning and sampling data distributions with an iterative denoising process, facilitating the image~\cite{dhariwal2021diffusion,song2020denoising}, text~\cite{yu2022latent}, and shape generation~\cite{luo2021diffusion}. With flexible conditioning, it is further extended to the language-conditioned image~\cite{saharia2022photorealistic,nichol2021glide,rombach2022high}, video~\cite{ho2022imagen,singer2022make}, and 3D generation~\cite{poole2022dreamfusion,tevet2022human,zhang2022motiondiffuse}. Notably, Janner \etal~\cite{janner2022planning} integrate the generation and planning into the same sampling framework for behavior synthesis. To our best knowledge, \model is the first framework that models the 3D scene-conditioned generation with a diffusion model and integrates the generation, optimization, and planning into a unified framework.

\section{Background}

\subsection{Problem Definition}

Given a 3D scene $\cS$, we aim to generate the optimal solution for completing the tasks (\eg, navigation, manipulation) given the goal $\cG$ in the scene. We denote the state and action of an agent as $(\bs, \ba)$. The dynamic model defines the state transition as $p(\bs_{i+1} | \bs_i, \ba_i)$, which is often deterministic in scene understanding (\ie, $f(\bs_i, \ba_i)$). The trajectory is defined as $\boldsymbol \btau = (\bs_0, \ba_0, \cdots, \bs_i, \ba_i, \cdots, \bs_N)$, where $N$ denotes the horizon of task solving in discrete time. 

\subsection{Planning with Trajectory Optimization}

The scene-conditional trajectory optimization is defined as maximizing the task objective:
\begin{equation}
    \btau^* = \argmax_{\btau} \cJ(\btau | \cS,\cG).
    \label{eq:traj}
\end{equation}
The dynamic model is usually known for trajectory optimization. Considering the future actions and states with predictable dynamics, the entire trajectory $\btau$ can be optimized jointly and non-progressively with traditional~\cite{lavalle2006planning} or data-driven~\cite{choudhury2018data} planning algorithms. Trajectory-based optimization benefits from its global awareness of history and future states, thus can better model the long-horizon tasks compared with single-step models in \ac{rl}, where $\ba_{0:N}^* = \argmax_{\ba_{0:
N}} \sum_{i=0}^N r(\bs_i, \ba_i | \cS,\cG)$. 

\subsection{Diffusion Model}

Diffusion models~\cite{hyvarinen2005estimation,sohl2015deep,ho2020denoising} are a class of generative models that represent the data generation with an iterative denoising process from Gaussian noise. It consists of a forward and a reverse process. The forward process $q(\btau^t | \btau^{t-1})$ gradually destroys data $\btau^0 \sim q(\btau^0) $ into Gaussian noise. The parametrized reverse process $p_{\theta}(\btau^{t-1} | \btau^t)$ recovers the data from noise with the learned normal distribution from a fixed timestep. The training objective for $\theta$ is denoising score matching over multiple noise scale~\cite{hyvarinen2005estimation,vincent2011connection}. Please refer to the \cref{app:sec:bg_diffuse} for detailed descriptions of the diffusion model and its variants.

\section{\model}\label{sec:model}

\model models planning as trajectory optimization and solves the aforementioned problem with the spirit of \textit{planning as sampling}, where the trajectory optimization is achieved by sampling trajectory-level distribution learned by the model. Leveraging the diffusion model with gradient-based sampling and flexible conditioning, \model models the scene-conditioned goal-oriented trajectory $p(\btau^0 | \cS,\cG)$:
\begin{equation}
    \begin{aligned}
        p(\btau^0|\cS,\cG)=&\ \frac{p_{\theta}(\btau^0|\cS)p_{\phi}(\cG|\btau^0,\cS)}{p(\cG|\cS)}\\
        \propto &\ p_{\theta}(\btau^0|\cS)p_{\phi}(\cG|\btau^0,\cS).
    \end{aligned}
    \label{eq:model}
\end{equation}

\paragraph{Generation}

$p_{\theta}(\btau^0 | \cS)$ characterizes the probability of generating certain trajectories with the scene condition. It can be modeled using a conditional diffusion model~\cite{sohl2015deep,ho2020denoising} with an iterative denoising process:
\begin{equation}
    \small
    \begin{aligned}
        p_{\theta}(\btau^0 | \cS)  & = p(\btau^T)\prod_{t=1}^T p(\btau^{t-1}|\btau^t,\cS), \\
        p(\btau^{t-1}|\btau^t,\cS)  & = \cN(\btau^{t-1}; \bmu_{\theta}(\btau^t, t, \cS), \bSigma_{\theta}(\btau^t,t,\cS)).
    \end{aligned}
    \label{eq:generation}
\end{equation}

\paragraph{Optimization and Planning}

$p_{\phi}(\cG|\btau^0,\cS)$ represents the probability of reaching the goal with the sampled trajectory, where the goal can be flexibly defined by customized objective functions in various tasks. As shown in \cref{eq:opt_plan}, the precise definition of this probability is $p_{\phi}(\cO=1|\btau^0, \cS, \cG)$, where $O$ is an optimality indicator that represents if the goal were achieved.
Intuitively, the trajectory objective in \cref{eq:traj} can be a good indicator for such optimality. We therefore expand $p_{\phi}(\cG|\btau^t,\cS)$ as its exponential in \cref{eq:opt_exponential}:
\begin{align}
    \label{eq:opt_plan}
    p_{\phi}(\cG|\btau^t,\cS) = &\ p_{\phi}(\cO=1|\btau^t,\cS,\cG) \\
    \label{eq:opt_exponential}
    \propto & \exp(\cJ(\btau^t | \cS,\cG)) \\
    = & \exp(\obj_{p}(\btau^t | \cS, \cG) + \obj_{o}(\btau^t | \cS).
\end{align}
Here, $\obj_o(\btau^t | \cS)$ denotes the objective for optimizing the trajectory with scene condition and is independent of task goal $\cG$. In scene understanding, $\obj_o$ usually denotes plausible physical relationships (\eg, collision, contact, and intersection). $\obj_p(\btau^t | \cS,\cG)$ indicates the objective for planning (\ie, goal-reaching) with scene condition. Both $\obj_o$ and $\obj_p$ can be explicitly defined or implicitly learned from observed trajectories with proper parametrization. 

\subsection{Learning}

$p_{\theta}(\btau^0|\cS)$ is the scene-conditioned generator, which can be learned by the conditional diffusion model with the simplified objective of estimating the noise $\bepsilon$~\cite{ho2020denoising,dhariwal2021diffusion,ho2022cascaded}, where
\begin{equation}
    \small
    \begin{aligned}
        \mathcal{L}_\theta(\btau^0|\cS) = & \Eb{t,\bepsilon,\btau^0}{\bepsilon - \bepsilon_{\theta}(\sqrt{\hat{\alpha}^t}\btau^0+\sqrt{1 - \hat{\alpha}^t}\bepsilon, t, \cS)}\\
        = & \Eb{t,\bepsilon,\btau^0}{\bepsilon - \bepsilon_{\theta}(\btau^t, t, \cS)},
    \end{aligned}
    \label{eq:origin_ddpm}
\end{equation}
where $\hat{\alpha}^t$ is the pre-determined function in the forward process. With the learned $p_{\theta}(\btau^0|\cS)$, we sample $p(\btau^0|\cS,\cG)$ by taking the advantage of the diffusion model's flexible conditioning~\cite{dhariwal2021diffusion,janner2022planning}. Specifically, we approximate $p_{\phi}(\cG|\btau^t,\cS)$ using the Taylor expansion around $\btau^t = \bmu$ at timestep $t$ as
\begin{equation*}
    \log p_{\phi}(\cG|\btau^t,\cS) \approx (\btau^t - \bmu)\bg + C,
\end{equation*}
where $C$ is a constant, $\bmu = \bmu_{\theta}(\btau^t,t,\cS)$ and $\bSigma = \bSigma_{\theta}(\btau^t,t,\cS)$ are the inferred parameters of original diffusion process, and
\begin{equation}
    \begin{aligned}
        \bg = & \nabla_{\btau^t}\log p_{\phi}(\cG|\btau^t,\cS)|_{\btau^t=\bmu} \\
        = & \nabla_{\btau^t}(\obj_{o}(\btau^t | \cS) + \obj_{p}(\btau^t | \cS, \cG))|_{\btau^t=\bmu}.
    \end{aligned}
\end{equation}
Therefore, we have
\begin{equation}
    p(\btau^{t-1}|\btau^{t},\cS,\cG) = \cN(\btau^{t-1}; \bmu + \lambda\bSigma\bg, \bSigma),\label{eq:reverse_final}
\end{equation}
where $\lambda$ is the scaling factor for the guidance. With \cref{eq:reverse_final}, we can sample $\btau^t$ with the guidance of optimizing and planning objectives. 

Of note, $\obj_p$ and $\obj_o$ serve as the pre-defined guidance for tilting the original trajectory with physical and goal constraints. However, they can also be learned from the observed trajectories. During training, we first fix the learned base model of $p_{\theta}(\btau^0|\cS)$, then learn $\phi_o$ and $\phi_p$ for optimization and planning with the following objective:
\begin{equation}
    \mathcal{L}_\phi(\btau^0|\cS,\cG) = \Eb{t,\bepsilon,\btau^0}{\bepsilon - \bepsilon_\theta(\btau^t, t, \cS)-\bSigma\bg}.
    \label{eq:origin_ddpm_short}
\end{equation}
\cref{alg:diffuser:training} summarizes the training procedure.

\begin{algorithm}[ht!]
    \fontsize{8.5}{8.5}\selectfont{}
    \caption{Training of the \model}
    \label{alg:diffuser:training}
    \texttt{// train base generation model}\\
    \SetKwInOut{module}{Modules}
    \KwIn{Trajectory in 3D scene $(\btau^0, \cS)$}
    \Repeat{\text{converged}}{  
        $\btau^0 \sim p(\btau^0 | \cS)$ \\
        $\bepsilon \sim \mathcal{N}({\bf 0}, {\bf I})$, $t \sim \mathcal{U}(\{1,\cdots,T\})$\\
        $\btau^t = \sqrt{\hat{\alpha}_t}\btau_0 + \sqrt{1 - \hat{\alpha}_t}\bepsilon$ \\
        $\theta = \theta - \eta\nabla_\theta \|{\bf\bepsilon} - {\bf\bepsilon}_{\theta}(\btau^t, t, \cS)\|^2_2$
    }
    \texttt{// (optional) train optimization and planning model} \\
    \KwIn{Trajectory in 3D scene with goal $(\btau^0, \cS, \cG)$, learned $\theta$ for $p_\theta(\btau^0|\cS)$}
    \Repeat{\text{converged}}{  
    $\btau^0 \sim p(\btau^0 | \cS, \cG)$ \\
    $\bepsilon \sim \mathcal{N}({\bf 0}, {\bf I})$, $t \sim \mathcal{U}(\{1,\cdots,T\})$\\
    $\bmu = \bmu_{\theta}(\btau^t,t,\cS)$, $\bSigma = \bSigma_{\theta}(\btau^t,t,\cS)$\\
    $\bg = \nabla_{\btau^t}\log p_{\phi}(\cG|\btau^t,\cS)|_{\btau^t=\bmu}$ \\
    $\btau^t = \sqrt{\hat{\alpha}_t}\btau_0 + \sqrt{1 - \hat{\alpha}_t}\bepsilon$ \\
    $\phi = \phi - \eta\nabla_\phi \|{\bf\bepsilon} - {\bf\bepsilon}_{\theta}(\btau^t, t, \cS) - \lambda\bSigma\bg\|^2_2$
}
\end{algorithm}

\subsection{Sampling}

With different sampling strategies, \model can generate, optimize, and plan the trajectory in 3D scenes, under a unified framework of guided sampling. \cref{alg:diffuser:sample} summarizes the detailed sampling algorithm.

\begin{algorithm}[ht!]
    \fontsize{8.5}{8.5}\selectfont{}
    \caption{Sampling of the \model for generation, optimization, and planning}
    \label{alg:diffuser:sample}
    \SetKwInOut{module}{Modules}
    \SetKwProg{Fn}{function}{:}{}
    \SetKwFunction{sample}{{\bf sample}}
    \module{Model $p_{\theta}(\cdot | \cS)$, optimization objective $\obj_o(\cdot|\cS)$, and planner objective $\obj_{p}(\cdot | \cS, \cG)$}
    \texttt{// one-step guided sampling}\\
    \Fn{\sample$(\btau^{t},\cJ)$}{
        $\bmu = \bmu_{\theta}(\btau^t,t,\cS)$, $\bSigma = \bSigma_{\theta}(\btau^t,t,\cS)$\\
        $\btau^{t-1} = \cN(\btau^{t-1}; \bmu + \lambda\bSigma\nabla_{\btau^t}(\cJ(\btau^t | \cS,\cG))|_{\btau^t=\bmu}, \bSigma)$\\
        \Return $\btau^{t-1}$
    }
    \texttt{// physics-based generation}\\
    \KwIn{initial trajectory $\btau^{T}\sim\cN({\bf 0}, {\bf I})$}
    \For{$t=T,\cdots,1$}{
        \texttt{// sampling with optimization}\\
        $\btau^{t-1}=\sample(\btau^{t},\obj_o(\cdot | \cS))$\\
    }
    \Return $\btau^0$ \\
    \texttt{// goal-oriented planning}\\
    \KwIn{planning steps $N$, starting state $\hat{\bs}_0$, initial plan $\btau^{T}_0\sim\cN({\bf 0}, {\bf I})$} 
    $i = 1$ \\
    \While{not done and planning step $i<N$}{
        \For{$t=T,\cdots,1$}{
        
            $\btau^{t-1}_{i} = \sample(\btau^{t}_{i}, \obj_o(\cdot | \cS) + \obj_p(\cdot | \cS,\cG))$\\
            \texttt{// planning as inpainting}\\
            $\btau^{t-1}_{i}[0:i] = \hat{\bs}_{0:i}$\\
        }
        Act $\hat{a}_{i-1}$ to reach $\hat{\bs}_{i} = \btau^{0}_{i}[i]$, $\hat{\bs}_{0:i} = \hat{\bs}_{0:i-1}\cup\hat{\bs}_{i}$\\
        Increment planning step $i = i + 1$\\
    }
\end{algorithm}

\paragraph{Scene-aware Generation}

Sampling $\btau^0$ from the distribution $p_{\theta}(\btau^0|\cS)$ in \cref{eq:generation} directly solves the conditional generation tasks. The sampled trajectories represent diverse modes and possible interactions with the 3D scenes.

\paragraph{Physics-based Optimization}

The physical relations between each state and the environment are defined by $\obj_o$ in \cref{eq:opt_plan} in a differentiable manner. For general optimization without the planning objective, the task goal $\cG$ is to sample a plausible trajectory in 3D scenes. Therefore, we can draw physically plausible trajectories in 3D scenes by sampling from $p(\btau^0|\cS,\cG)$ with \cref{eq:reverse_final}.

\paragraph{Goal-oriented Planning}

The goal-oriented planning can be formulated as motion inpainting under the sampling framework. Given the start state $\hat{\bs}_s$ and the goal state $\hat{\bs}_g$, the planning module returns trajectory $\hat{\btau} = (\hat{\bs}_0,\hat{\ba}_0,\cdots,\hat{\bs}_i,\hat{\ba}_i,\cdots,\hat{\bs}_g)$ that can reach the goal state. We set the first state as $\hat{\bs}_0 = \hat{\bs}_s$ and define the goal state and reward of goal-reaching in $\obj_p$. For each step $i$, we first keep the previous states and inpaint the remaining trajectory by sampling the goal-oriented \model with an iterative denoising process. Next, we take the action that can reach the next sampled state with $(\hat{\ba}_{i-1},\hat{\bs}_{i})$. As illustrated in \cref{alg:diffuser:sample}, we repeat the planning steps until reaching the goal or the maximal planning step. Our planner leverages the trajectory-level generator, thus more generalizable to long-horizon trajectories and novel scenes.

\begin{figure}
    \centering
    \includegraphics[width=\linewidth]{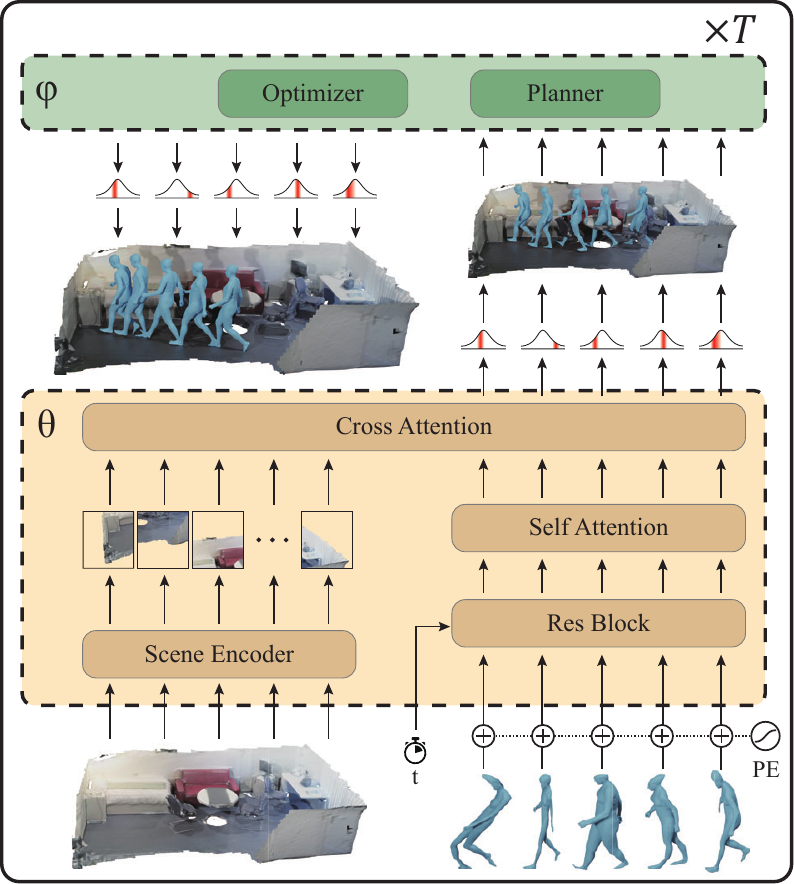}
    \caption{\textbf{Model architecture of the \model.} We use cross-attention to learn the relation between the input trajectory and scene condition. The optimizer and planner serve as the guidance for physically-plausible and goal-oriented trajectories.}
    \label{fig:framework}
\end{figure}

\subsection{Model Architecture}

The design of \model follows the practices of conditional diffusion model~\cite{rombach2022high,saharia2022photorealistic,ho2022cascaded}. Specifically, we augment the time-conditional diffusion model with cross-attention~\cite{vaswani2017attention} for flexible conditioning. As shown in \cref{fig:framework}, for each sampling step, the model computes the cross-attention between the 3D scene condition and input trajectory, wherein the key and value are learned from the condition, and the query is learned from the input trajectory. The computed vector is fed into a feed-forward layer to estimate the noise $\bepsilon$. The 3D scene is processed by a scene encoder (\ie, Point Transformer~\cite{zhao2021point} or PointNet~\cite{qi2017pointnet}). Please refer to the \cref{app:sec:architecture} for details.

\subsection{Objective Design}

For optimization and planning objectives discussed in \cref{sec:model}, we consider two types of trajectory objectives: (i) trajectory-level objective, and (ii) the accumulation of step-wise objective.
For optimization, we consider step-wise collision and contact objective, as well as trajectory level smoothness objective, \ie, $\{\obj_{o}^{\text{collision}}, \obj_{o}^{\text{contact}}, \obj_{o}^{\text{smoothness}}\}$.
For planning, we consider the accumulation of simple step-wise distance \ie, $\obj_{p}^{L_2}$.
Please refer to the \cref{app:sec:objective} for implementation details of our objective design. Empirically, we observe that parameterizing the objectives with timestep $t$ and increasing the guidance during the last several diffusion steps will enhance the effect of guidance. 

\section{Experiments}

To demonstrate \model is general and applicable to various scenarios, we evaluate \model on five scene understanding tasks. For generation, we evaluate the scene-conditioned human pose and motion generation and object-conditioned dexterous grasp generation. For planning, we evaluate the path planning for 3D navigation and motion planning for robot arms. We first introduce the compared methods used in our experiments, followed by detailed settings, results analyses, and ablative studies for each task. Due to the page limit, we refer to the Appendix for more details about the implementation, experimental settings, and additional results and ablations.

\subsection{Compared Methods}

For conditional generation tasks, we primarily compare \model with the widely-adopted \ac{cvae} model~\cite{li2019putting,zhangyan2020generating,jiang2021hand,wang2022humanise,li2022gendexgrasp} and its variants. We also compare with strategies for optimizing the physics of the trajectory in the \ac{cvae}, including integrating into training as loss and plugging upon as the post-optimization. For planning, we compare with a stochastic planner learned by imitation learning using \ac{bc} and a simple heuristic-based deterministic planner guided by $L_2$ distance.

\subsection{Human Pose Generation in 3D Scenes}\label{sec:exp:pose_gen}

\paragraph{Setup}

Scene-conditioned human pose generation aims to generate semantically plausible and physically feasible single-frame human bodies within the given 3D scenes. We evaluate the task on the 12 indoor scenes provided by PROX~\cite{hassan2019resolving} and the refined version of PROX'S per-frame SMPL-X parameters from LEMO~\cite{zhang2021learning}. The input is the colored point cloud extracted by randomly downsampling the scene meshes provided in PROX. Training/testing splits are created following the literature~\cite{zhangyan2020generating,wang2021synthesizing}, resulting in $\sim{}53$k frames in 8 scenes for training and others for testing.

\begin{table*}[t!]
    \centering
    \caption{\textbf{Quantitative results of human pose generation in 3D scenes.} We report metrics for physical plausibility and diversity.}
    \label{tab:pose_gen_quan}
    \resizebox{\linewidth}{!}{%
        \setlength{\tabcolsep}{3pt}%
        \begin{tabular}{cccccccccc}%
            \toprule
            model & plausible rate$\uparrow$ & non-collision score$\uparrow$ & contact score$\uparrow$ & APD (trans.)$\uparrow$ & std (trans.)$\uparrow$ & APD (param)$\uparrow$ & std (param)$\uparrow$ & APD (marker)$\uparrow$ & std (marker)$\uparrow$\\
            \midrule
            \ac{cvae} (w/o. $L_{\text{HS}}$) ~\cite{zhangyan2020generating} & 12.57 & 99.78 & 96.42 & \textbf{1.218} & \textbf{0.494} & 2.878 & 0.166 & 3.638 & 0.172\\
            \ac{cvae} (w/ $L_{\text{HS}}$)~\cite{zhangyan2020generating} & 14.64 & 99.75 & 99.25 & 1.013 & 0.416 & 2.994 & 0.170 & 3.614 & 0.169\\
            our (w/o opt.) & 24.83 & 99.74 & \textbf{99.43} & 0.776 & 0.331 & 3.204 & 0.195 & 3.483 & 0.167\\
            our (w/ opt.) & \textbf{49.35} & \textbf{99.93} & 98.05 & 1.009 & 0.413 & \textbf{3.297} & \textbf{0.197} & \textbf{3.679} & \textbf{0.177}\\
            \bottomrule
        \end{tabular}%
    }%
\end{table*}

\begin{table*}[t!]
    \centering
    \caption{\textbf{Quantitative results of human motion generation in 3D scenes.} We report model variants with and without the start pose.}
    \label{tab:motion_gen_quan}
    \resizebox{\linewidth}{!}{%
        \setlength{\tabcolsep}{3pt}%
        \begin{tabular}{cccccccccc}%
            \toprule
            model & plausible rate$\uparrow$ & non-collision score$\uparrow$ & contact score$\uparrow$ & APD (trans.)$\uparrow$ & std (trans.)$\uparrow$ & APD (param)$\uparrow$ & std (param)$\uparrow$ & APD (marker)$\uparrow$ & std (marker)$\uparrow$ \\
            \midrule
            \ac{cvae} (w/o start)~\cite{wang2022humanise} & 5.88 & 99.86 & 86.26 & \textbf{1.628} & \textbf{0.613} & \textbf{2.766} & \textbf{0.155} & 3.275 & 0.150\\
            ours (w/o start) & \textbf{24.70} & \textbf{99.71} & \textbf{97.92} & 0.568 & 0.237 & 2.339 & 0.126 & 3.299 & 0.151\\
            ours (w/o start \& w/ opt.) & 23.53 & 99.70 & 97.84 & 0.542 & 0.226 & 2.338 & 0.125 & \textbf{3.301} & \textbf{0.151}\\
            \midrule
            \ac{cvae} (w/ start)~\cite{wang2022humanise} & 16.24 & \textbf{99.88} & 95.44 & \textbf{0.478} & \textbf{0.188} & \textbf{1.747} & \textbf{0.091} & \textbf{2.308} & \textbf{0.105}\\
            Ours (w/ start) & 41.76 & 99.85 & \textbf{99.63} & 0.193 & 0.081 & 1.372 & 0.065 & 1.568 & 0.072\\
            Ours (w/ start \& w/ opt.) & \textbf{42.30} & 99.85 & 99.62 & 0.192 & 0.080 & 1.368 & 0.063 & 1.565 & 0.075\\
            \bottomrule
        \end{tabular}%
    }%
\end{table*}

\begin{figure*}[t!]
    \centering
    \includegraphics[width=\linewidth]{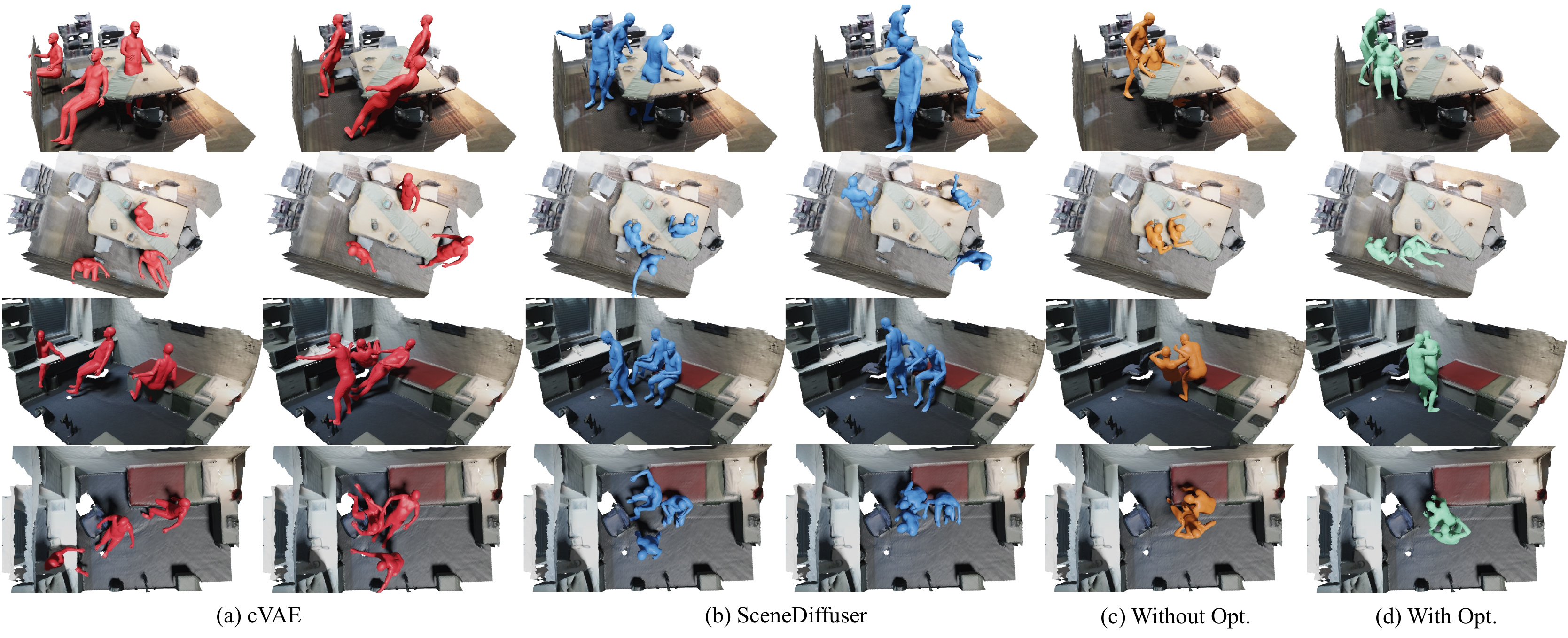}
    \caption{\textbf{Qualitative results of human pose generation in 3D scenes.} From left to right: (a) \ac{cvae} generation, (b) \model generation without optimization, and poses generated (c) with and (d) without applying our optimization-guided sampling.}
    \label{fig:pose_gen_qual}
\end{figure*}

\paragraph{Metrics}

We evaluate the physical plausibility of generated poses with both direct human evaluations and indirect collision and contact scores. For the direct measure, we randomly selected 1000 frames in the four test scenes and instructed seven participants to decide whether the generated human pose was plausible. We compute the mean percentage of plausible generation and term this metric as the plausible rate. For indirect measures, we report (i) the non-collision score of the generated human bodies by calculating the proportion of the scene vertices with positive SDF to the human body and (ii) the contact score by checking if the body contact with the scene in a distance~\cite{hassan2019resolving} below a pre-defined threshold. Following the literature~\cite{yuan2020dlow,zhangyan2020generating}, we evaluate the diversities of global translation, generated SMPL-X parameters, and the marker-based body-mesh representation~\cite{zhang2021we}. Specifically, we calculate the diversity of generated pose with the Average Pairwise Distance (APD) and standard deviation (std). 

\paragraph{Results}

\cref{tab:pose_gen_quan} quantitatively demonstrates that \model generates significantly better poses while maintaining generation diversity. We further provide qualitative comparisons between baseline models and \model in \cref{fig:pose_gen_qual}. While achieving a comparable performance of diversity, collision, and contact, our model generates results that contain considerably more physically plausible poses (\eg, floating, severe collision). This is reflected by the significant superiority (\ie, over 30\%) over \ac{cvae}-based baselines on plausible rates. We observe this large improvement both quantitatively from the plausible rate and non-collision score and qualitatively in \cref{fig:pose_gen_qual}. Notably, our optimization-guided sampling improves the generator with 25\% on the plausible rate, showing the efficacy of the proposed optimization-guided sampling strategy and its potential for a broader range of 3D tasks with physic-based constraints or objectives.

\subsection{Human Motion Generation in 3D Scenes}

\paragraph{Setup}

We consider generating human motion sequences under two different settings: (1) condition solely on the 3D scene, and (2) condition on both the starting pose and the 3D scene. We use the same human and scene representation as in \cref{sec:exp:pose_gen} and clip the original LEMO motion sequence into segments with a fixed duration (60 frames). In total, we obtain 28k motion segments with the distance between each start and end pose being longer than 0.2 meters. We follow the same split in \cref{sec:exp:pose_gen} for training/testing and the same evaluation metrics for the pose generation. We report the average values of pose metrics over motion sequence as our performance measure.

\begin{figure}[t!]
    \centering
   \includegraphics[width=\linewidth]{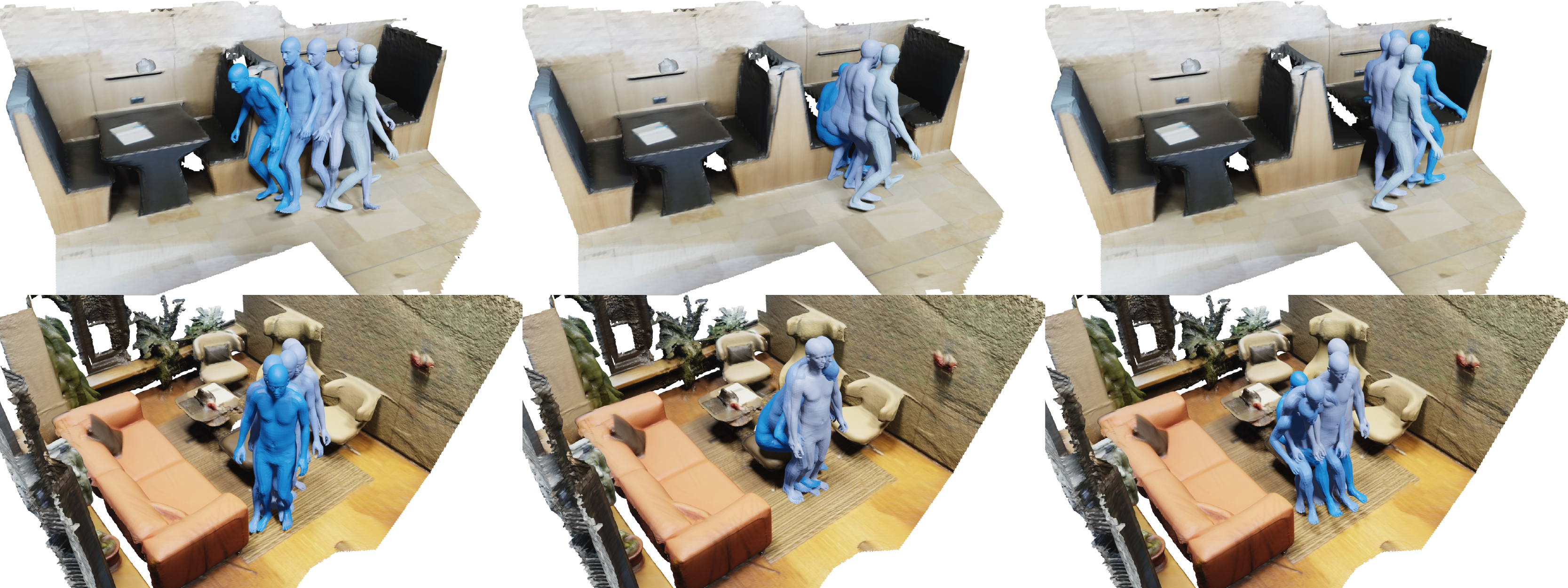}
    \caption{\textbf{Human motions generated by \model.} Each row shows sampled human motions from the same start pose.}
    \label{fig:motion_gen_qual}
\end{figure}

\begin{figure*}[b!]
    \centering
    \begin{subfigure}{0.12\linewidth}\hfill
    \includegraphics[width=0.95\linewidth]{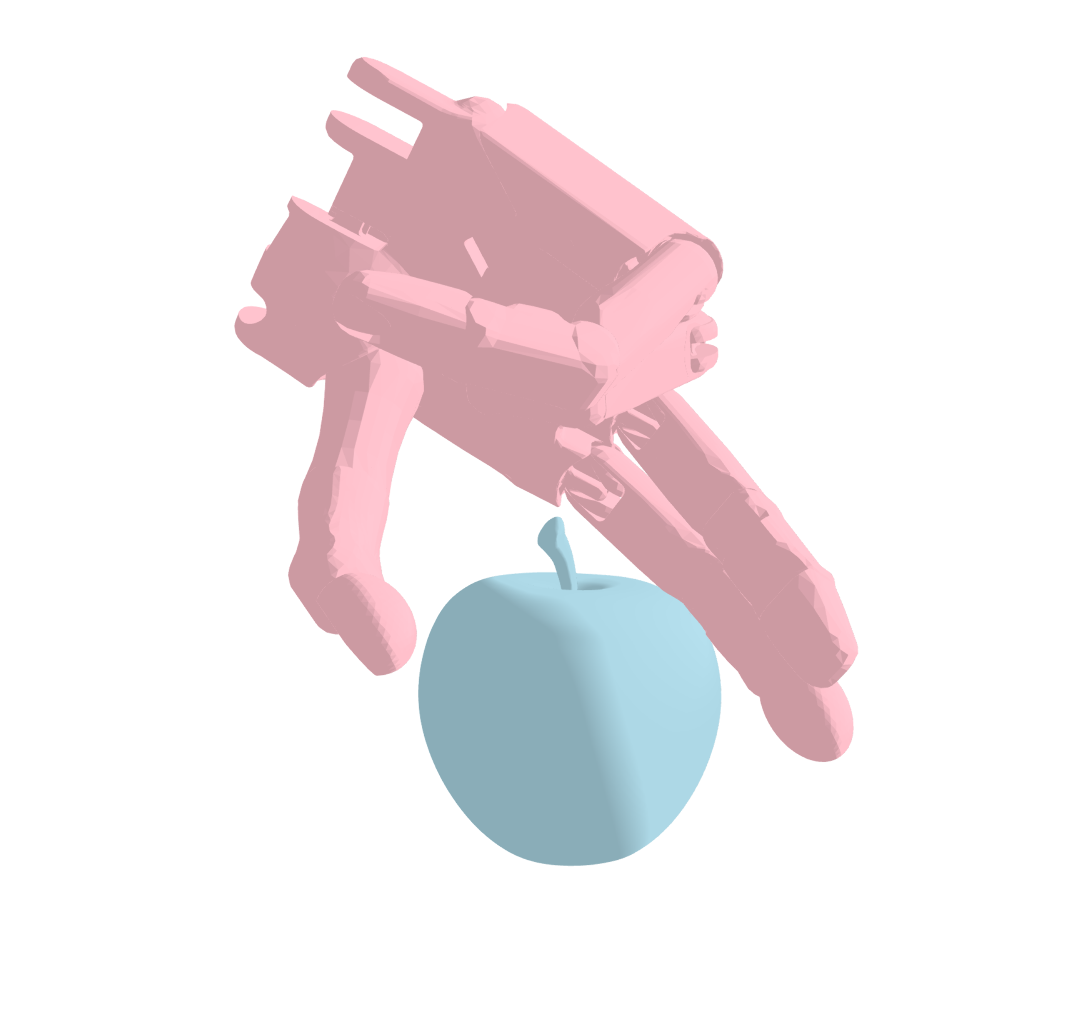} 
    \hfill\end{subfigure} \hfill
    \begin{subfigure}{0.12\linewidth}\hfill
    \includegraphics[width=0.95\linewidth]{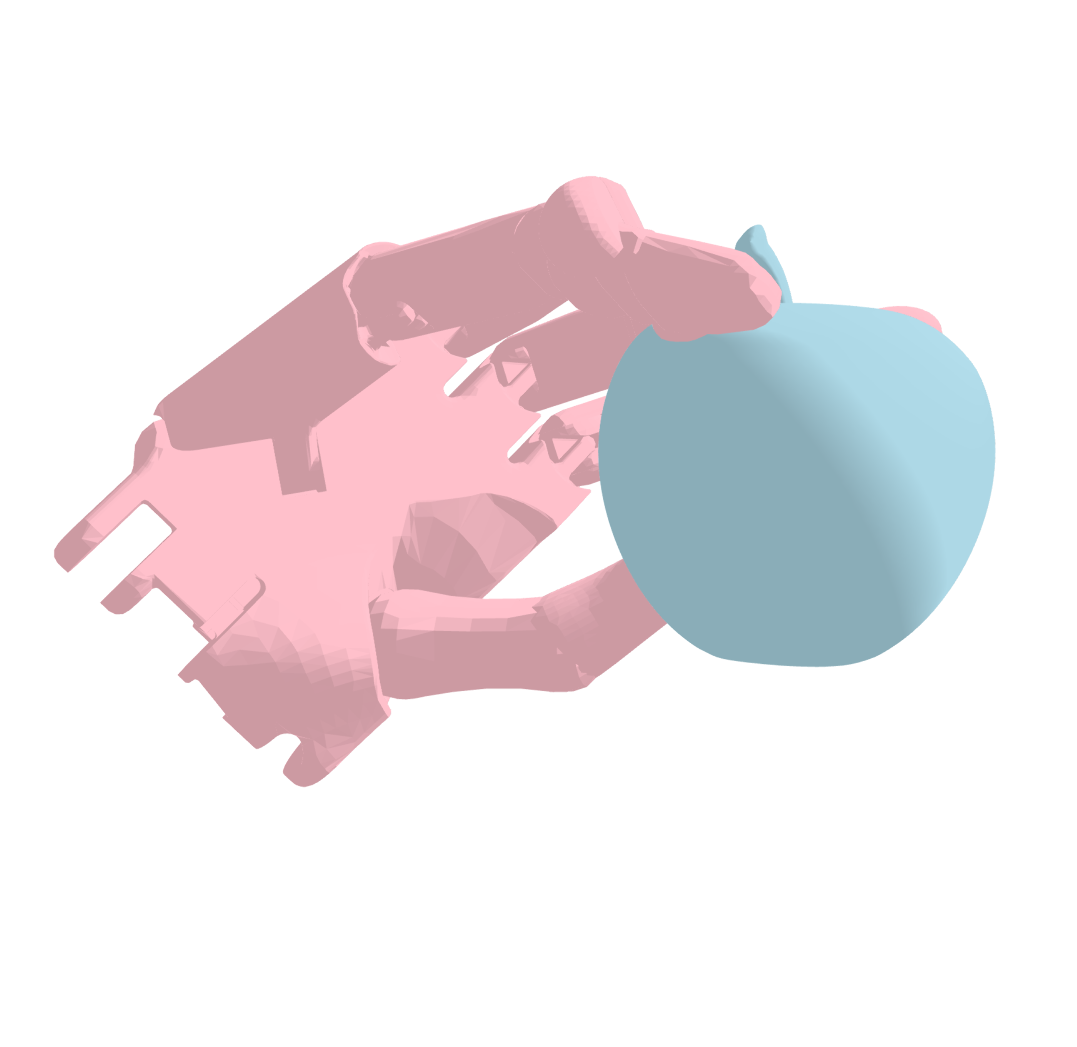} 
    \hfill\end{subfigure} \hfill
    \begin{subfigure}{0.12\linewidth}\hfill
    \includegraphics[width=0.95\linewidth]{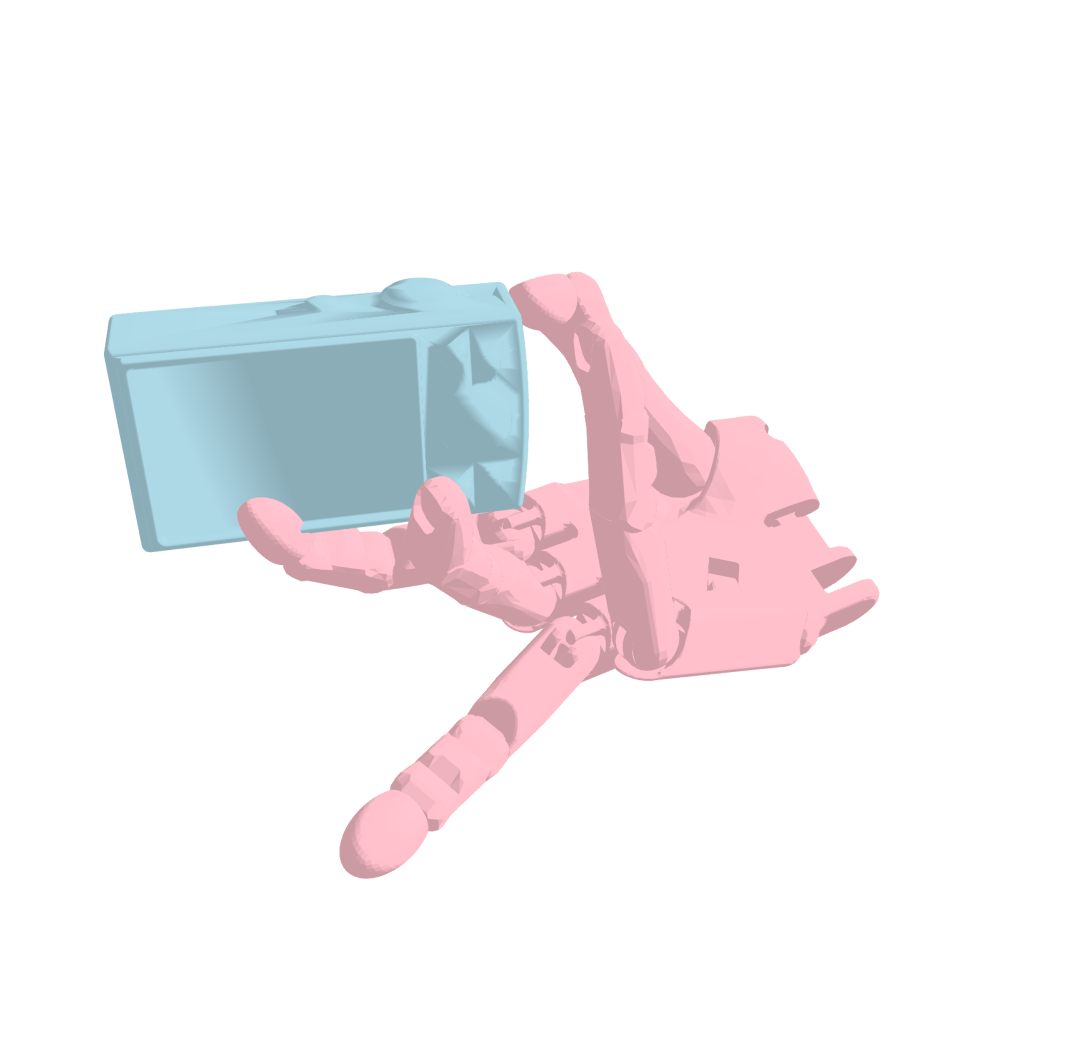} 
    \hfill\end{subfigure} \hfill
    \begin{subfigure}{0.12\linewidth}\hfill
    \includegraphics[width=0.95\linewidth]{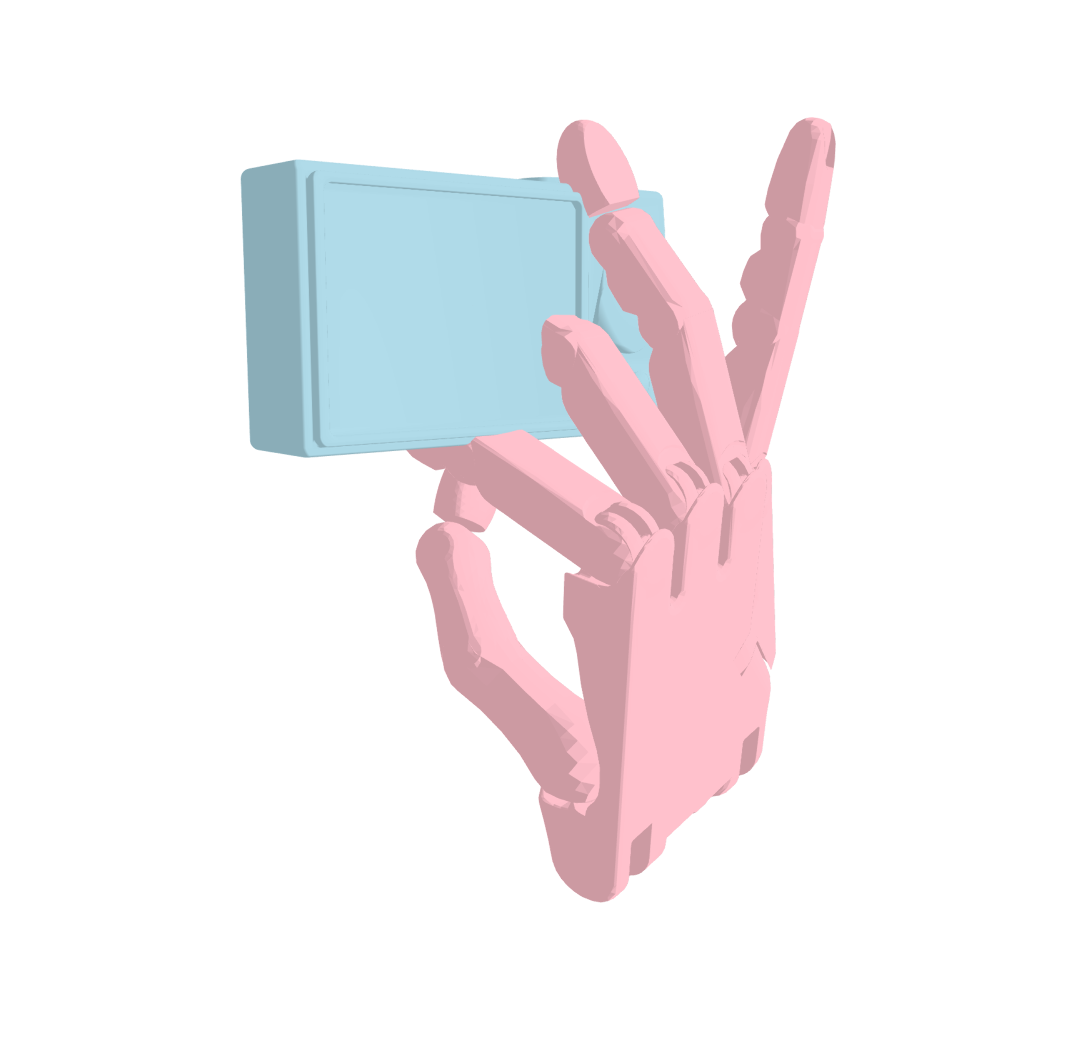} 
    \hfill\end{subfigure} \hfill
    \begin{subfigure}{0.12\linewidth}\hfill
    \includegraphics[width=0.95\linewidth]{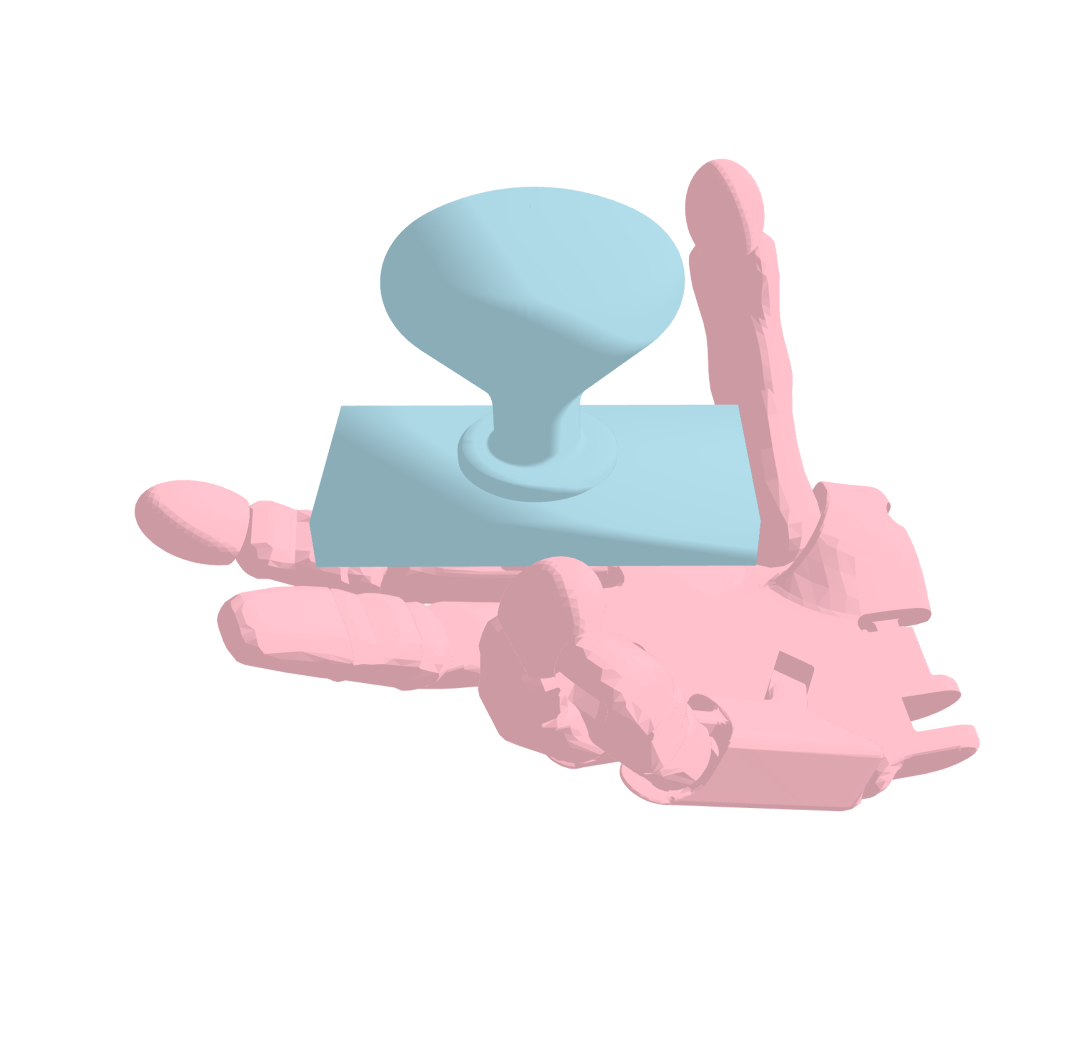} 
    \hfill\end{subfigure} \hfill
    \begin{subfigure}{0.12\linewidth}\hfill
    \includegraphics[width=0.95\linewidth]{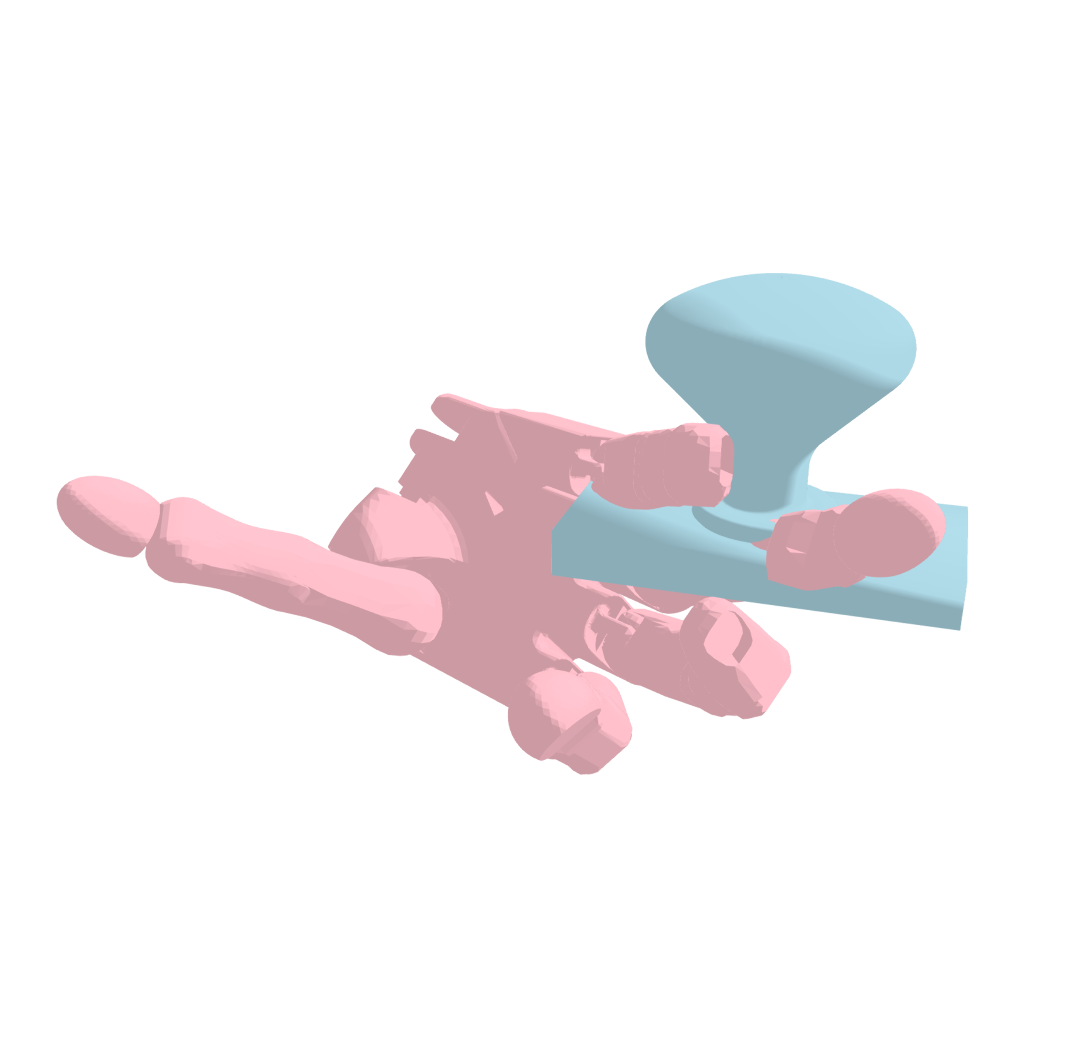} 
    \hfill\end{subfigure} \hfill
    \begin{subfigure}{0.12\linewidth}\hfill
    \includegraphics[width=0.95\linewidth]{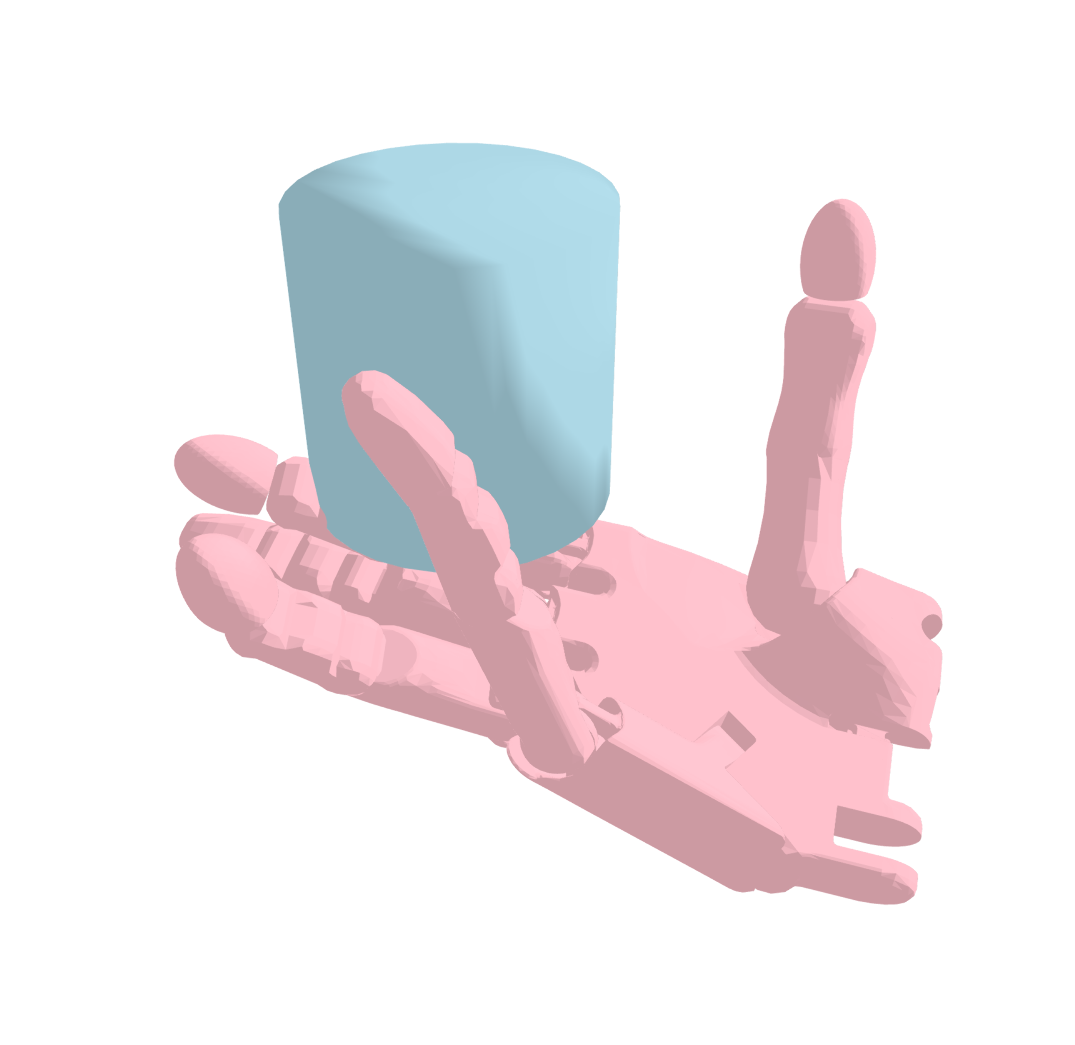} 
    \hfill\end{subfigure} \hfill
    \begin{subfigure}{0.12\linewidth}\hfill
    \includegraphics[width=0.95\linewidth]{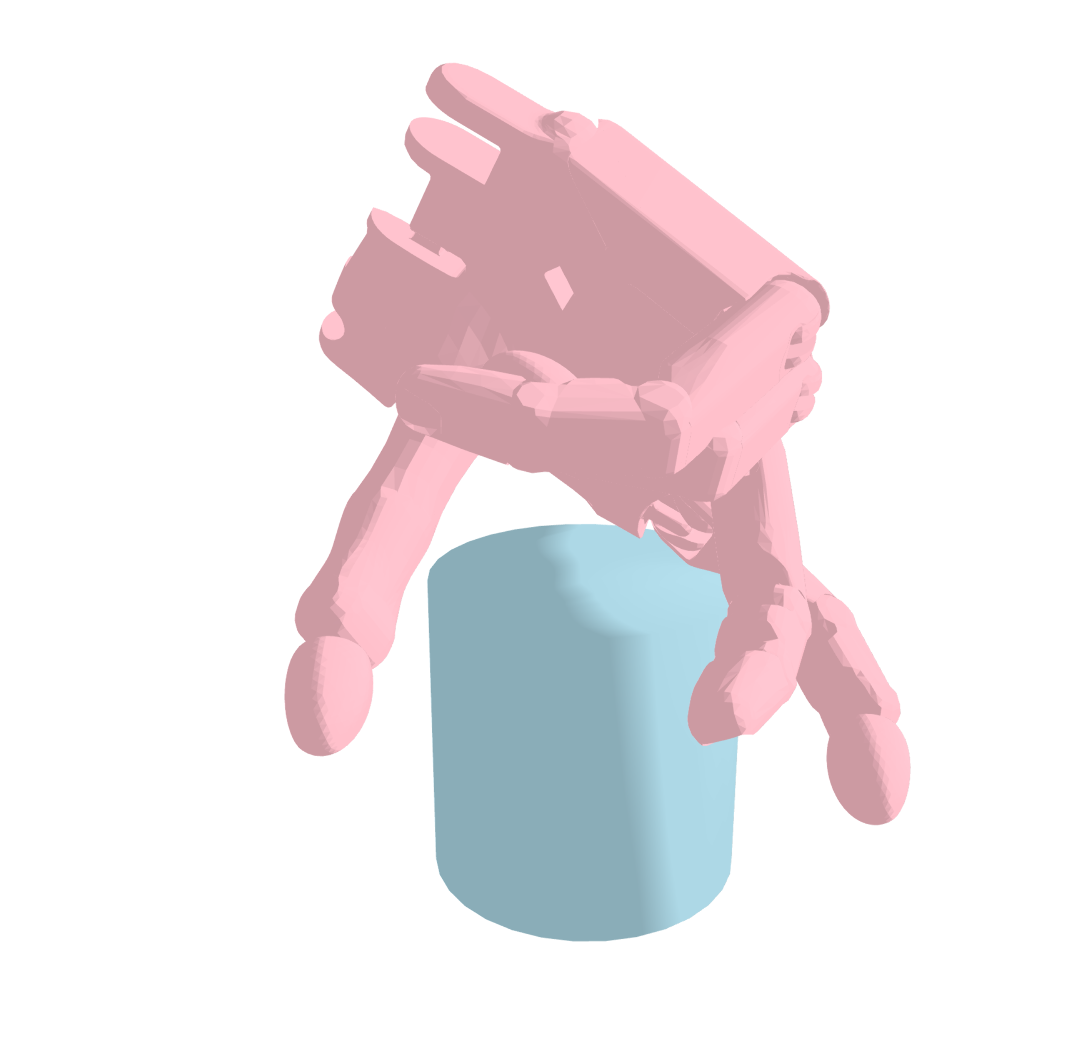} 
    \hfill\end{subfigure} 
    \\
    \begin{subfigure}{0.12\linewidth}\hfill
    \includegraphics[width=0.95\linewidth]{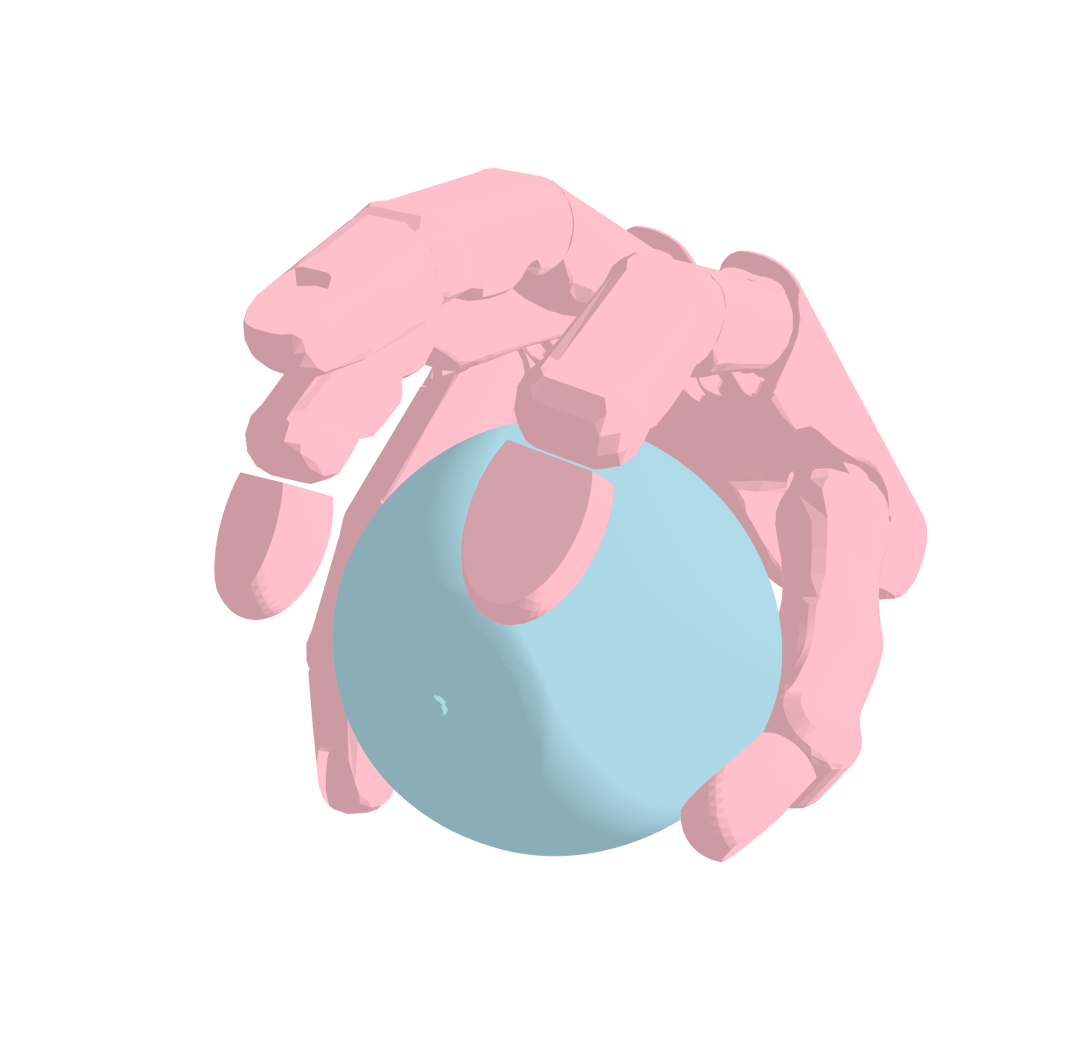} 
    \hfill\end{subfigure} \hfill
    \begin{subfigure}{0.12\linewidth}\hfill
    \includegraphics[width=0.95\linewidth]{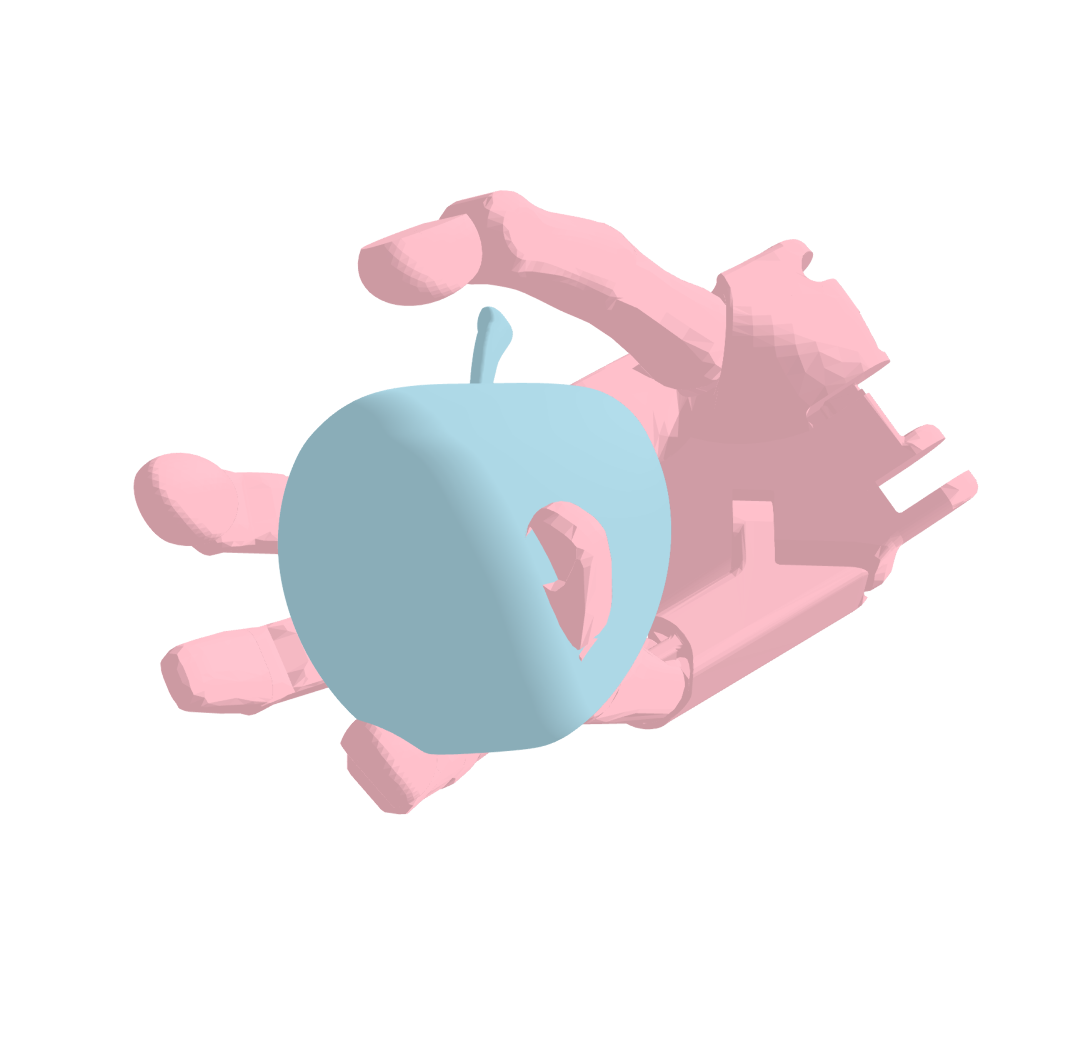} 
    \hfill\end{subfigure} \hfill
    \begin{subfigure}{0.12\linewidth}\hfill
    \includegraphics[width=0.95\linewidth]{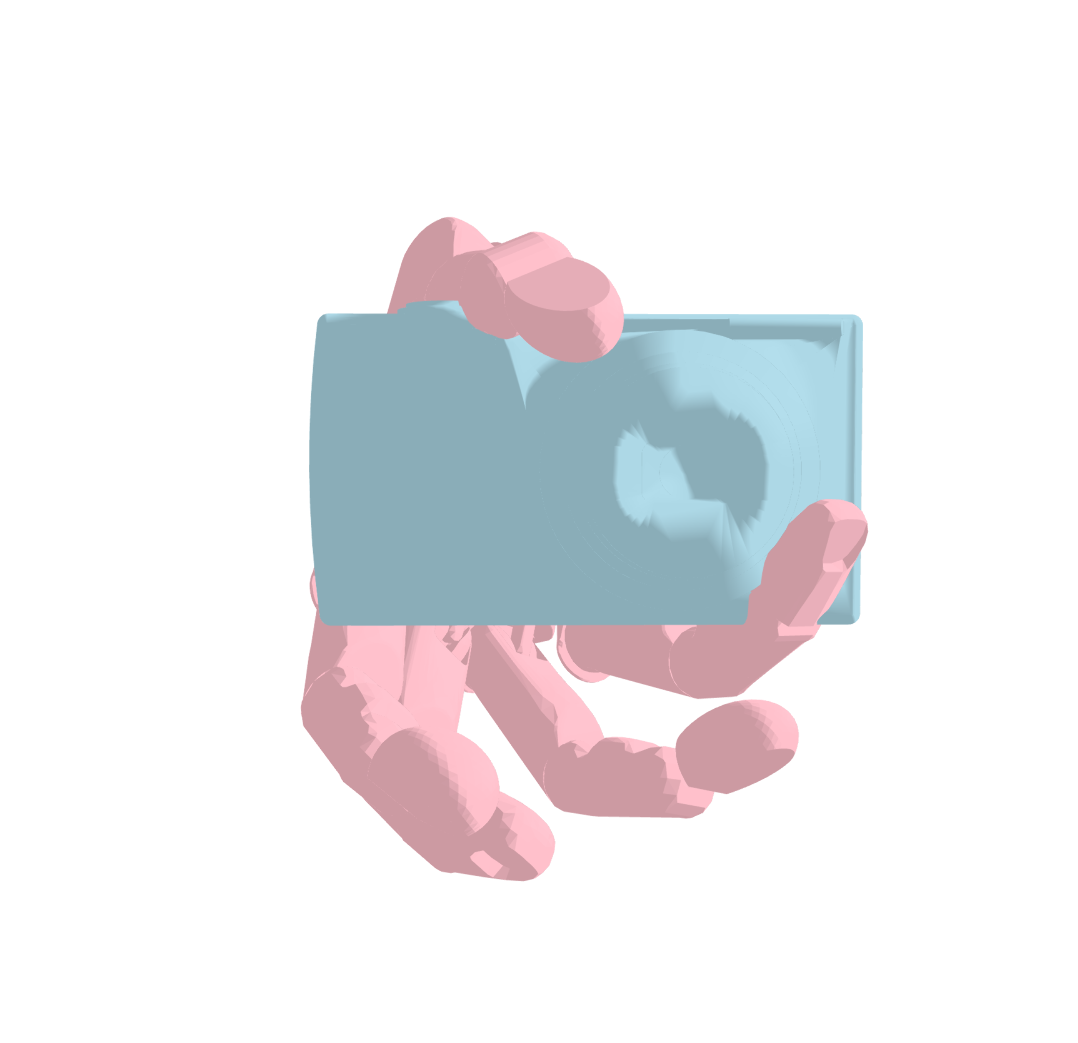} 
    \hfill\end{subfigure} \hfill
    \begin{subfigure}{0.12\linewidth}\hfill
    \includegraphics[width=0.95\linewidth]{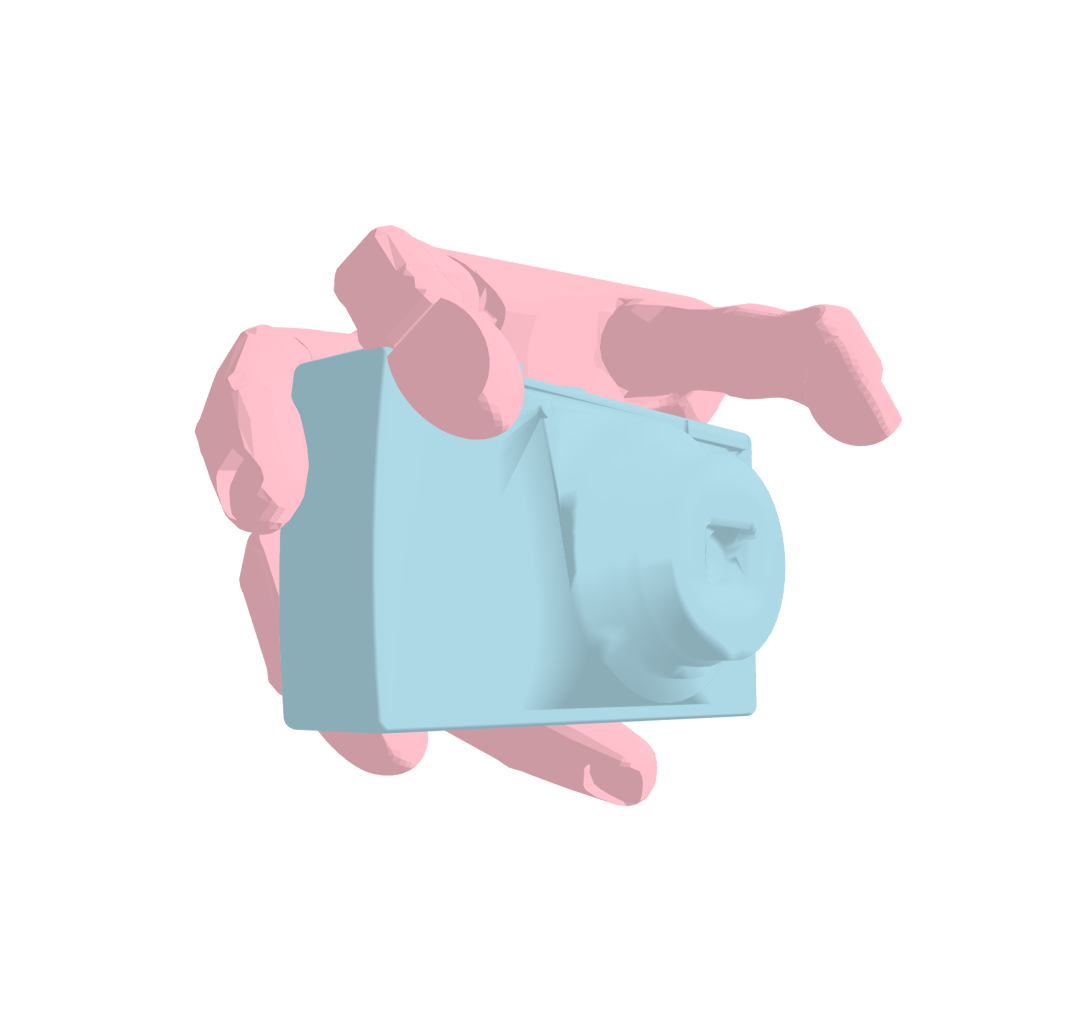} 
    \hfill\end{subfigure} \hfill
    \begin{subfigure}{0.12\linewidth}\hfill
    \includegraphics[width=0.95\linewidth]{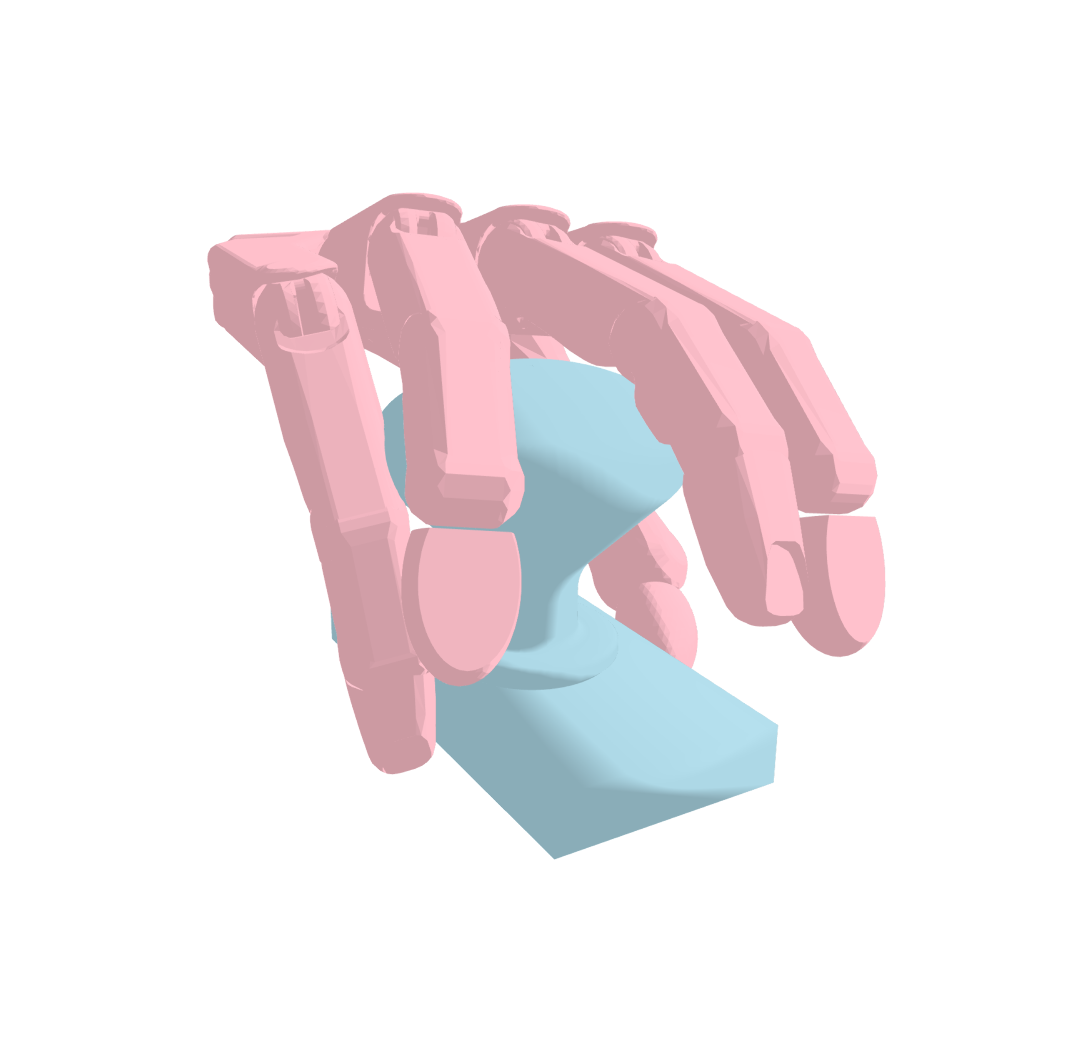} 
    \hfill\end{subfigure} \hfill
    \begin{subfigure}{0.12\linewidth}\hfill
    \includegraphics[width=0.95\linewidth]{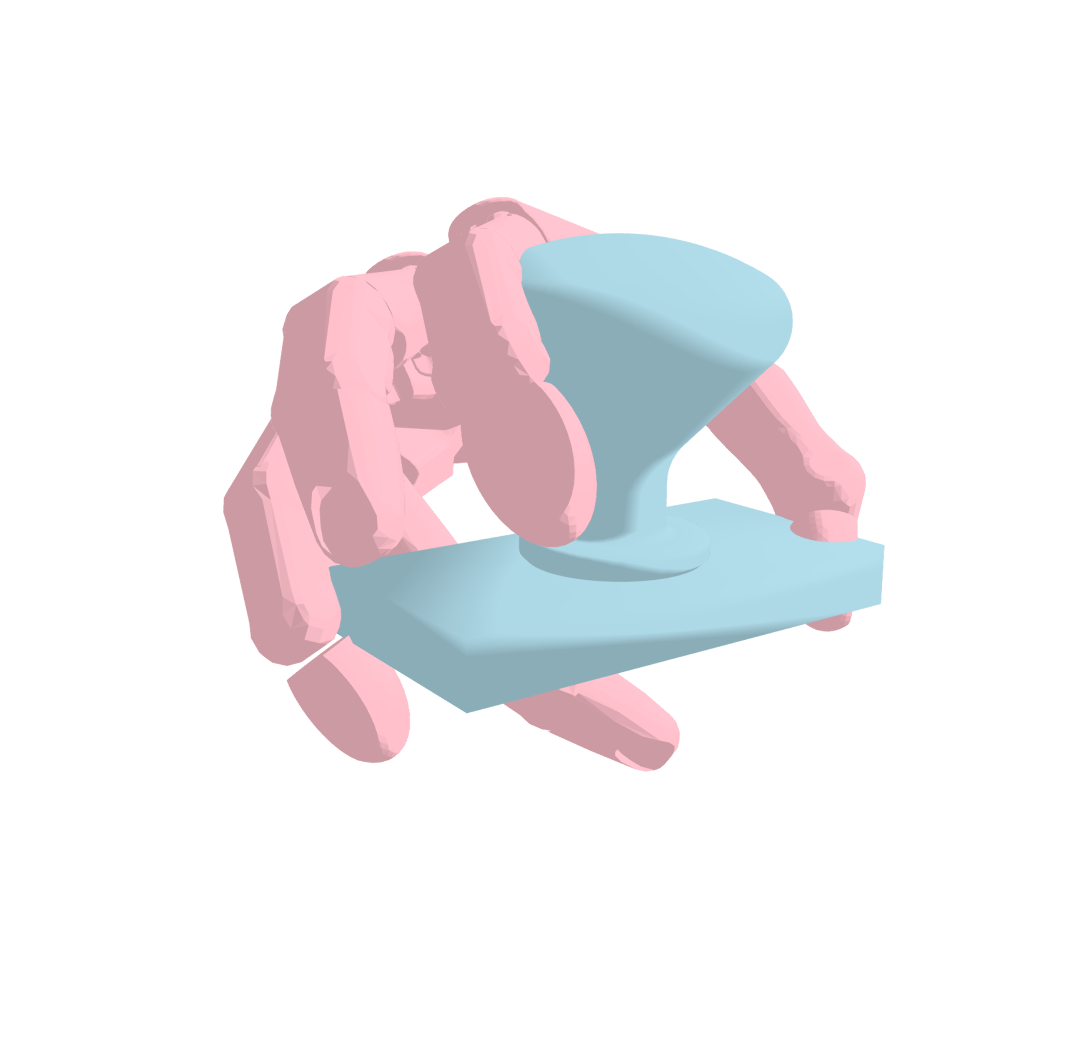} 
    \hfill\end{subfigure} \hfill
    \begin{subfigure}{0.12\linewidth}\hfill
    \includegraphics[width=0.95\linewidth]{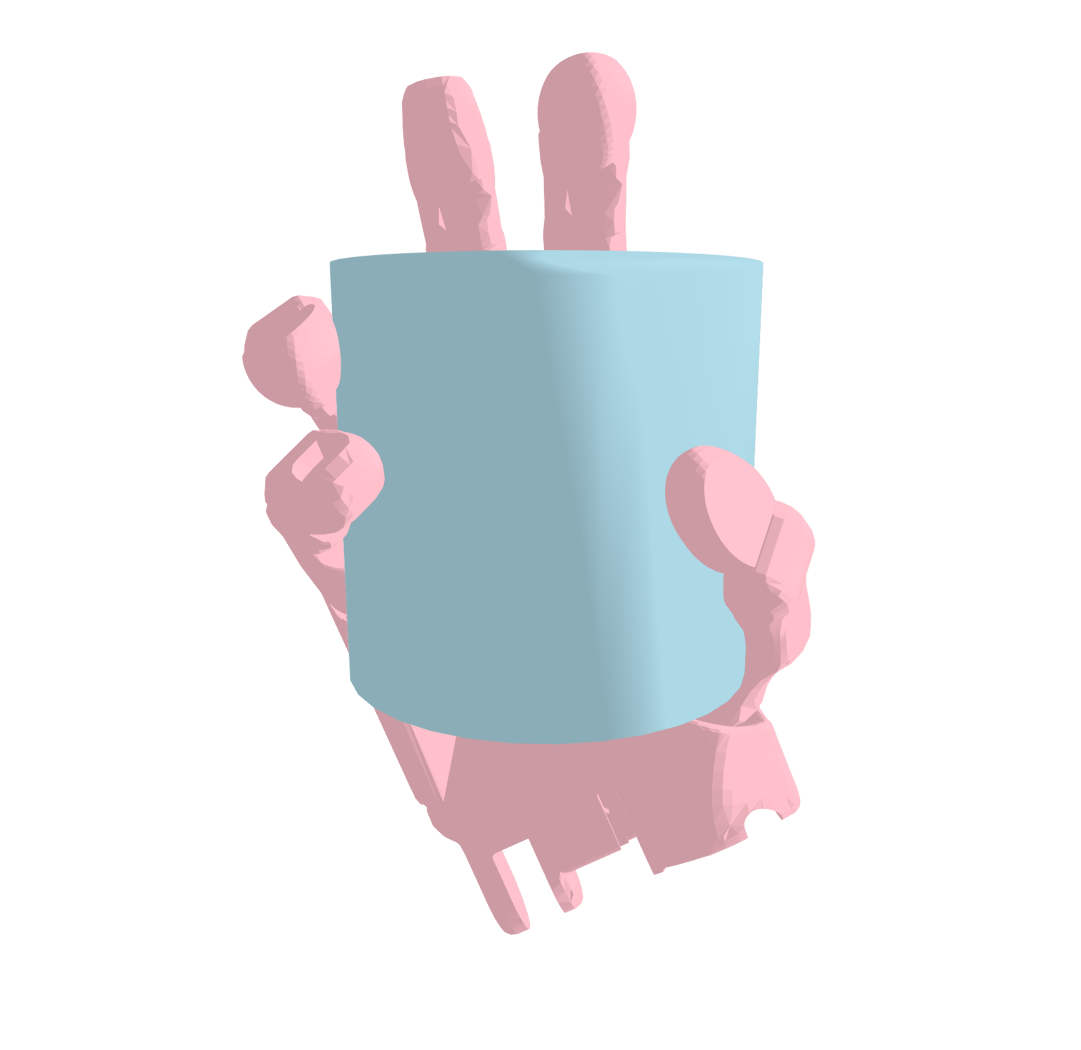} 
    \hfill\end{subfigure} \hfill
    \begin{subfigure}{0.12\linewidth}\centering
    \includegraphics[width=0.95\linewidth]{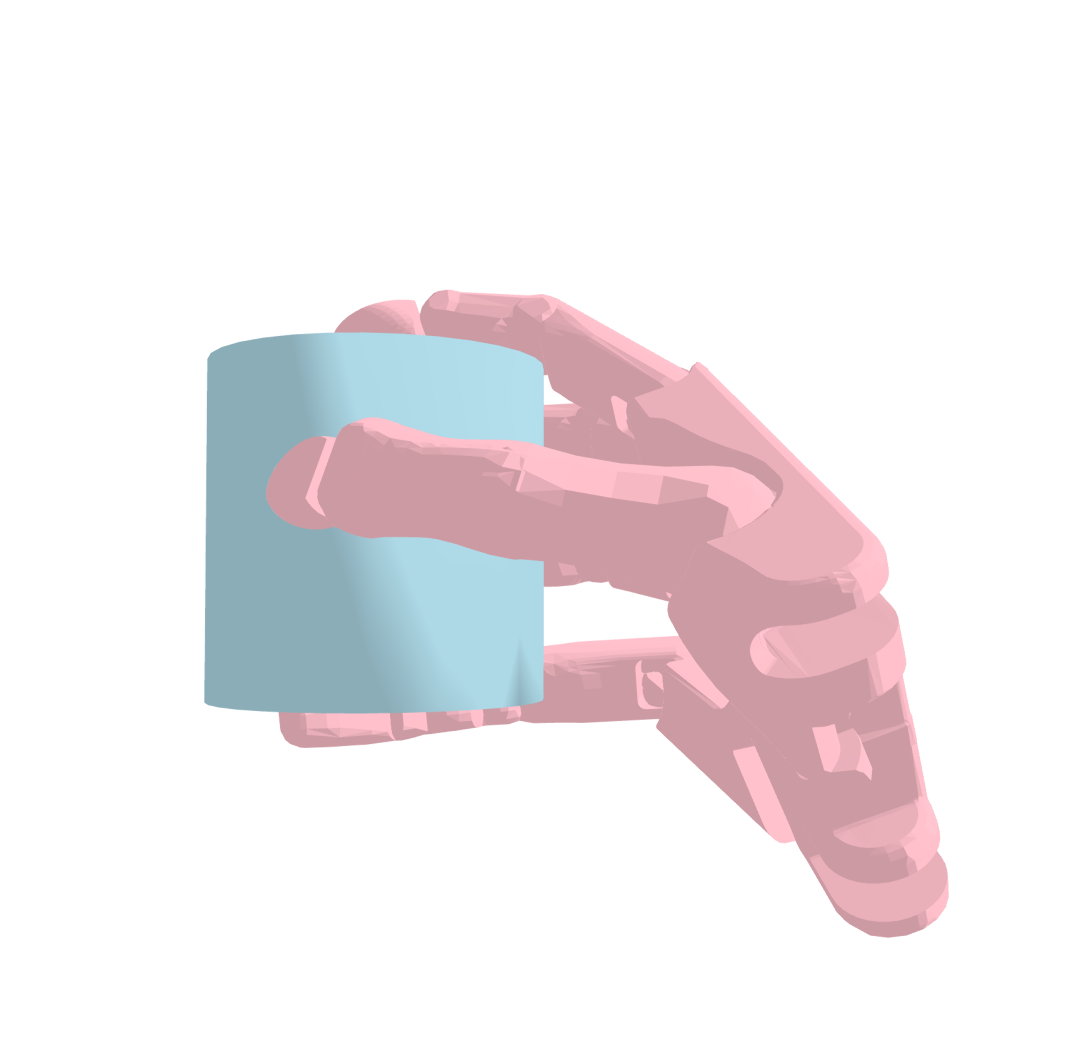} 
    \end{subfigure} 
    \\
    \caption{\textbf{Qualitative results of dexterous grasp generation.} Compared to grasps generated by \ac{cvae} (first row), \model (second row) generates fewer colliding or floating poses, which helps to achieve a higher success rate.}
    \label{fig:grasp_gen_qual}
\end{figure*}

\paragraph{Results}

As quantitatively shown in \cref{tab:motion_gen_quan}, \model consistently generates high-quality motion sequences compared to \ac{cvae} baselines. Specifically, our generated motion outperforms baseline models on both plausible rate and contact scores. This performance gain indicates better coverage of motion that involves rich interaction with the scene while remaining physically plausible. It also causes lower diversity in metrics (\eg, translation variance) since the plausible space for the motion is limited compared with \ac{cvae}. Empirically, we observe that providing the start position of motion as a condition constrains possible future motion sequences and leads to a drop in generation diversity for all models. In addition, providing the start condition benefits the physical plausibility since the motion starts from a plausible pose. We also note only a marginal performance improvement after applying optimization-guided sampling. One potential reason is that the generated motions are already plausible and receive small guidance from the optimization. As qualitatively shown in \cref{fig:motion_gen_qual}, \model generates diverse motions (\eg, ``sit,'' ``walk'') from the same start position in unseen 3D scenes.

\subsection{Dexterous Grasp Generation for 3D Objects}

\paragraph{Setup}

Dexterous grasp generation aims to generate diverse and stable grasping poses for the given object with a human-like dexterous hand. We use the Shadowhand subset of the MultiDex~\cite{li2022gendexgrasp} dataset, which contains diverse dexterous grasping poses for 58 daily objects. We represent the pose of Shadowhand as $q \coloneq (t, R, \theta) \in \bbR^{33}$, where $t \in \bbR^{3}$ and $R \in \bbR^{6}$ denote the global translation and orientation respectively, and $\theta \in \bbR^{24}$ describes the rotation angles of the revolute joints. An object is represented by its point cloud $\cO \in \bbR^{2048 \times 3}$. We split the dataset into 48 seen objects and 10 unseen objects for training and testing, respectively.

\paragraph{Metrics}

We evaluate models in terms of success rate, diversity, and collision depth. We test if a grasp is successful in IsaacGym~\cite{makoviychuk2021isaac} by applying external forces to the object and measuring the movement of the object. To measure how learned models capture the diversity of successful grasping pose in the training data, we report the success rate of generated poses that lies at different variance levels from the mean pose of training data. We measure the collision depth as the maximum depth that the hand penetrates the object in each successful grasp for testing models' performance on physically correct grasps. In all cases, we ignore the root transformation of the hand as it does not contribute to the diversity of grasping types.

\begin{table}[t!]
    \centering
    \caption{\textbf{Quantitative results of dexterous grasp generation on MultiDex~\cite{li2022gendexgrasp} dataset.} We measure the success rates under different diversities and depth collisions. TTA. denotes test-time optimization with physics and contact.}
    \label{tab:grasp_pose_quan}
    \resizebox{\linewidth}{!}{%
        \begin{tabular}{ccccc}
            \toprule
            \multirow{2}[2]{*}{\textbf{model}} & \multicolumn{3}{c}{\textbf{succ. rate} (\%)$\uparrow$} & \multirow{2}[2]{*}{\textbf{depth coll.} (mm)$\downarrow$} \\
            \cmidrule{2-4}
             & $\sigma$ & $2\sigma$  & all &  \\
             \midrule
             \ac{cvae}~\cite{jiang2021hand} & 0.00 & 10.09 & 14.06 & 22.98\\
             \ac{cvae} (w/ TTA.)~\cite{jiang2021hand} & 0.00 & 21.91 & 17.97 & 15.19\\
             \midrule
             ours (w/o opt.) & 70.65 & \textbf{71.25} & \textbf{71.25} & 17.34\\
             ours (w/ opt.) & \textbf{71.27} & 69.84 & 69.84 & \textbf{14.61}\\
            \bottomrule
        \end{tabular}%
    }%
\end{table}

\paragraph{Results}

\cref{tab:grasp_pose_quan} quantitatively demonstrates that \model generates significantly better grasp poses in terms of success rate while correctly balancing the diversity of generation and grasp success. This result indicates that the \model achieves a consistently high success rate without much performance drop when the generated pose diverges from the mean pose in the training data. We also show that, by applying optimizer upon \model, the guided sampling process can reduce the violation of physically implausible grasping poses, outperforming the state-of-the-art baseline~\cite{jiang2021hand} without additional training or intermediate representation (\ie, contact maps). We provide qualitative results in \cref{fig:grasp_gen_qual} for visualization.

\subsection{Path Planning for 3D Scene Navigation}\label{sec:exp:path_planning}

\paragraph{Setup}

\begin{figure}[t!]
    \centering
    \begin{subfigure}{0.5\linewidth}
        \includegraphics[width=\linewidth]{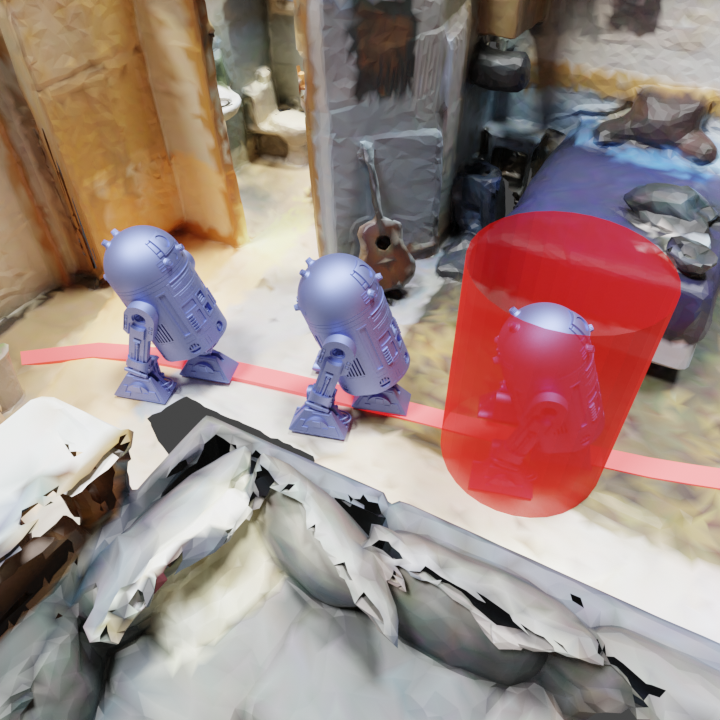}
        \caption{path planning}
        \label{fig:path_planning_setting}
    \end{subfigure}%
    \begin{subfigure}{0.5\linewidth}
	    \includegraphics[width=\linewidth]{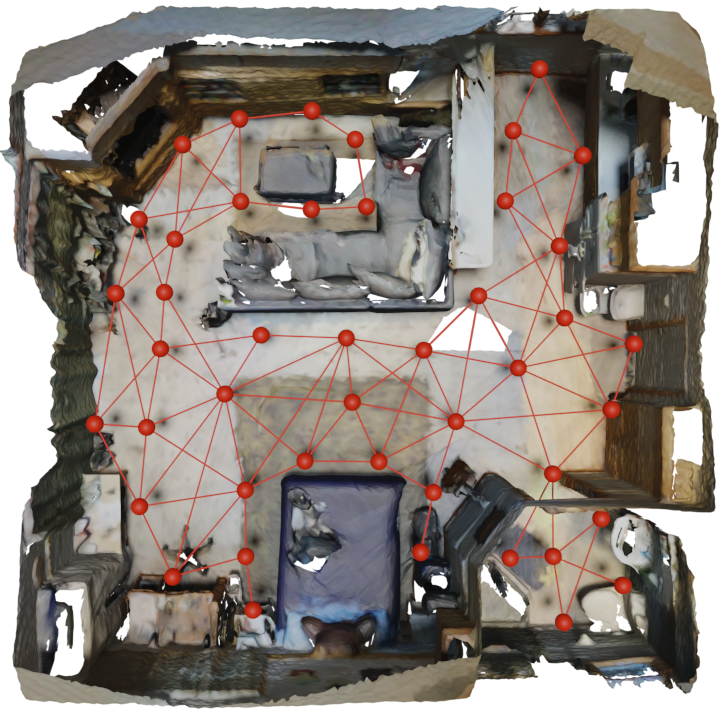}
        \caption{navigation graph}
        \label{fig:path_planning_graph}
    \end{subfigure}%
    \caption{\textbf{Path planning for 3D scene navigation.} \model generates trajectories in long-horizon tasks.}
    \label{fig:path_planning}
\end{figure}

We manually selected 61 indoor scenes from ScanNet~\cite{dai2017scannet} to construct room-level scenarios for navigational path planning and annotated these scenes with navigation graphs. As shown in \cref{fig:path_planning_graph}, these annotations are more spatially dense and physically plausible compared to previous methods~\cite{anderson2018vision}. We represent the physical robot with a cylinder to simulate physically plausible trajectories; see \cref{fig:path_planning_setting}. In total, we collected around 6k trajectories by searching the shortest paths between the randomly selected start and target nodes on the graph. We use trajectories in 46 scenes for training and trajectories in the rest 15 scenes for evaluation. Models take the input as the scene point cloud $\cS \in \bbR^{32768 \times 3}$, a given start position $\hat{\bs}_0 \in \bbR^{2} $, and a target position $\cG \in \bbR^{2}$ on the floor plane.

\paragraph{Metrics}

We evaluate the planned results by checking if the ``robot'' can move from the start to the target without collision along the planned trajectory. We report the average success rate and planning steps over all test cases.

\paragraph{Results}

As shown in \cref{tab:planning_quan}, \model outperforms both the \ac{bc} and the deterministic planner baseline. These results indicate the efficacy of guided sampling with the planning objective, especially given that all test scenes are unseen during training. Crucially, as simple heuristics (like $L_2$) oftentimes lead to dead-ends in path planning, \model can correctly combine past knowledge on the scene-conditioned trajectory distribution and planning objective under specific unseen scenes to redirect planning direction, which helps to avoid obstacles and dead-ends to reach the goal successfully.
Compared with the baseline models, our model also requires fewer planning steps while maintaining a higher success rate.
This suggests that \model successfully navigates to the target without diverging even in long-horizon tasks, where classic RL-based stochastic planners suffer (\ie, the low performance of \ac{bc}).

\begin{table}[t!]
    \centering
    \caption{\textbf{Quantitative results of path planning in 3D navigation and motion planning for robot arms.}}
    \label{tab:planning_quan}
    \resizebox{\linewidth}{!}{%
        \begin{tabular}{cccc}%
            \toprule
            \textbf{task} & \textbf{model} & \textbf{succ. rate}(\%)$\uparrow$ & \textbf{planning steps}$\downarrow$ \\
            \midrule
            \multirow{3}{*}{path plan} & \ac{bc}   & 0 & 150 \\
            & deterministic($L_2$) & 13.50 & 137.98 \\
            & ours & \textbf{73.75} & \textbf{90.38}\\
            \midrule
            \multirow{3}{*}{arm motion} & \ac{bc} & 0.31 & 299.08 \\ 
            & deterministic($L_2$) & 72.87 & \textbf{141.28} \\
            & ours & \textbf{78.59} & 147.60\\
            \bottomrule
        \end{tabular}%
    }%
\end{table}

\subsection{Motion Planning for Robot Arms}

\paragraph{Setup}

\begin{figure}[b!]
    \centering
    \begin{subfigure}{0.5\linewidth}
        \includegraphics[width=\linewidth]{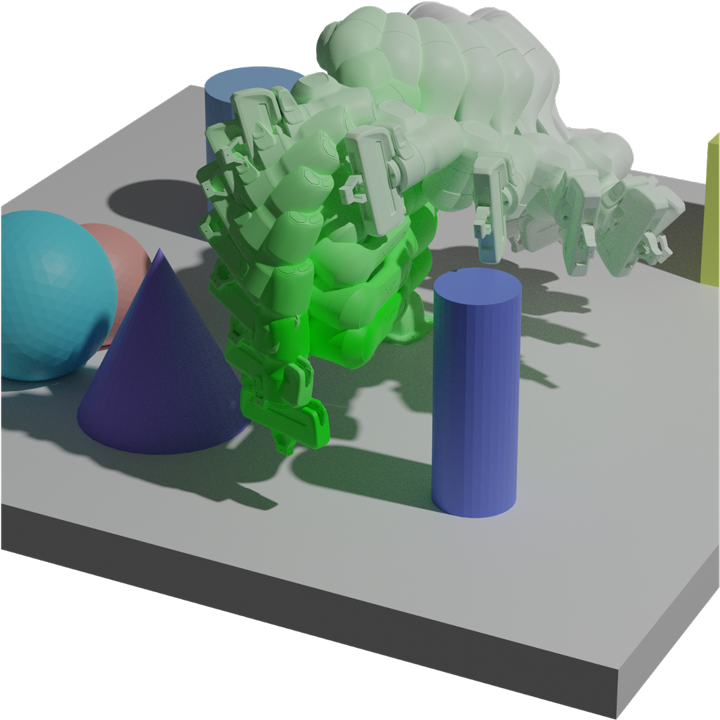}
        \caption{motion planning}
        \label{fig:motion_planning_setting}
    \end{subfigure}%
    \begin{subfigure}{0.5\linewidth}
        \includegraphics[width=\linewidth]{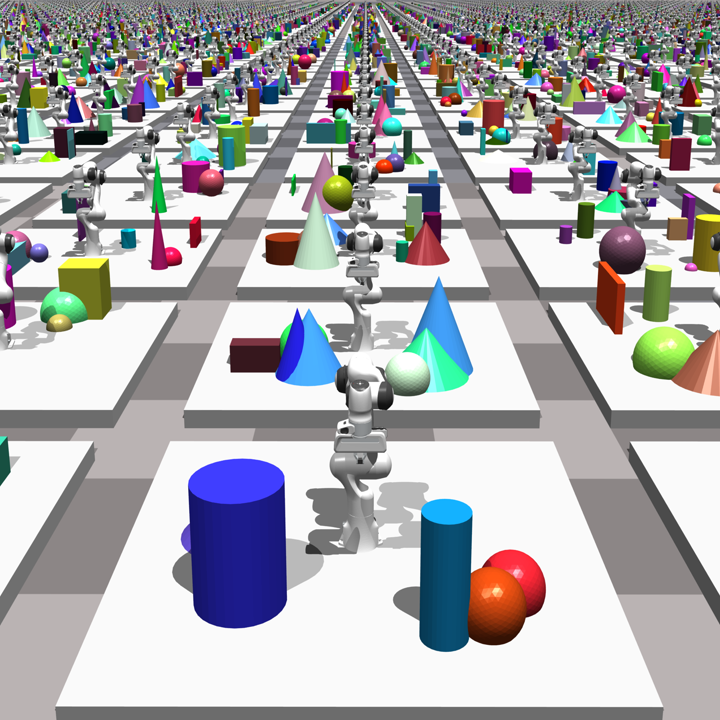}
        \caption{cluttered scenes}
        \label{fig:motion_planning_scene}
    \end{subfigure}%
    \caption{\textbf{Motion planning for robot arms.} \model generates arm motions in tabletop scenes with obstacles.}
    \label{fig:motion_planning}
\end{figure}

Aiming to generate valid robot arm motion trajectories in cluttered scenes, we used the Franka Emika arm with seven revolute joints and collected 19,800 trajectories over 200 randomly generated cluttered scenes using the MoveIt 2.0~\cite{moveit2}, as shown in \cref{fig:motion_planning}. We represent the scene with point clouds $\mathcal{S}\in \mathbb{R}^{4096\times3}$ and the robot arm trajectory with a sequence of joint angles $\mathcal{R}\in[-\pi,\pi]$. We train our model on 160 scenes and test on 40 unseen scenes.

\paragraph{Metrics}

Similar to \cref{sec:exp:path_planning}, we evaluate the generated trajectories by success rate on unseen scenes and the average number of planning steps. We consider a trajectory successful if the robot arm reaches the goal pose by a certain distance threshold within a limited number of steps.

\paragraph{Results}

We observe similar overall performance as in \cref{sec:exp:path_planning}. \cref{tab:planning_quan} shows that \model consistently outperforms both the RL-based \ac{bc} baseline and the deterministic planner baseline. \model's planning steps for successful trials are also comparable with the deterministic planner, showing the efficacy of the planner in long-horizon scenarios.

\subsection{Ablation Analyses}

We explore how the scaling coefficient $\lambda$ influences the human pose generation results. We report the diversity and physics metrics of sampling results under different $\lambda$s, ranging from 0.1 to 100.
As shown in \cref{tab:ablation}, $\lambda$ balances generation collision/contact and diversity in human pose generation. Specifically, $\lambda=1.0$ leads to the best physical plausibility while larger $\lambda$ values lead to diverse generation results. We attribute this effect to the optimization as with bigger $\lambda$s; the optimization will draw poses away from the scene. Due to the page limit, we provide more ablative studies in \cref{app:sec:abl}, including the sampling steps, choices and hyperparameters of objectives, and model architectures.

\begin{table}[t!]
    \centering
    \caption{\textbf{Ablation of the scale coefficient for optimization.}}
    \label{tab:ablation}
    \resizebox{\linewidth}{!}{%
        \begin{tabular}{c|ccccccccc}%
            \toprule
            metric & $\lambda=0.1$ & $\lambda=1.0$ & $\lambda=10.0$ & $\lambda=100.0$ \\
            \hline
            plausible rate $\uparrow$ & 28.75 & \textbf{52.5} & 21.25 & 0\\
            APD (trans.) $\uparrow$ & 0.764 & 0.886 & 1.564 & \textbf{23.96}\\
            APD (param) $\uparrow$ & 3.206 & 3.243 & 9.040 & \textbf{573.6}\\
            non-collison score $\uparrow$ & 99.76 & \textbf{99.87} & 99.85 & 74.9\\
            contact score $\uparrow$ & \textbf{99.70} & 99.65 & 81.75 & 0.0\\
            \bottomrule
        \end{tabular}%
    }%
\end{table}

\subsection{Limitation}

The primary limitation of the \model is its slow training and test speed compared to previous scene-conditioned generative models, a common issue of diffusion-based methods. We also observe that the optimization and planning are highly dependent on the objective designs, which requires efforts on hyper-parameter tuning.

\section{Conclusion}

We propose the \model as a general conditional generative model for generation, optimization, and planning in 3D scenes. \model is designed with appealing properties including scene-aware, physics-based, and goal-oriented. We demonstrate that the \model outperforms previous models by a large margin on various tasks, establishing its efficacy and flexibility.

A promising future direction is extending \model to richer 3D representations, including RGB-D images, semantic images, bird-eye view (BEV) images, videos, 3D meshes, and neural radiance field (NeRF)~\cite{mildenhall2021nerf}. Such flexible conditions consume a tremendous amount of 3D training data, which is also a significant challenge. We also hope to extend the \model to outdoor scenes, \eg, the autonomous driving scenarios~\cite{casas2021mp3}. Moreover, the \model can be combined with recent large language models (LLMs)~\cite{brown2020language} for automatic generation and planning with natural language instructions in 3D scenes, which is promising for the vision and robotics community. Finally, \model can serve as the tool for analyzing the behaviors of humans and agents if we can properly learn the planning objective, which naturally encodes the values and preferences that underlie the trajectories.

\section{Acknowledgement}

We thank Ruiqi Gao and Ying Nian Wu for their helpful discussions and suggestions. This work is supported in part by the National Key R\&D Program of China (2021ZD0150200) and the Beijing Nova Program.

{\small
\balance
\bibliographystyle{ieee_fullname}
\bibliography{reference}
}
\appendix
\renewcommand\thefigure{A\arabic{figure}}
\setcounter{figure}{0}
\renewcommand\thetable{A\arabic{table}}
\setcounter{table}{0}
\renewcommand\theequation{A\arabic{equation}}
\setcounter{equation}{0}
\pagenumbering{arabic}%
\renewcommand*{\thepage}{A\arabic{page}}
\setcounter{footnote}{0}

\section{Background for Diffusion Model}\label{app:sec:bg_diffuse}

A diffusion model~\cite{sohl2015deep,ho2020denoising} is defined by a forward process that gradually corrupts data $\btau^0 \sim q(\btau^0)$ over $T$ timesteps
\begin{align*}
    q(\btau^{1:T} | \btau^0) &= \prod_{t=1}^T q(\btau^t|\btau^{t-1}) \\
    q(\btau^t|\btau^{t-1}) &= \mathcal{N}(\btau^t; \sqrt{1-\beta^t}\btau^{t-1},\beta\bI)
\end{align*}
and a reverse process $p_\theta(\btau^0) = \int p_\theta(\btau^{0:T})d\btau^{1:T}$ where
\begin{align*}
    p_\theta(\btau^{0:T}) &= p(\btau^T)\sum_{t=1}^T p_\theta(\btau^{t-1}|\btau^t) \\
    p_\theta(\btau^{t-1} | \btau^t) &= \mathcal{N}(\btau^{t-1};\bmu_\theta(\btau^t, t), \bSigma_\theta(\btau^t,t)).
\end{align*}
The forward process hyperparameters $\beta^t$ are set so that $\btau^T$ is approximately distributed according to a standard normal distribution, so $\btau^T$ is set to a standard normal prior as well. The reverse process is trained to match the joint distribution of the forward process by optimizing the evidence lower bound (ELBO)~\cite{sohl2015deep,ho2020denoising}. As suggested by the literature~\cite{ho2020denoising,nichol2021glide}, we can use the reverse process parametrizations as:
\begin{align*}
    \bmu_\theta(\btau^t,t) &= \frac{1}{\sqrt{\alpha^t}}(\btau^t - \frac{\beta^t}{\sqrt{1-\hat{\alpha^t}}}\bepsilon_\theta(\btau^t,t)) \\
    \bSigma_\theta^{ii}(\btau^t,t) & = \exp(\log \hat{\beta}^t + (\log\beta^t - \log\hat{\beta}^t)v_\theta^i(\btau^t,t))
\end{align*}
where $\alpha^t = 1 - \beta^t$, $\hat{\alpha}^t=\sum^t_{s=1}\alpha^s$, and $\hat{\beta}^t = \frac{1-\hat{\alpha}^{t-1}}{1-\hat{\alpha}^t}\beta^t$.

We can optimize modified loss instead of the ELBO to improve the sample quality, depending on whether we learn $\bSigma$ or treat it as a fixed hyper-parameter. For the non-learned case, we use the simplified loss:
\begin{align*}
    \mathcal{L}_{simple}(\theta) &= \Eb{t,\bepsilon,\btau^0}{\bepsilon - \bepsilon_\theta(\sqrt{\hat{\alpha}^t\btau^0}+\sqrt{1-\hat{\alpha}^t}\bepsilon,t)}\\
    & = \Eb{t,\bepsilon,\btau^0}{\bepsilon - \bepsilon_\theta(\btau^t, t)}
\end{align*}
It is a weighted form of the ELBO that resembles denoising score matching over multiple noise scale~\cite{ho2020denoising,song2019generative}.

\paragraph{Conditional Diffusion Model}

The goal of the conditional diffusion model is to learn a conditional distribution $p_\theta(\btau^0|\bc)$. We modify the diffusion model to include the condition $\bc$ as input to the inverse process:
\begin{align*}
    p_\theta(\btau^{0:T}|\bc) &= p(\btau^T)\prod^T_{t=1}p_\theta(\btau^{t-1}|\btau^t, \bc) \\
    p_\theta(\btau^{t-1}|\btau^t,\bc) &= \mathcal{N}(\btau^{t-1};\bmu_\theta(\btau^t,t,\bc), \bSigma_\theta(\btau^t,t,\bc))
\end{align*}

\section{Model Architectures}\label{app:sec:architecture}

For the tasks of human pose/motion generation in 3D scenes and path planning for 3D scene navigation, we use the same scene encoder, \ie, the PointTransformer~\cite{zhao2021point} adopted from the original architecture. We pre-train the scene encoder with indoor scene semantic segmentation task on ScanNet dataset and freeze it while training \model. The outputs of the scene encoder are used as the key and value of the cross-attention module.

For processing the trajectory, we employ an FC layer and positional embedding to obtain the high-dimensional feature of the trajectory. We then fuse the trajectory feature with denoising timestep embedding with a ResBlock. After that, we feed the fused feature vectors to a self-attention module and use them as the query of the cross-attention module. Finally, the computed vector is fed into a feedforward layer to estimate the noise $\bepsilon$.

For the task of dexterous grasp generation for 3D objects, we use PointNet~\cite{qi2017pointnet} as the 3D object encoder. Before the cross-attention module, the outputs of PointNet are reshaped to $(N_{\text{token}}, N_{\text{feat}})$, where $N_{\text{token}}$ refers to the number of tokens and $N_{\text{feat}}$ refers to the dimensions of the feature.

For the task of motion planning for robot arms, we adopt PointTransformer~\cite{zhao2021point} as the scene encoder, which is jointly trained from scratch with \model.

\section{Objective Design}\label{app:sec:objective}

For human pose and motion generation in 3D scenes, we encourage contact and non-collision between the generated human body meshes and the scene meshes. Following~\cite{zhangyan2020generating}, we design optimization objective $\obj_o(\btau^t | \cS)=\alpha_{1}\obj_{o}^{\text{collision}} + \alpha_{2}\obj_{o}^{\text{contact}}$ for pose generation and $\obj_o(\btau^t | \cS)=\alpha_{1}\obj_{o}^{\text{collision}} + \alpha_{2}\obj_{o}^{\text{contact}} + \alpha_{3}\obj_{o}^{\text{smoothness}}$ for motion generation. $\alpha$ is the balancing weight.
$\obj_{o}^{\text{collision}}$ minimizes the negative signed-distance values of the body mesh vertices given the negative signed distance field (SDF) of the 3D scene $\Phi_s^-(\cdot)$, which is formulated as
\begin{equation} \label{eq:obj_coll}
    \obj_{o}^{\text{collision}} = -\bbE \left[ |\Phi_{s}^-(\mathcal{M}^t)| \right],
\end{equation}
where $\mathcal{M}^t$ is the SMPL-X body mesh at denoising step $t$.
$\obj_{o}^{\text{contact}}$ minimize the distance between contact body parts of the generated body mesh and the scene mesh, which is formulated as 
\begin{equation}
    \obj_{o}^{\text{contact}} = -\sum_{v_c \in C(\mathcal{M}^t)} \min_{v_s \in \cS} |v_c - v_s|,
\end{equation}
where $C(\cdot)$ is the operation of selecting contact part vertices from the SMPL-X body mesh according to the annotation in~\cite{hassan2019resolving}.
We design the smoothness objective to smooth the motion over time by minimizing the velocity difference of consecutive frames, which is formulated as
\begin{equation}
    \obj_{o}^{\text{smoothness}} = -\sum_{v \in \mathcal{M}^t}\sum_{i=1}^{L-2} \| v^{i+2} - 2v^{i+1} + v^{i}\|^2,
\end{equation}
where $L$ is the length of the motion sequence.
We empirically set $\alpha_1 = 1.0$, $\alpha_2 = 0.02$, and $\alpha_3 = 0.001$.

For dexterous grasp generation, we punish the collision between the robotic hand mesh and the object mesh. We design optimization $\obj_o(\btau^t | \cS)=\obj_{o}^{\text{collision}}$. $\obj_{o}^{\text{collision}}$ is similar to \cref{eq:obj_coll}, where 3D scene is represented as an object and $\mathcal{M}^t$ as the robotic hand mesh at denoising step $t$.

For path planning for the 3D scene navigation task, we design an optimization objective $\obj_{o}$ and $\obj_{p}$ for generating collision-free paths toward the goals.
The collision-free objective maximizes the distance between the robot and the scene vertices in the robot cylinder, formulated as
\begin{equation}
    \obj_{o} = -\sum_{i=1}^{L}\sum_{v_s \in \cS} \text{ReLU}(r - \text{dist}(v_s, \tau^t_i)),
\end{equation}
where $\text{ReLU}(x) = \text{max}(0, x)$, $r$ is the radius of the robot cylinder, and $\text{dist}(\cdot)$ compute the Euler distance between scene vertices and robot position on the 2D plane.
The planning objective $\obj_{p}$ encourages the generated paths toward the target position. In our work, we formulate it as
\begin{equation} \label{eq:obj_plan}
    \obj_{p} = \sum_{i=1}^{L} \exp \left(\frac{1}{\|\cG - \tau^t_i\|_1} \right).
\end{equation}

For robot arm motion planning, we design the planning objective $\obj_{p}$ similar to \cref{eq:obj_plan}.
The objective is defined as
\begin{equation}
    \obj_{p} = \sum_{i=1}^{L} \exp \left(\frac{1}{\sum_{j=1}^{N} \| \cG_j - \tau^t_{ij} \|_1} \right).
\end{equation}
where $N$ denotes to the number of revolute joints and $j$ refers to $j$-th revolute joint.

\section{Implementation Details}\label{app:sec:imple}

\subsection{Human Pose Generation in 3D Scenes}

Following prior work~\cite{zhangyan2020generating}, we represent the human body with the SMPL-X model. We denote the parameters of SMPL-X in our setting as $x_h \coloneq (t, R, \beta, \theta_b)^T \in \bbR^{79}$, where $t$ is the global translation in meters, $R$ is the global orientation represented in axis-angle, $\beta \in \bbR^{10}$ is the body shape feature, and $\theta_b \in \bbR^{63}$ is the axis-angle representation of 21 body joints.
SMPL-X can map these low-dimensional parameters into a watertight mesh with a fixed topology, enabling physical collision and contact modeling.
Unlike~\cite{zhangyan2020generating} using scene depth and semantics, we directly represent the scene with a point cloud $\cS \in \bbR^{32768 \times 3}$, which provides raw information about the 3D scene.

For quantitative evaluation, we randomly sample 1000 examples in each test scene to compute the diversity and physics metrics. Specifically, we separately calculate the Average Pairwise Distance (APD) and standard deviation (std) for global translation $t \in \bbR^{3}$, the rest of local SMPL-X parameters $(R, \beta, \theta_b)^T \in \bbR^{76}$, and the marker-based representation~\cite{zhang2021we} of generated bodies without global translation.
We also report the non-collision score of the generated human bodies by calculating the proportion of the scene vertices with positive SDF to the human body and the contact score by checking whether the body contacts with the scene within a pre-defined distance threshold, \ie, 0.02m.

To train \model, we use Adam~\cite{kingma2014adam} optimizer with 0.0001 as the learning rate. We use 4 NVIDIA A100 GPUs to train 100 epochs with a batch size of 128. The number of diffusion steps $T$ in this task is set as 100. For optimization guidance sampling, we empirically set scale coefficient $\lambda=2.5$.

\subsection{Human Motion Generation in 3D Scenes}

For the two different settings (with and without start position) of human motion generation in 3D scenes, we represent the single-frame human body of the motion sequence as the same as the pose generation. 
To collect training data, we clip the motion sequences in the PROX dataset into motion segments with a fixed duration, \ie, 60 frames.
We use the same evaluation metrics as pose generation and report the average values over motion sequence as the motion generation performance measure.
In this task, the optimizer is Adam, and the learning rate is 0.0001. We use 4 NVIDIA A100 GPUs to train 300 epochs with 200 diffusion steps and 128 batch size. For optimization guidance sampling, we empirically set scale coefficient $\lambda=2.5$.

\subsection{Dexterous Grasp Generation for 3D Objetcs}

We use Shadowhand as our dexterous robotic hand and denote qpos as $q \coloneq (t, R, \theta) \in \bbR^{33}$, where $t \in \bbR^{3}$ and $R \in \bbR^{6}$ represent the global translation and orientation respectively, $\theta \in \bbR^{24}$ describes the rotation angles of the revolute joints. We split the MultiDex~\cite{li2022gendexgrasp} into 48 seen objects and 10 unseen objects for training and testing.

For each grasp, we apply $0.5\mathrm{ms}^{-2}$ acceleration to the object along $\pm xyz$ directions, and the grasping is successful if the movements of the object are all within $2\mathrm{cm}$. For the diversity, we first capture the mean $\mu_i$ and the standard deviation $\sigma_i$ of $i$ revolute joint in the training data grasping pose. We define the mean pose as $\mu_q \coloneq (\mu_1, \mu_2, ..., \mu_{24}) \in \bbR^{24}$ and the standard deviation pose as $\sigma_q \coloneq (\sigma_1, \sigma_2, ..., \sigma_{24}) \in \bbR^{24}$. We report the success rate of generated poses that lie at the $k$ standard deviation level, which means these poses $q$ satisfy the constraint as $\mu_q - k \sigma \leq q \leq \mu_q + k \sigma$. For the depth collision computation, we sample the surface points $\mathcal{H} \in \bbR^{3200 \times 3}$ on the ShadowHand related to the pose $q$ and the surface points with normal $\mathcal{O} \in \bbR^{4096 \times 6}$ on the object. We compute the collision for ShadowHand surface to the object and report the depth collision among $\mathcal{H}$ to show the quality of generated poses.

To train \model on this task, we use Adam optimizer, set the learning rate as 0.0001, and use 1 NVIDIA A100 GPU to train 2100 epochs with 64 batch size. For optimization guidance sampling, we empirically set scale coefficient $\lambda=1.0$.

\subsection{Path Planning for 3D Scene Navigation}

In this task, we consider 3D navigation in realistic scenes,  where the goal is to plan plausible trajectories for a physical robot from the given start position $\hat{\bs}_0$ to the given target position $\cG$ in a furnished 3D indoor scene $\cS$. We represent the hallucinated physical robot as a cylinder to simulate physically plausible trajectories which are collision-free in the 3D scene. The robot can move in all directions within a distance in each step without height change. We set the maximum moving distance as 0.08m, the robot radius as 0.08m, and the robot height as infinite, which means the robot can only move on the floor that is not occupied.

\begin{figure*}[t!]
    \centering
    \includegraphics[width=\linewidth]{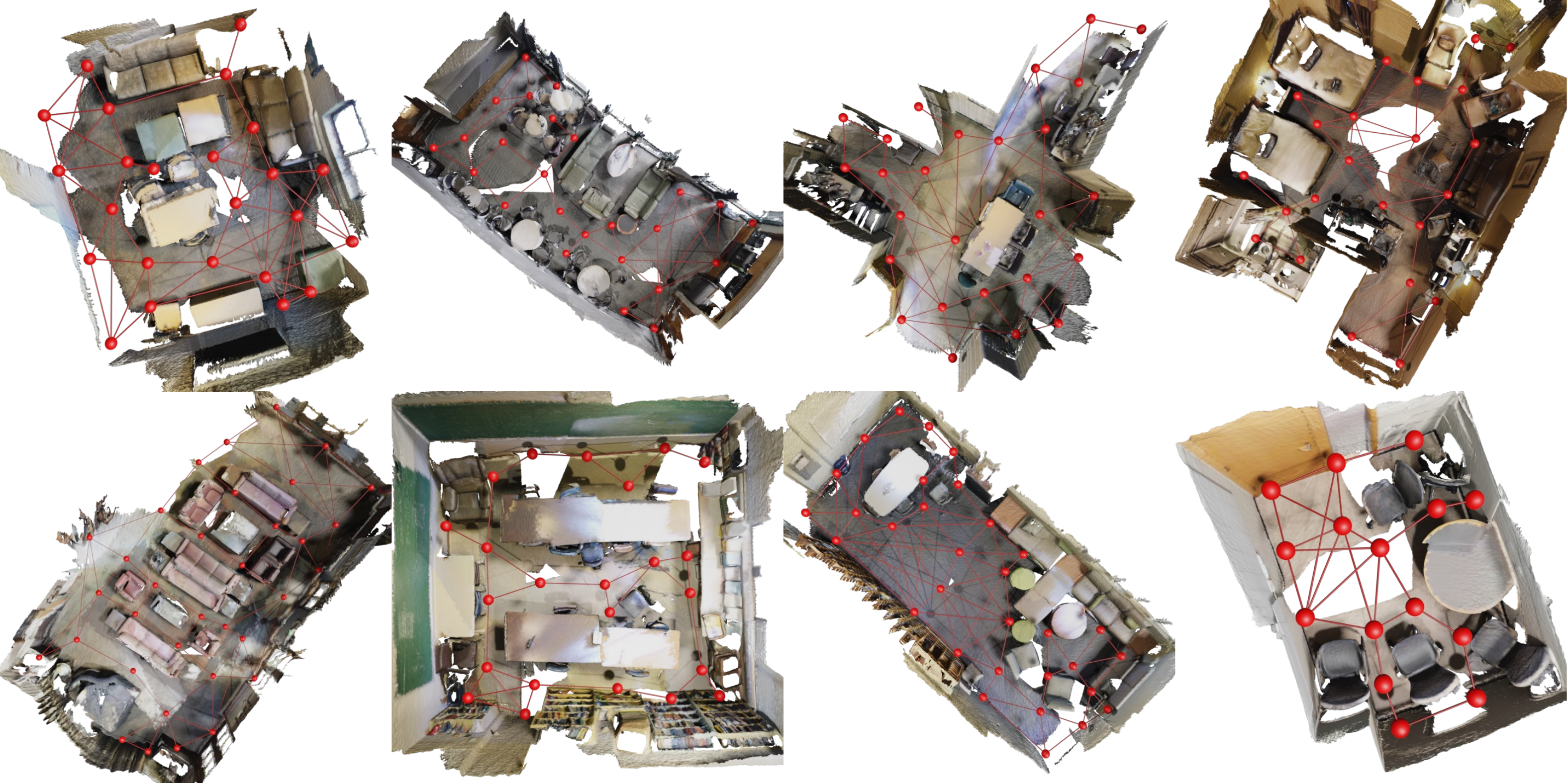}
    \caption{\textbf{Scenes and corresponding navigation graphs for path planning.} The selected scenes have various regions, diverse room types, and sufficient layout complexity.}
    \label{fig:supp:scene_graph}
\end{figure*}

To construct room-level realistic scenarios for path planning, we manually select 61 indoor scenes from ScanNet~\cite{dai2017scannet}, as shown in \cref{fig:supp:scene_graph}.
We annotate these scenes with spatially dense and physically plausible navigation graphs and collect about 6.3k trajectories by searching the shortest paths between the randomly selected start and target graph nodes. As the distance between nodes may be too long for a robot to move in one step, we refined the trajectories according to the maximum moving distance. These trajectories have an average step of 60.0, a minimal step of 32, and a maximum step of 120. We use 4.7k trajectories in 46 scenes as the training data and the rest 1.6k trajectories in 15 unseen scenes for evaluation. We set the maximum number of planning steps as 150.

During training, we set the fixed trajectory horizon as 32. We use 4 NVIDIA A100 GPUs to train 50 epochs with 100 diffusion steps and a batch size of 128. The optimizer is Adam, and the learning rate is 0.0001. During inference, we empirically set the scale coefficient of optimization guidance as 1.0 and the scale coefficient of planning guidance as 0.2.

\subsection{Motion Planning for Robot Arm}

We use the Franka Emika with seven revolute joints as the robot arm and randomly generate cluttered tabletop scenes with primitives following specific heuristics. For each scene, we position the robot arm at the center of the table and use moveit2 motion planner~\cite{moveit2} to synthesize trajectories constrained by a pair of start and goal poses of the end effector. We collected 19,800 collision-free trajectories over 200 clustered scenes.

During inference, we execute the planned motions of \model in IsaacGym~\cite{makoviychuk2021isaac}. We consider the planning is successful if our robot arm reaches the goal pose by a certain $L_2$ norm distance (e.g., $0.2$) in the space of revolute joints. Note that the simulation can not run infinitely; therefore, we set a limited number of simulation steps (e.g., $300$). For the efficiency evaluation, we capture the average number of simulation steps.

To train \model on this task, we use Adam optimizer, set the learning rate as 0.0001, and use 4 NVIDIA A100 GPUs to train 200 epochs with 128 batch size. We empirically set the scale coefficient of planning guidance as 0.2 during inference.

\subsection{Scaling Factor for the Guidance}

Similar to Ho \etal~\cite{ho2020denoising}, we notice that the parameter $\Sigma$ in \cref{eq:reverse_final} decreases as the denoising step $t$ decreases, which gradually weakens the guidance during the denoising process. Instead of using a constant as the scaling factor, we empirically schedule the scaling factor by dividing it by $\Sigma$. It reformulates \cref{eq:reverse_final} as
\begin{equation*}
    p(\btau^{t-1}|\btau^{t},\cS,\cG) \approx \cN(\btau^{t-1}; \bmu + \lambda\bg, \bSigma)
\end{equation*}

\section{Additional Ablative Experiments}\label{app:sec:abl}

We ablate different model architectures, including the scene encoder and noise prediction module in \model, diffusion steps and scale coefficient in the optimizer of dexterous grasp generation task, and fixed frames and planning objectives of path planning for 3D scene navigation task.

\subsection{Model Architecture}

As shown in \cref{tab:supp:abl_scene_enc}, we study how different scene model influences the dexterous grasp generation results. We use PointNet~\cite{qi2017pointnet} and PointNet++~\cite{qi2017pointnet++} as different scene models to extract the object feature. For more diversity evaluation, we capture the mean standard deviation among all revolute joints of the robotic hand qpos. We find that the global feature extracting from PointNet makes it easier for the model to learn a mean pose to obtain a higher grasping success rate. In contrast, the local feature extracting from PointNet++ makes the generated grasp pose more diverse. 

\begin{table}[ht!] 
    \centering
    \caption{\textbf{Ablation on different scene encoder.}}
    \label{tab:supp:abl_scene_enc}
    \resizebox{\linewidth}{!}{%
        \begin{tabular}{cccccc}
            \toprule
            \multirow{2}[2]{*}{\textbf{Scene Encoder}} & \multicolumn{3}{c}{\textbf{Succ. Rate} (\%)$\uparrow$} & \multirow{2}[2]{*}{\textbf{Div.} (rad.)$\uparrow$} &\multirow{2}[2]{*}{\textbf{Coll.} (mm)$\downarrow$} \\
            \cmidrule{2-4}
             & $\sigma$ & $2\sigma$  & all &  \\
             \midrule
             PointNet (w/o opt.) & 70.65 & 71.25 & 71.25 & 0.0718 & 17.34 \\
             PointNet (w/ opt.) & 71.27 & 70.32 & 69.84 & 0.0838 & 14.61 \\
             \midrule
             PointNet++ (w/o opt.) & 56.47 & 66.29 & 66.25 & 0.1568 & 18.53 \\
             PointNet++ (w/ opt.) & 64.33 & 60.51 & 59.53 & 0.1670 & 14.37 \\
            \bottomrule
        \end{tabular}%
    }%
\end{table}

As shown in \cref{tab:supp:abl_architecture}, we ablate the module for noise prediction. We compare the design of cross-attention and self-attention for processing the condition and input. Cross-attention indicates learning query from the input $\btau_t$ and learning key and value from the scene condition $\cS$. Self-attention indicates concatenating $\btau_t$ and scene features $\cS$ and learning with self-attention. Through our experiments, we find that with self-attention, the model learns better to capture the joint distribution of input and condition. This leads to a slightly lower diversity but better generation quality and success rate.

\begin{table}[ht!] 
    \centering
    \caption{\textbf{Ablation on different model architecture.}}
    \label{tab:supp:abl_architecture}
    \resizebox{\linewidth}{!}{%
        \begin{tabular}{cccccc}
            \toprule
            \multirow{2}[2]{*}{\textbf{Epsilon Model}} & \multicolumn{3}{c}{\textbf{Succ. Rate} (\%)$\uparrow$} & \multirow{2}[2]{*}{\textbf{Div.} (rad.)$\uparrow$} &\multirow{2}[2]{*}{\textbf{Coll.} (mm)$\downarrow$} \\
            \cmidrule{2-4}
             & $\sigma$ & $2\sigma$  & all &  \\
             \midrule
             CrossAttn. (w/o opt.) & 70.65 & 71.25 & 71.25 & 0.0718 & 17.34 \\
             CrossAttn. (w/ opt.) & 71.27 & 70.32 & 69.84 & 0.0838 & 14.61 \\
             \midrule
             SelfAttn. (w/o opt.) & 74.27 & 75.94 & 75.94 & 0.0535 & 16.49\\
             SelfAttn. (w/ opt.) & 72.01 & 71.56 & 71.09 & 0.0605 & 13.94 \\
            \bottomrule
        \end{tabular}%
    }%
\end{table}

\subsection{Diffusion Steps}

We study different diffusion steps $T$ in \cref{tab:supp:abl_steps_scale}, where we use PointNet++ as the scene encoder with cross-attention design. We report the success rate, diversity, and depth collision of sampling results in the test set under different diffusion steps, ranging from 30 to 1000. $T$ balance the diversity and success rate in dexterous grasp generation, where $T = 30$ leads to the best diversity of generated grasp pose and $T = 1000$ leads to the best \textit{all} success rate.

\begin{table*}[ht!]
    \centering
    \caption{\textbf{Ablation on diffusion steps and scale coefficient.}}
    \resizebox{0.75\linewidth}{!}{%
        \begin{tabular}{ccccccccc}
            \toprule
            \multirow{2}[2]{*}{\textbf{Time Steps}} & \multirow{2}[2]{*}{\textbf{Optimizer Scale}} & \multicolumn{4}{c}{\textbf{Succ. Rate} (\%)$\uparrow$} & \multirow{2}[2]{*}{\textbf{Diversity} (rad.)$\uparrow$} &\multirow{2}[2]{*}{\textbf{Depth Collision} (mm)$\downarrow$} \\
            \cmidrule{3-6}
             & & $\sigma$ & $2\sigma$ &$3\sigma$ & all & & \\
             \midrule
             30 & w/o & 0.00 & 60.01 & 50.94 & 48.13 & 0.3418 & 21.19 \\
             30 & 0.1 & 0.00 & 58.72 & 54.90 & 51.09 & 0.3415 & 19.96 \\
             30 & 0.5 & 0.00 & 64.24 & 51.63 & 47.81 & 0.3397 & 17.41 \\
             30 & 1.0 & 0.00 & 60.41 & 48.76 & 43.59 & 0.3393 & 16.05 \\
             \midrule
             100 & w/o & 0.00 & 66.62 & 60.12 & 58.91 & 0.2865 & 19.07 \\
             100 & 0.1 & 0.00 & 66.54 & 60.60 & 59.53 & 0.2836 & 17.55 \\
             100 & 0.5 & 0.00 & 61.23 & 56.71 & 53.75 & 0.2898 & 14.63 \\
             100 & 1.0 & 0.00 & 56.79 & 53.13 & 48.91 & 0.2920 & 14.53 \\
             \midrule
             500 & w/o & 75.00 & 67.50 & 67.34 & 67.34 & 0.1753 & 19.29 \\
             500 & 0.1 & 68.56 & 65.19 & 65.00 & 65.00 & 0.1733 & 17.68 \\
             500 & 0.5 & 62.83 & 60.25 & 58.94 & 58.75 & 0.1814 & 15.12 \\
             500 & 1.0 & 62.21 & 57.76 & 55.17 & 54.37 & 0.1872 & 14.36 \\
             \midrule
             1000 & w/o & 56.47 & 66.29 & 66.26 & 66.25 & 0.1568 & 18.53 \\
             1000 & 0.1 & 73.24 & 71.43 & 71.04 & 71.09 & 0.1572 & 16.88 \\
             1000 & 0.5 & 70.18 & 65.99 & 65.55 & 65.62 & 0.1611 & 14.37 \\
             1000 & 1.0 & 64.33 & 60.51 & 59.61 & 59.53 & 0.1670 & 14.37 \\
            \bottomrule
        \end{tabular}%
    }%
    \label{tab:supp:abl_steps_scale}
\end{table*}

\subsection{Scale Coefficient}

Among different time steps $T$, we ablate scale coefficient $\lambda$ of the optimization guidance in dexterous grasp generation in \cref{tab:supp:abl_steps_scale}, ranging from $0.0$ (denoted as w/o in the table) to $1.0$. Through extensive experiments, we observe that, in general, the $\alpha$ trade off the depth collision and grasp success rate. A larger $\alpha$ value leads to fewer collisions and draws the grasp pose away from the object simultaneously, which losses the grasp stability and lowers the success rate.

We also ablate the scale coefficient of the planner in path planning for 3D scene navigation, as shown in \cref{tab:supp:abl_planning_horizon_scale}. Too small or too large scale coefficients both lead to a performance drop. It is due to that a small value cannot provide sufficient guidance. In contrast, a large value diminishes trajectory diversity with strong guidance, preventing it from escaping obstacles and dead-ends.

\begin{table}[ht!]
    \centering
    \caption{\textbf{Ablation on different inpainting horizons and scale coefficients of the planning guidance.}}
    \label{tab:supp:abl_planning_horizon_scale}
    \resizebox{\linewidth}{!}{%
        \begin{tabular}{cccc}%
            \toprule
            \textbf{Fixed Frames} & \textbf{Planner Scale} & \textbf{Succ. Rate}(\%)$\uparrow$ & \textbf{Planning Steps}$\downarrow$ \\
            \midrule
            1  & 0.2 & 31.25 & 135.14 \\
            \midrule
            7  & 0.2 & 65.50 & 104.30 \\
            \midrule
            15 & 0.2 & 73.75 & 90.38 \\
            \midrule
            23 & 0.2 & 73.25 & 87.49 \\
            \midrule
            \multirow{4}{*}{31} & 0.1 & 53.50 & 106.23 \\
                                & 0.2 & 62.37 & 97.02  \\
                                & 0.3 & 56.81 & 101.54 \\
                                & 0.4 & 50.94 & 105.11 \\
            \bottomrule
        \end{tabular}%
    }%
\end{table}

\subsection{Fixed Frames for Planning}

Since we formulate the planning algorithm as inpainting, we also ablate the number of the fixed frame in it. In path planning for 3D scene navigation, we train the \model with a trajectory length of 32. Therefore, we compare the settings of fixing the first 1, 7, 15, 23, and 31 frames for inpainting during the denoising process. The results in \cref{tab:supp:abl_planning_horizon_scale} show that the model achieves the best performance while fixing the first 15 frames.

\subsection{Planning Objectives}

To explore the influence of different planning objectives, we design the following four planning objectives and compare them with \cref{eq:obj_plan}.

\begin{itemize}
    \item We compute the L1 distance between the last frame of the denoised trajectory and the target position, \ie,
\begin{equation}
    \obj_{p}=-\|\cG - \tau^t_L\|_1.
\end{equation}
    \item We summarize the L1 distance between all frames of the denoised trajectory and the target position, \ie,
\begin{equation}
    \obj_{p}=-\sum_{i=1}^{L}\|\cG - \tau^t_i\|_1.
\end{equation}
    \item Similar to \cref{eq:obj_plan}, we only consider the last frame of the denoised trajectory, \ie,
\begin{equation}
    \obj_{p}=\exp\left(\frac{1}{\|\cG - \tau^t_L\|_1}\right).
\end{equation}
    \item We compute the L1 distance between the target position and the frame closest to the target, \ie,
\begin{equation}
    \obj_{p}=-\min\limits_{i}\|\cG - \tau^t_i\|_1.
\end{equation}
\end{itemize}

The planning results in \cref{tab:supp:abl_planning_objective} indicate that encouraging all frames of the denoised trajectory to reach the target position surpasses considering only one frame. Besides, directly using $L1$ distance tends to achieve a better performance than additionally applying the exponential function.

\begin{table}[ht!]
    \centering
    \caption{\textbf{Ablation on different planning objectives.}}
    \label{tab:supp:abl_planning_objective}
    \resizebox{\linewidth}{!}{%
        \begin{tabular}{ccc}%
            \toprule
            \textbf{Objective} & \textbf{Succ. Rate}(\%)$\uparrow$ & \textbf{Planning Steps}$\downarrow$ \\
            \midrule
            $\obj_{p}=-\|\cG - \tau^t_L\|_1$ & 57.06 & 116.22 \\
            $\obj_{p}=-\sum_{i=1}^{L}\|\cG - \tau^t_i\|_1$ & 75.69 & 88.02 \\
            $\obj_{p}=\exp\left(\frac{1}{\|\cG - \tau^t_L\|_1}\right)$ & 34.31 & 131.74 \\
            $\obj_{p}=\sum_{i=1}^{L}\exp\left(\frac{1}{\|\cG - \tau^t_i\|_1}\right)$ & 73.75 & 90.38 \\
            $\obj_{p}=-\min\limits_{i}\|\cG - \tau^t_i\|_1$ & 56.00 & 109.02 \\
            \bottomrule
        \end{tabular}%
    }%
\end{table}

\section{Trainable Optimization and Planning}

\begin{figure}[t!]
    \centering
    \begin{subfigure}{0.5\linewidth}
        \includegraphics[width=\linewidth]{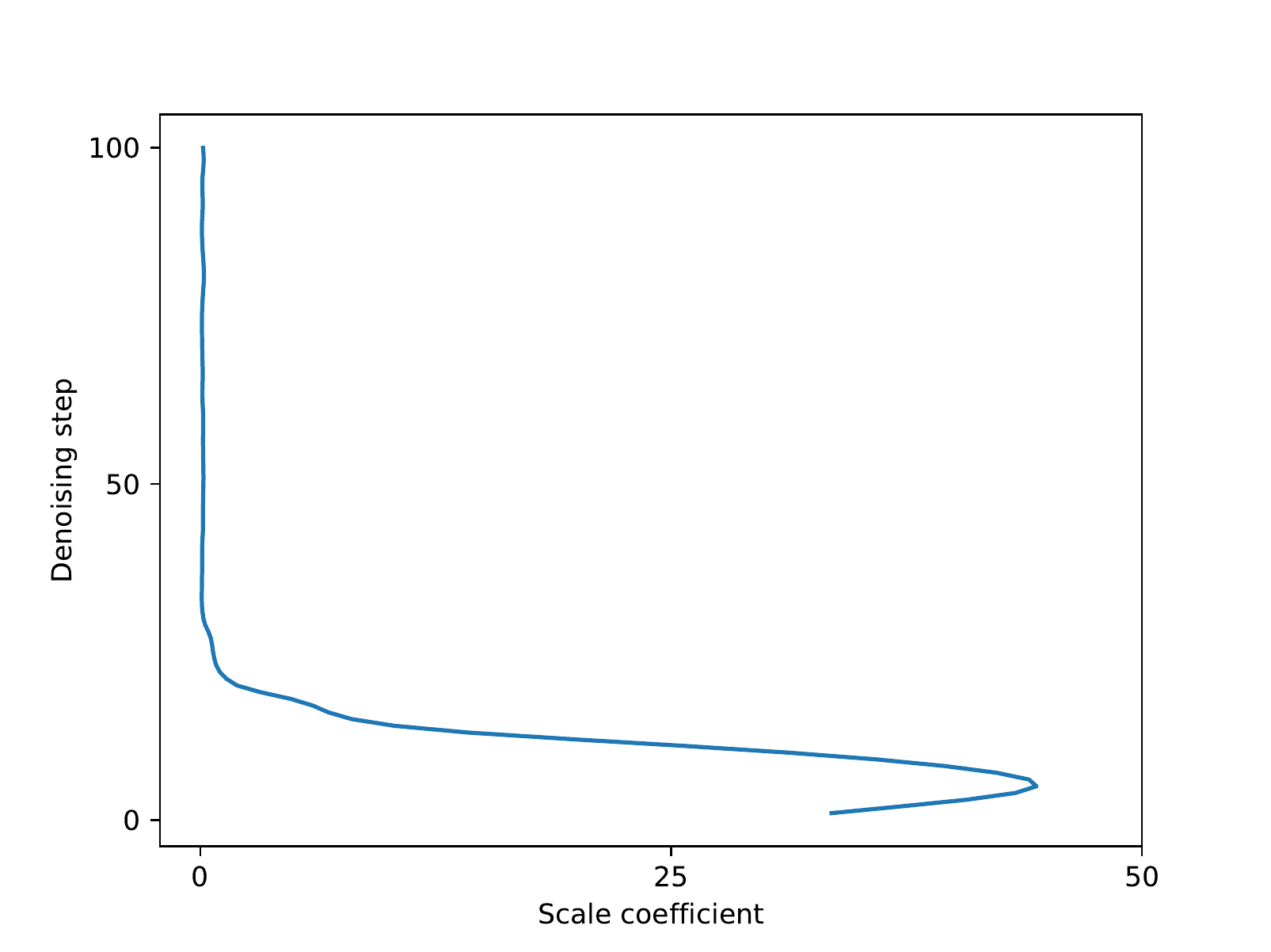}
        \caption{pose generation}
        \label{fig:supp:pose_generation}
    \end{subfigure}%
    \begin{subfigure}{0.5\linewidth}
	\includegraphics[width=\linewidth]{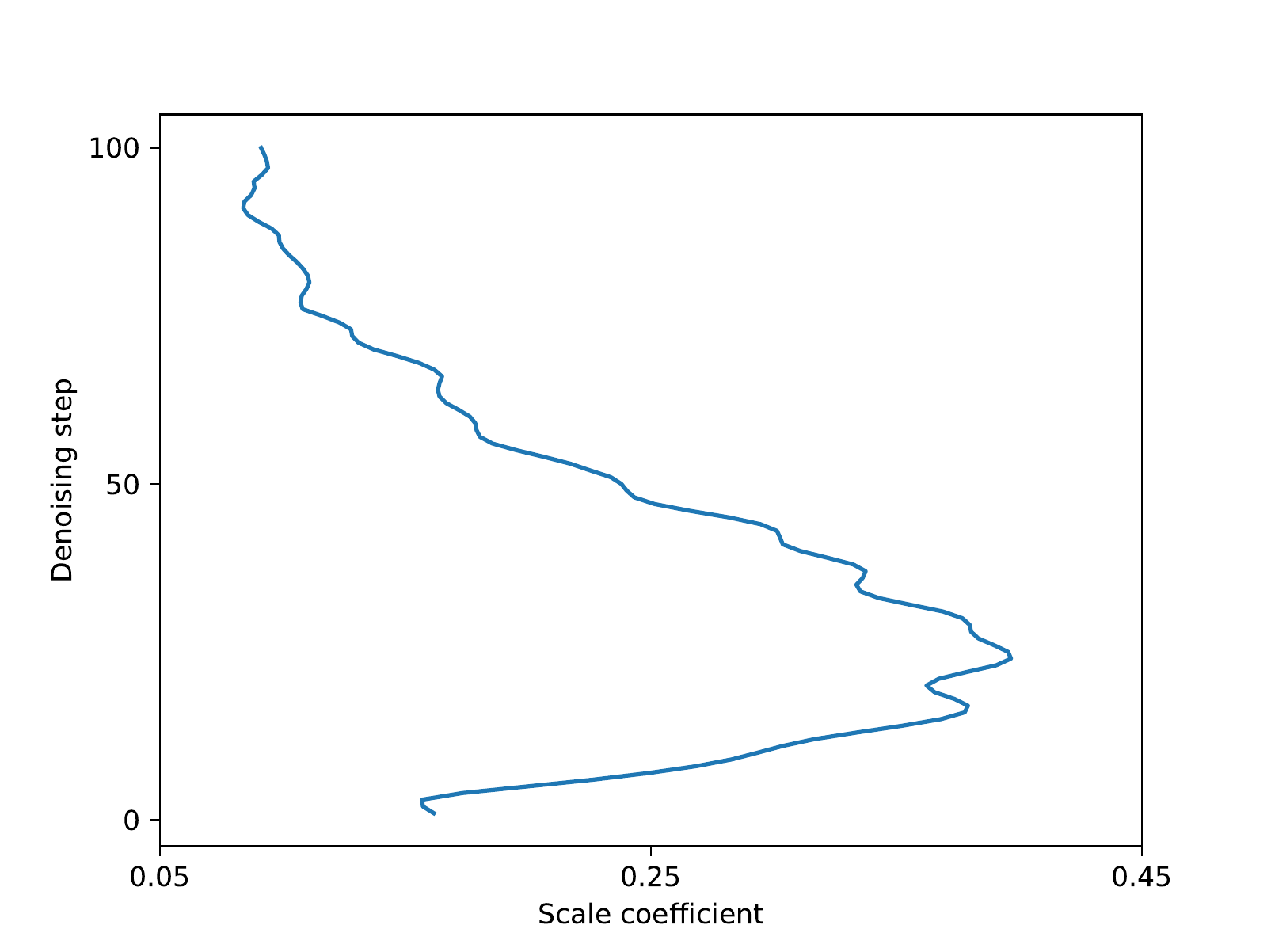}
        \caption{path planning}
        \label{fig:supp:path_planning}
    \end{subfigure}%
    \caption{\textbf{Trainable scaling factor varying with the denoising step.}}
    \label{fig:supp:trainable_scale}
\end{figure}

As shown in \cref{alg:diffuser:training}, we can optionally train the optimization and planning process with observed trajectories. To verify its efficacy, we optimize the trainable scaling factor $\lambda$ of the optimization guidance in pose generation and path planning tasks. Specifically, we use a small MLP model to map the timestep embedding of each step into a scalar, \ie, the scaling factor. During training, we only optimize the MLP while fixing the pre-trained diffusion model. We plot the learned scaling factor varying with the denoising step from 100 to 1, as shown in \cref{fig:supp:trainable_scale}. We observe that the scaling factor of the denoising process at the beginning is much smaller than at the end. We speculate that the target signal at the beginning of the denoising process is mostly noise so a large scaling factor cannot optimize it properly. The scaling factor decrease in the last several steps may be because this can alleviate excessive guidance and balance the guidance from other modules, such as the planner.

\section{More Qualitative Results}\label{app:sec:quali}

\paragraph{Pose Generation in 3D Scenes}

\begin{figure*}[t!]
    \centering
    \includegraphics[width=\linewidth]{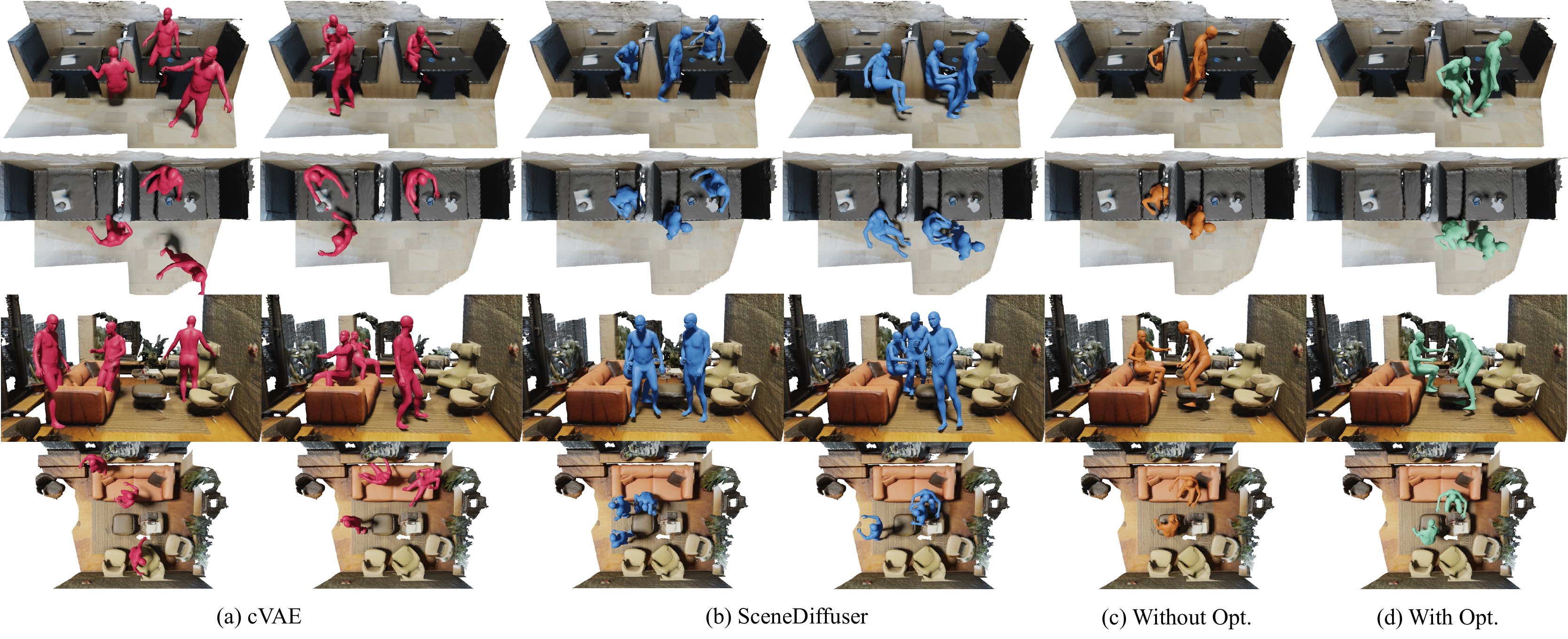}
    \caption{\textbf{More qualitative results of pose generation in 3D scenes.}}
    \label{fig:supp:pose_gen_qual_supp}
\end{figure*}

We show more qualitative results in  \cref{fig:supp:pose_gen_qual_supp}.

\paragraph{Motion Generation in 3D Scenes}

\begin{figure*}[t!]
    \centering
    \includegraphics[width=\linewidth]{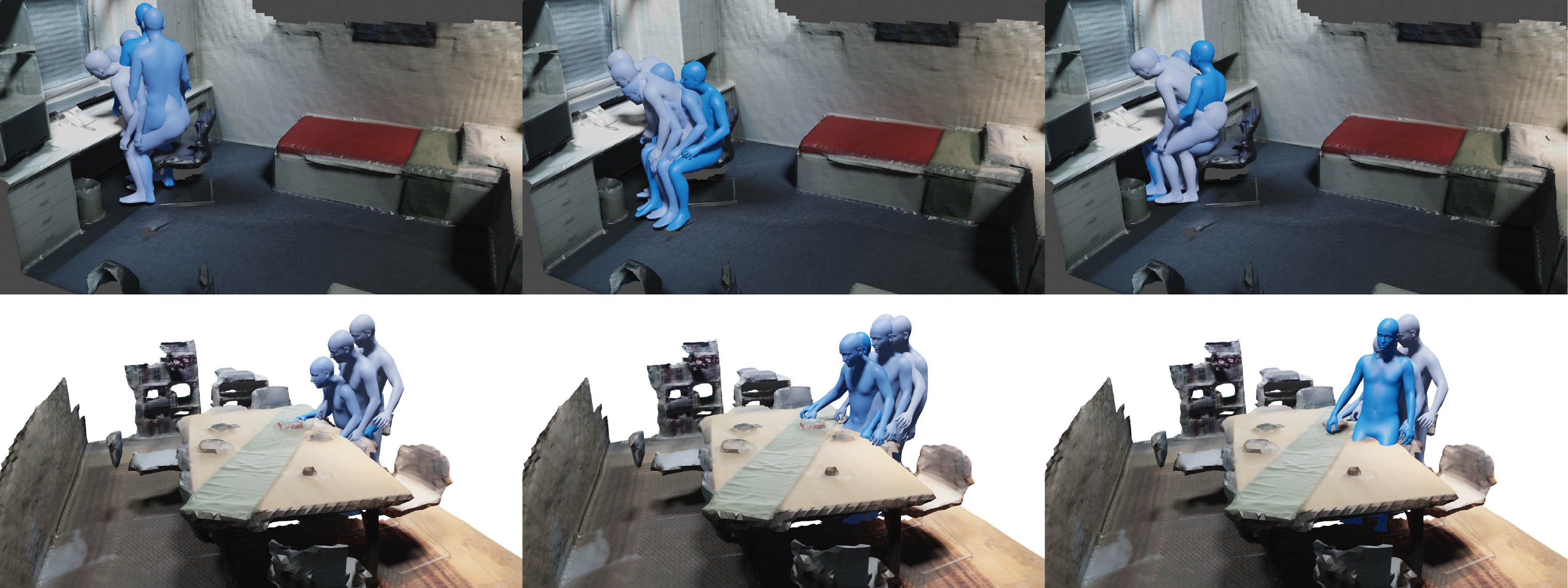}
    \caption{\textbf{More qualitative results of motion generation in 3D scenes.}}
    \label{fig:supp:motion_gen_qual_supp}
\end{figure*}

We provide more sampled human motions from the same start pose in other scenes, as shown in \cref{fig:supp:motion_gen_qual_supp}. Please refer to the supplemental demo video for better visualization with rendered animations.

\paragraph{Path Planning for 3D Scene Navigation}

\begin{figure*}[t!]
    \centering
    \includegraphics[width=\linewidth]{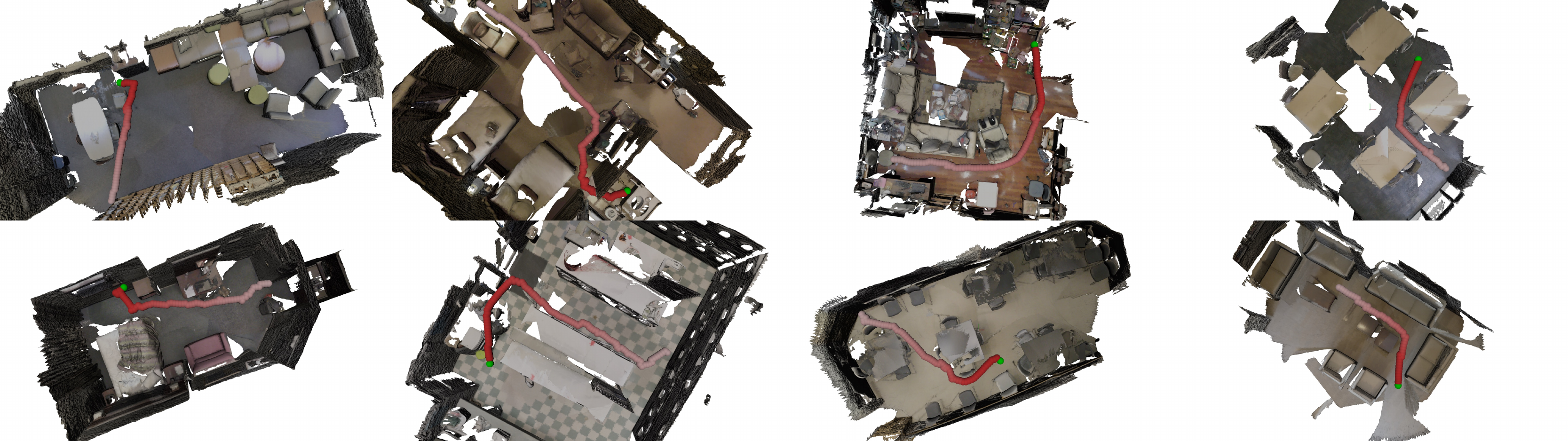}
    \caption{\textbf{Qualitative results of path planning for 3D scene navigation.} The red balls represent the planning result, starting with the lightest red ball and ending with the darkest red ball. The green ball indicates the target position.}
    \label{fig:supp:path_planning_qual_supp}
\end{figure*}

\cref{fig:supp:path_planning_qual_supp} shows some qualitative results of path planning for 3D scene navigation.

\paragraph{Dexterous Grasp Generation for 3D Objects}

\begin{figure*}
    \centering
    \begin{subfigure}{0.11\linewidth}\hfill
    \includegraphics[width=1.0\linewidth]{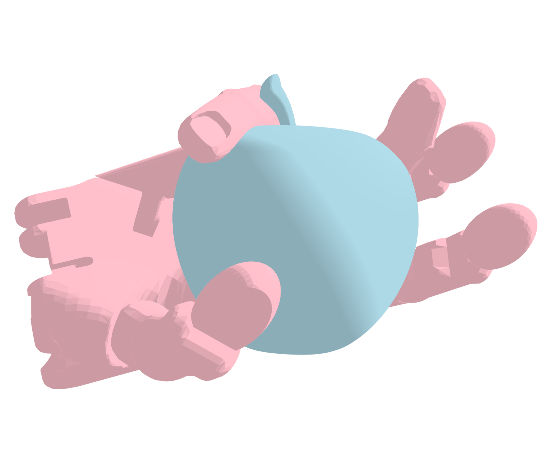} 
    \hfill\end{subfigure} \hfill
    \begin{subfigure}{0.11\linewidth}\hfill
    \includegraphics[width=1.0\linewidth]{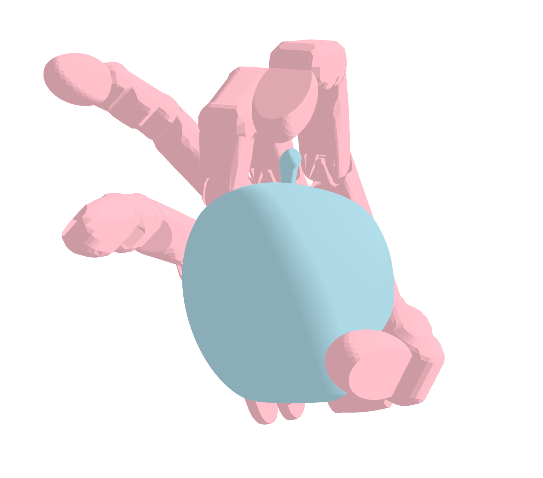}
    \hfill\end{subfigure} \hfill
    \begin{subfigure}{0.11\linewidth}\hfill
    \includegraphics[width=1.0\linewidth]{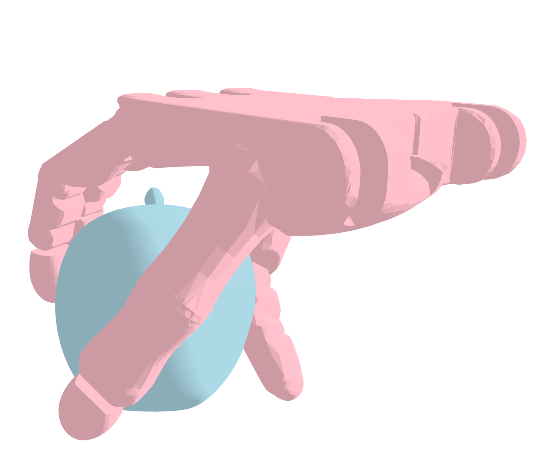}
    \hfill\end{subfigure} \hfill
    \begin{subfigure}{0.11\linewidth}\hfill
    \includegraphics[width=1.0\linewidth]{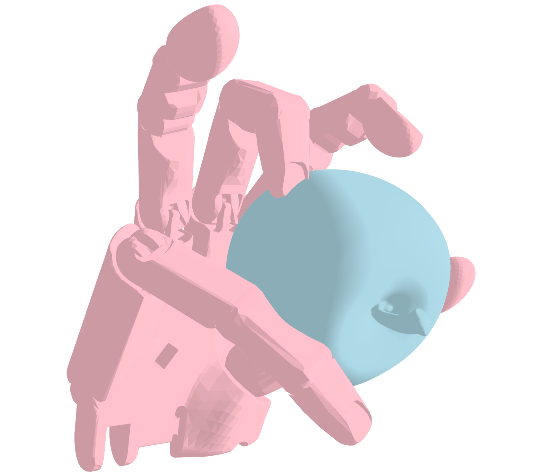}
    \hfill\end{subfigure} \hfill
    \begin{subfigure}{0.11\linewidth}\hfill
    \includegraphics[width=1.0\linewidth]{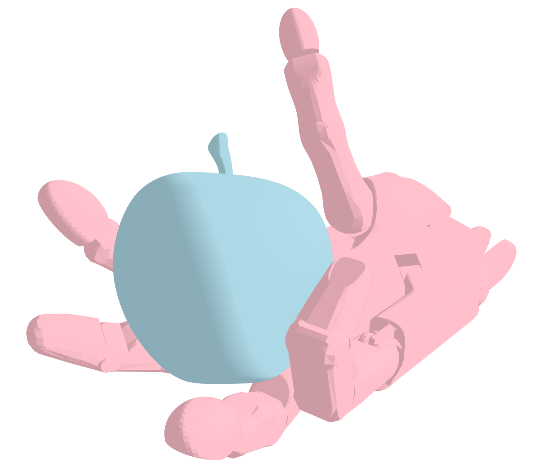}
    \hfill\end{subfigure} \hfill
    \begin{subfigure}{0.11\linewidth}\hfill
    \includegraphics[width=1.0\linewidth]{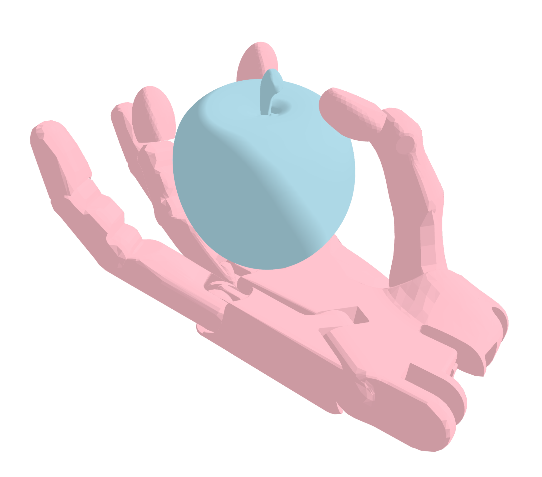}
    \hfill\end{subfigure} \hfill
    \begin{subfigure}{0.11\linewidth}\hfill
    \includegraphics[width=1.0\linewidth]{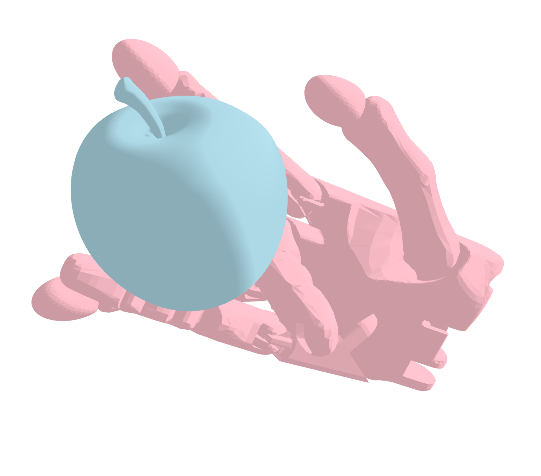}
    \hfill\end{subfigure} \hfill
    \begin{subfigure}{0.11\linewidth}\hfill
    \includegraphics[width=1.0\linewidth]{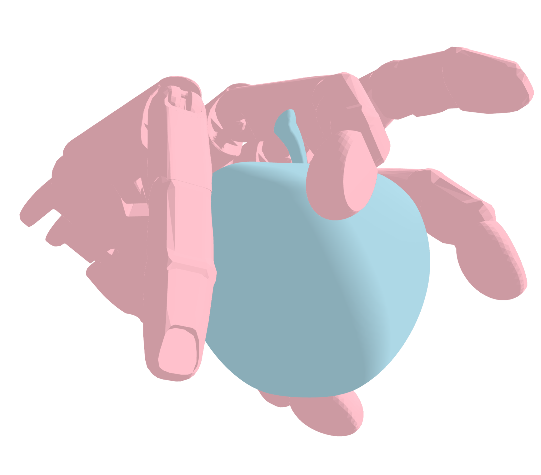}
    \hfill\end{subfigure} 
    \\
    
    \begin{subfigure}{0.11\linewidth}\hfill
    \includegraphics[width=1.0\linewidth]{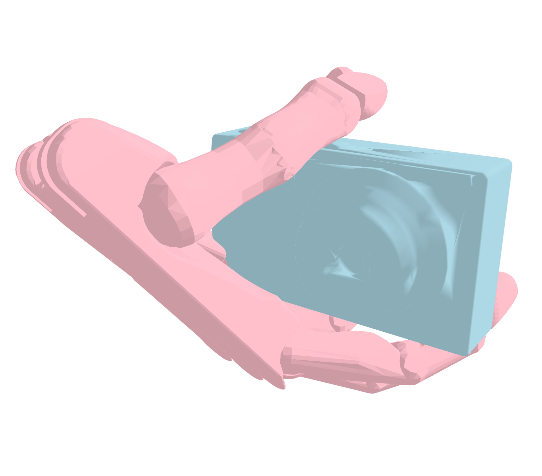} 
    \hfill\end{subfigure} \hfill
    \begin{subfigure}{0.11\linewidth}\hfill
    \includegraphics[width=1.0\linewidth]{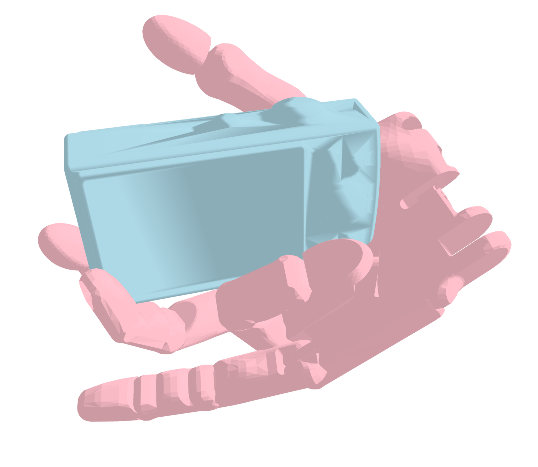} 
    \hfill\end{subfigure} \hfill
    \begin{subfigure}{0.11\linewidth}\hfill
    \includegraphics[width=1.0\linewidth]{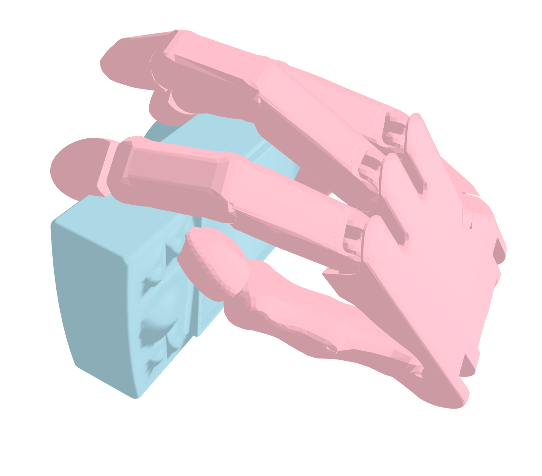} 
    \hfill\end{subfigure} \hfill
    \begin{subfigure}{0.11\linewidth}\hfill
    \includegraphics[width=1.0\linewidth]{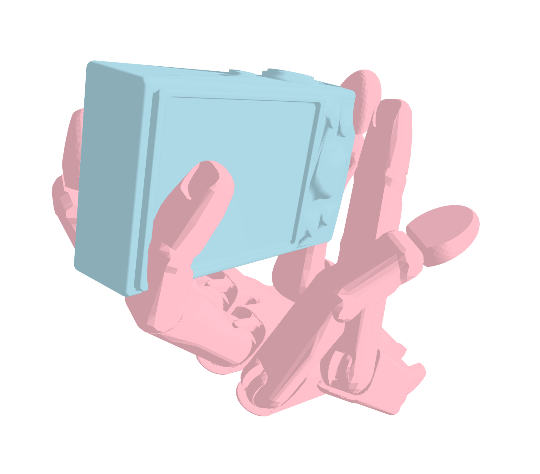} 
    \hfill\end{subfigure} \hfill
    \begin{subfigure}{0.11\linewidth}\hfill
    \includegraphics[width=1.0\linewidth]{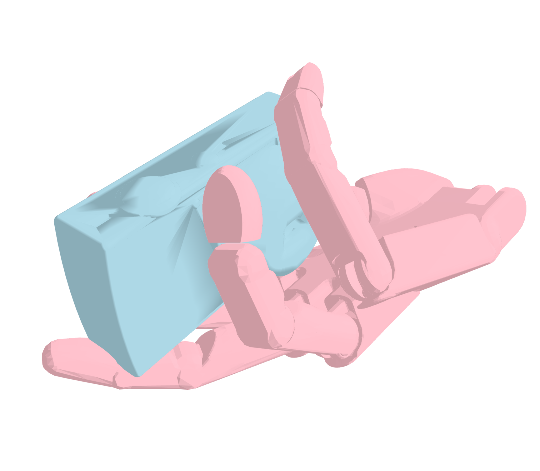} 
    \hfill\end{subfigure} \hfill
    \begin{subfigure}{0.11\linewidth}\hfill
    \includegraphics[width=1.0\linewidth]{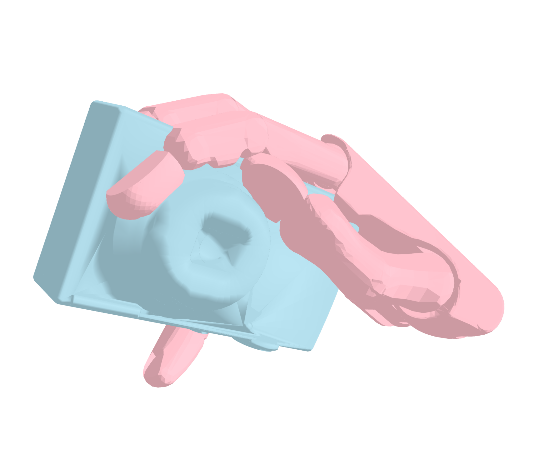} 
    \hfill\end{subfigure} \hfill
    \begin{subfigure}{0.11\linewidth}\hfill
    \includegraphics[width=1.0\linewidth]{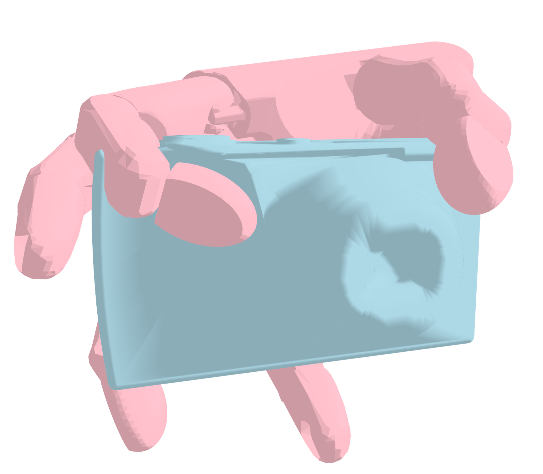} 
    \hfill\end{subfigure} \hfill
    \begin{subfigure}{0.11\linewidth}\hfill
    \includegraphics[width=1.0\linewidth]{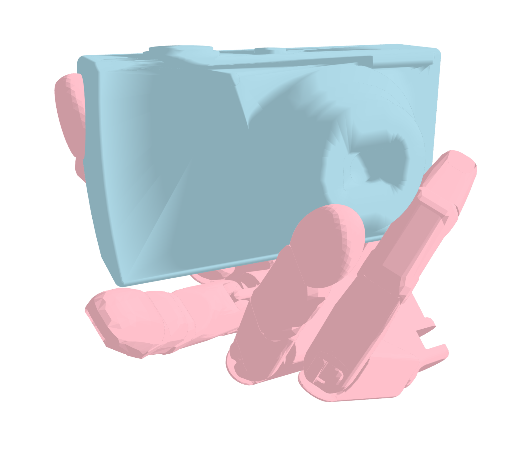} 
    \hfill\end{subfigure} 
    \\

    \begin{subfigure}{0.11\linewidth}\hfill
    \includegraphics[width=1.0\linewidth]{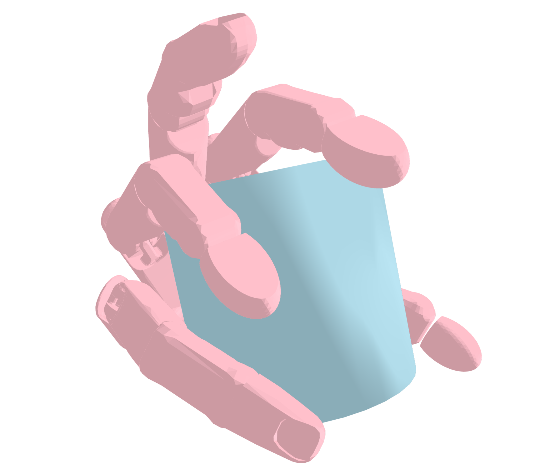} 
    \hfill\end{subfigure} \hfill
    \begin{subfigure}{0.11\linewidth}\hfill
    \includegraphics[width=1.0\linewidth]{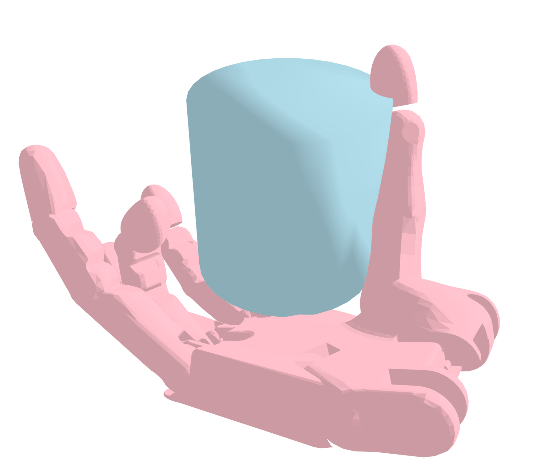}
    \hfill\end{subfigure} \hfill
    \begin{subfigure}{0.11\linewidth}\hfill
    \includegraphics[width=1.0\linewidth]{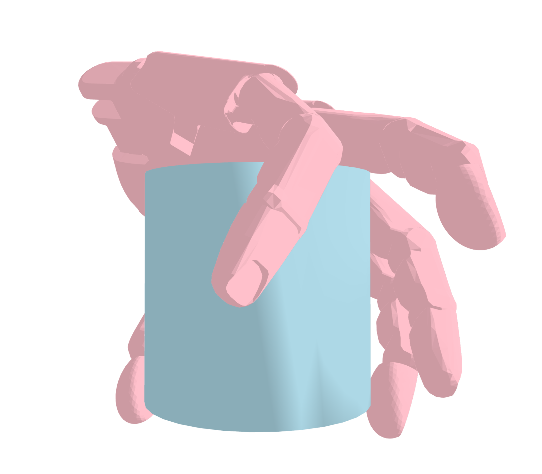} 
    \hfill\end{subfigure} \hfill
    \begin{subfigure}{0.11\linewidth}\hfill
    \includegraphics[width=1.0\linewidth]{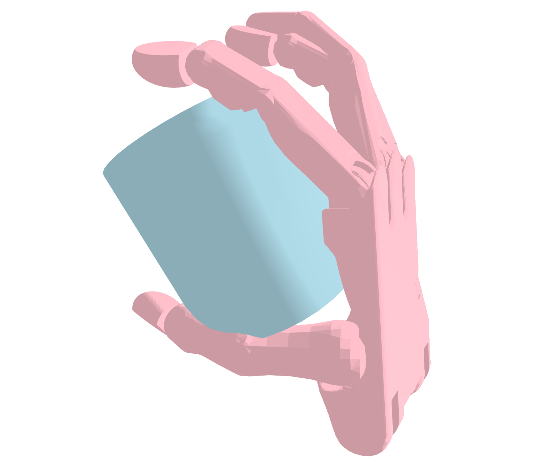} 
    \hfill\end{subfigure} \hfill
    \begin{subfigure}{0.11\linewidth}\hfill
    \includegraphics[width=1.0\linewidth]{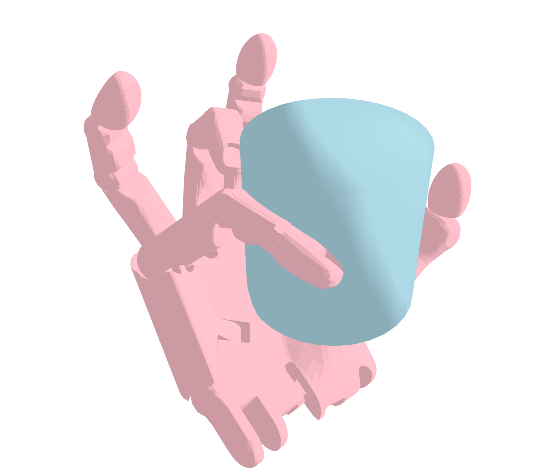} 
    \hfill\end{subfigure} \hfill
    \begin{subfigure}{0.11\linewidth}\hfill
    \includegraphics[width=1.0\linewidth]{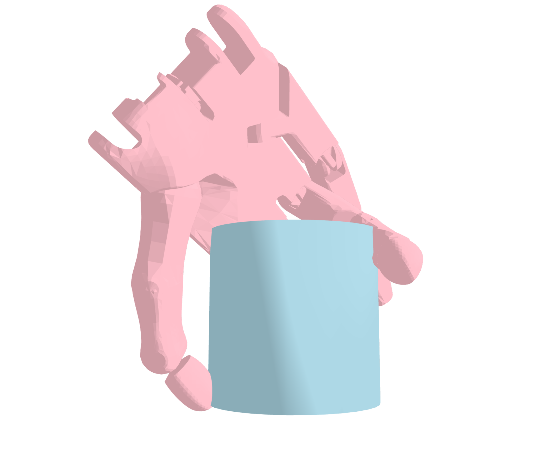} 
    \hfill\end{subfigure} \hfill
    \begin{subfigure}{0.11\linewidth}\hfill
    \includegraphics[width=1.0\linewidth]{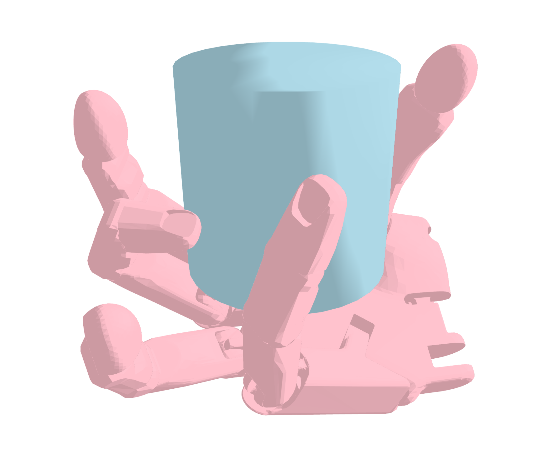} 
    \hfill\end{subfigure} \hfill
    \begin{subfigure}{0.11\linewidth}\hfill
    \includegraphics[width=1.0\linewidth]{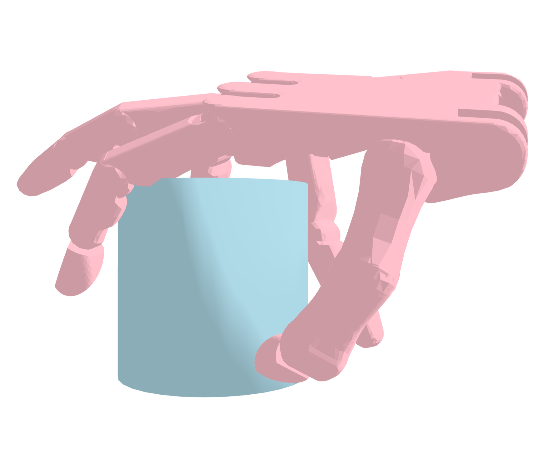} 
    \hfill\end{subfigure} 
    \\

    \begin{subfigure}{0.11\linewidth}\hfill
    \includegraphics[width=1.0\linewidth]{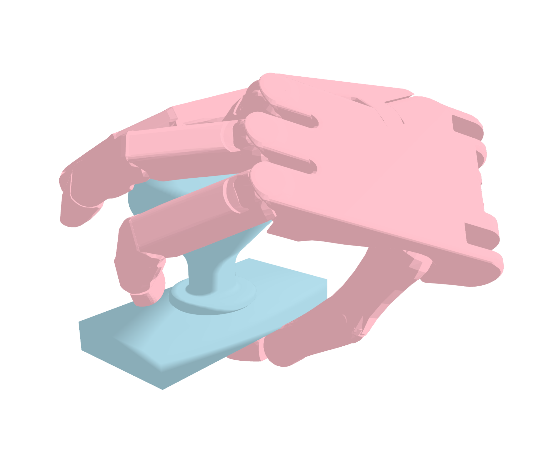}
    \hfill\end{subfigure} \hfill
    \begin{subfigure}{0.11\linewidth}\hfill
    \includegraphics[width=1.0\linewidth]{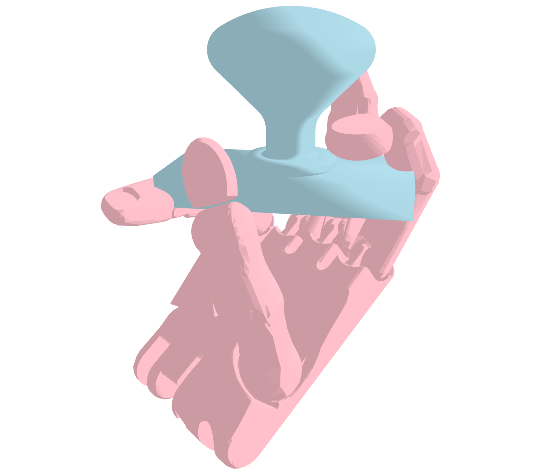}
    \hfill\end{subfigure} \hfill
    \begin{subfigure}{0.11\linewidth}\hfill
    \includegraphics[width=1.0\linewidth]{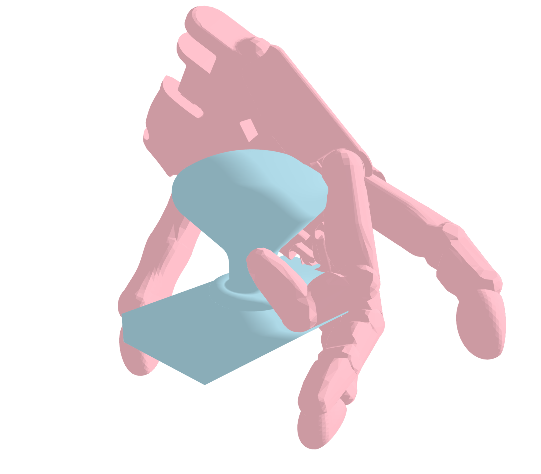}
    \hfill\end{subfigure} \hfill
    \begin{subfigure}{0.11\linewidth}\hfill
    \includegraphics[width=1.0\linewidth]{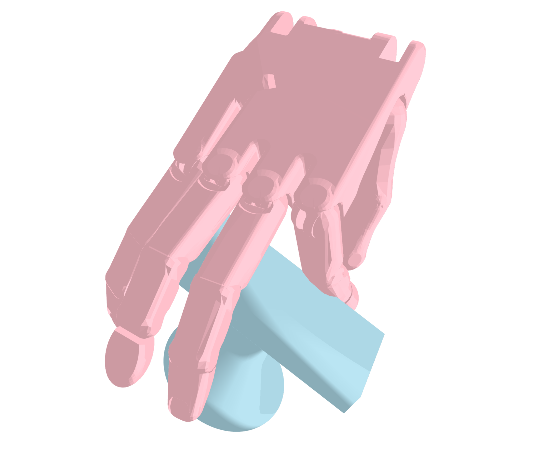}
    \hfill\end{subfigure} \hfill
    \begin{subfigure}{0.11\linewidth}\hfill
    \includegraphics[width=1.0\linewidth]{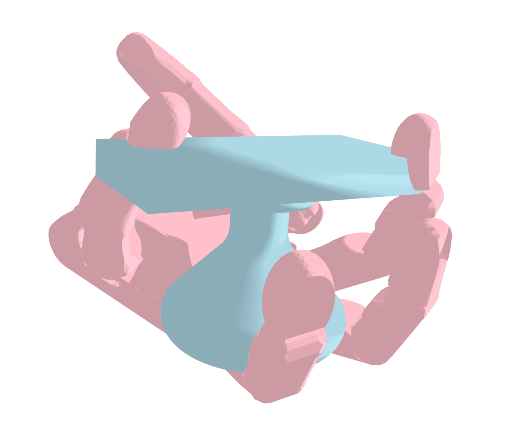}
    \hfill\end{subfigure} \hfill
    \begin{subfigure}{0.11\linewidth}\hfill
    \includegraphics[width=1.0\linewidth]{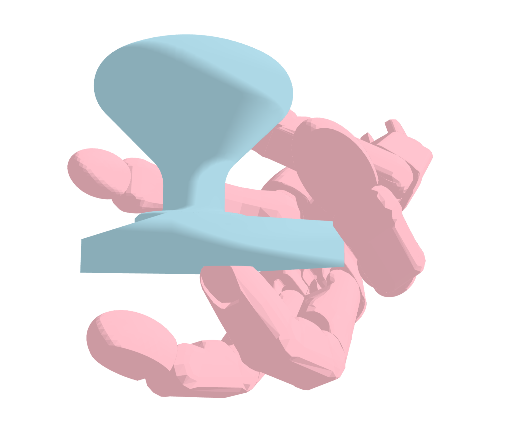}
    \hfill\end{subfigure} \hfill
    \begin{subfigure}{0.11\linewidth}\hfill
    \includegraphics[width=1.0\linewidth]{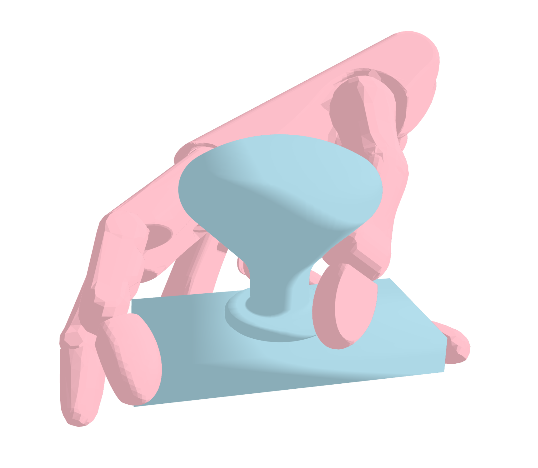}
    \hfill\end{subfigure} \hfill
    \begin{subfigure}{0.11\linewidth}\hfill
    \includegraphics[width=1.0\linewidth]{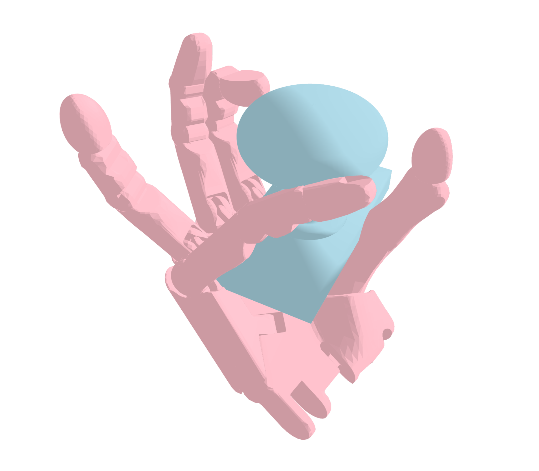}
    \hfill\end{subfigure} 
    \\

    \begin{subfigure}{0.11\linewidth}\hfill
    \includegraphics[width=1.0\linewidth]{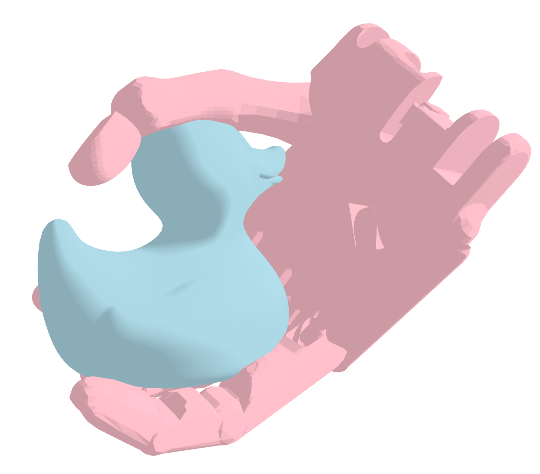}
    \hfill\end{subfigure} \hfill
    \begin{subfigure}{0.11\linewidth}\hfill
    \includegraphics[width=1.0\linewidth]{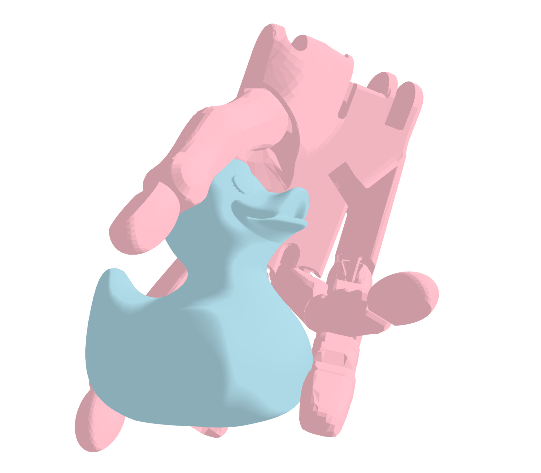}
    \hfill\end{subfigure} \hfill
    \begin{subfigure}{0.11\linewidth}\hfill
    \includegraphics[width=1.0\linewidth]{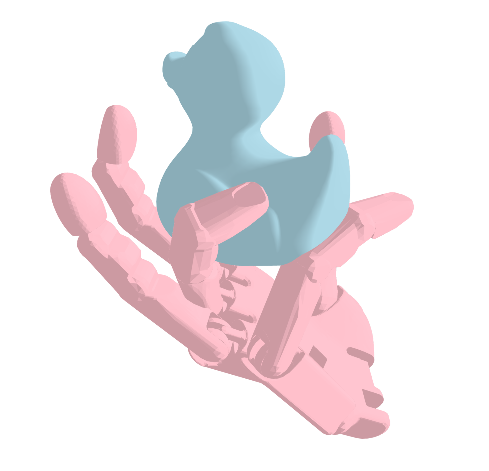}
    \hfill\end{subfigure} \hfill
    \begin{subfigure}{0.11\linewidth}\hfill
    \includegraphics[width=1.0\linewidth]{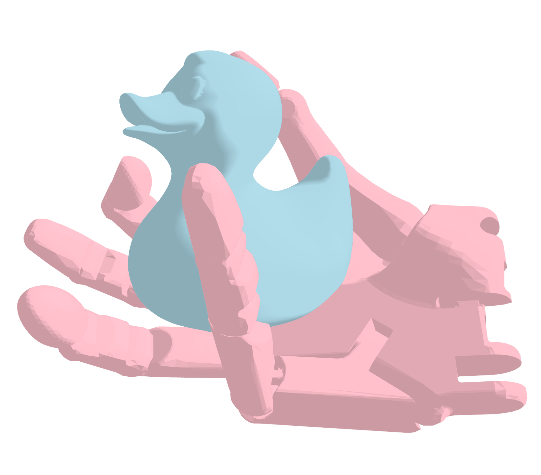}
    \hfill\end{subfigure} \hfill
    \begin{subfigure}{0.11\linewidth}\hfill
    \includegraphics[width=1.0\linewidth]{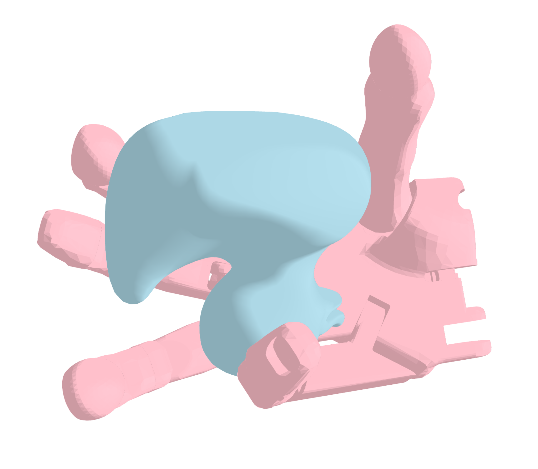}
    \hfill\end{subfigure} \hfill
    \begin{subfigure}{0.11\linewidth}\hfill
    \includegraphics[width=1.0\linewidth]{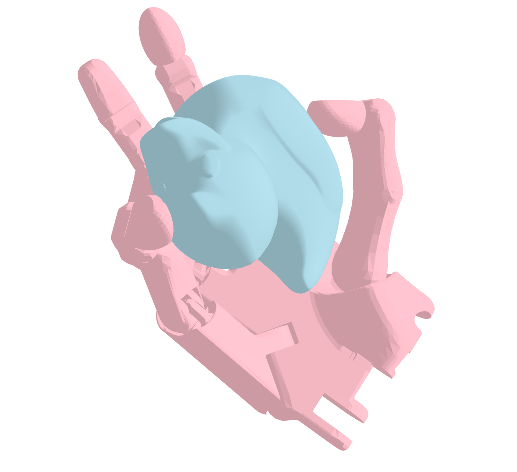}
    \hfill\end{subfigure} \hfill
    \begin{subfigure}{0.11\linewidth}\hfill
    \includegraphics[width=1.0\linewidth]{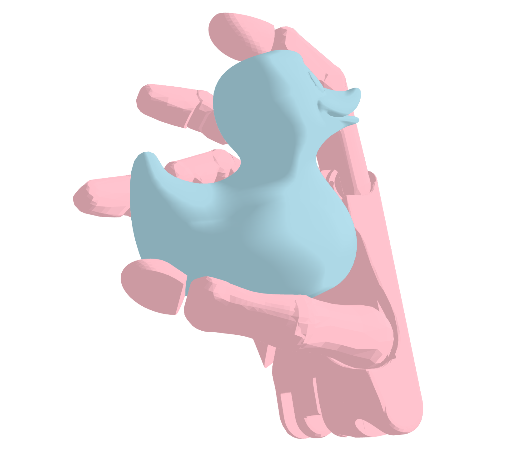}
    \hfill\end{subfigure} \hfill
    \begin{subfigure}{0.11\linewidth}\hfill
    \includegraphics[width=1.0\linewidth]{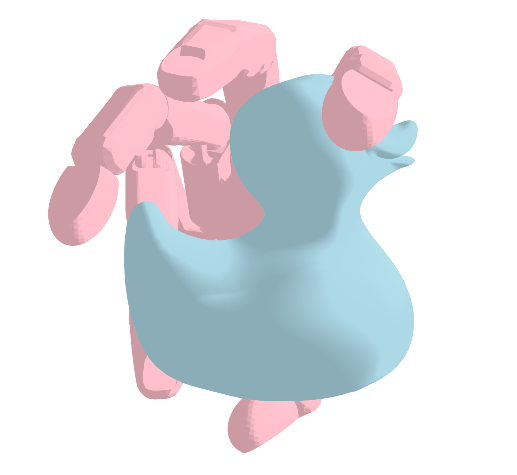}
    \hfill\end{subfigure} 
    \\

    \begin{subfigure}{0.11\linewidth}\hfill
    \includegraphics[width=1.0\linewidth]{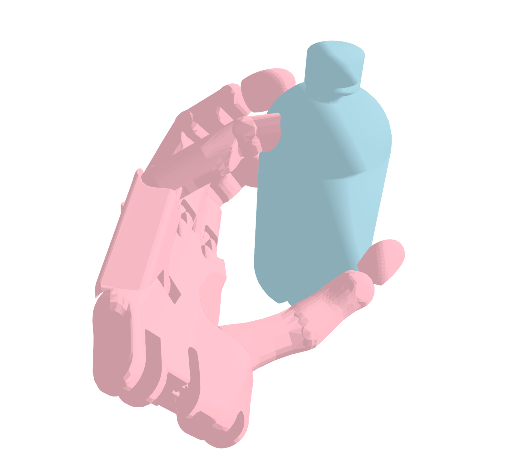}
    \hfill\end{subfigure} \hfill
    \begin{subfigure}{0.11\linewidth}\hfill
    \includegraphics[width=1.0\linewidth]{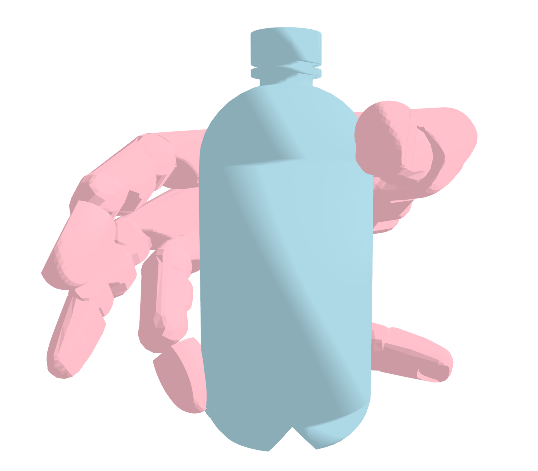}
    \hfill\end{subfigure} \hfill
    \begin{subfigure}{0.11\linewidth}\hfill
    \includegraphics[width=1.0\linewidth]{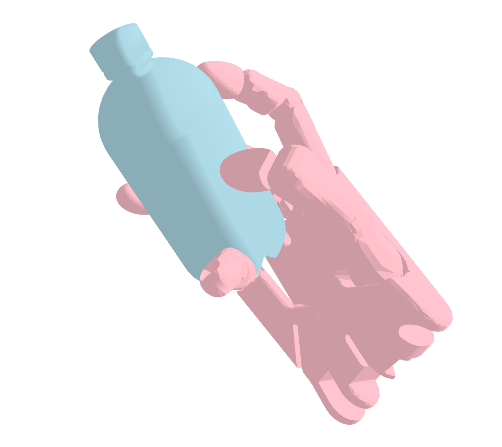}
    \hfill\end{subfigure} \hfill
    \begin{subfigure}{0.11\linewidth}\hfill
    \includegraphics[width=1.0\linewidth]{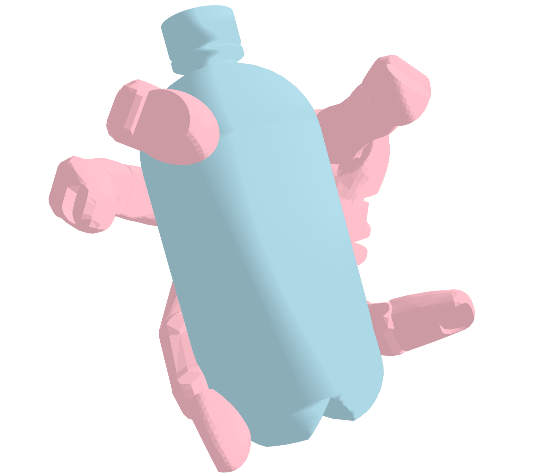}
    \hfill\end{subfigure} \hfill
    \begin{subfigure}{0.11\linewidth}\hfill
    \includegraphics[width=1.0\linewidth]{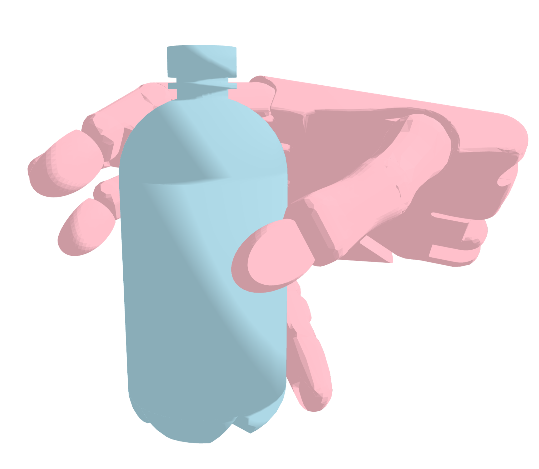}
    \hfill\end{subfigure} \hfill
    \begin{subfigure}{0.11\linewidth}\hfill
    \includegraphics[width=1.0\linewidth]{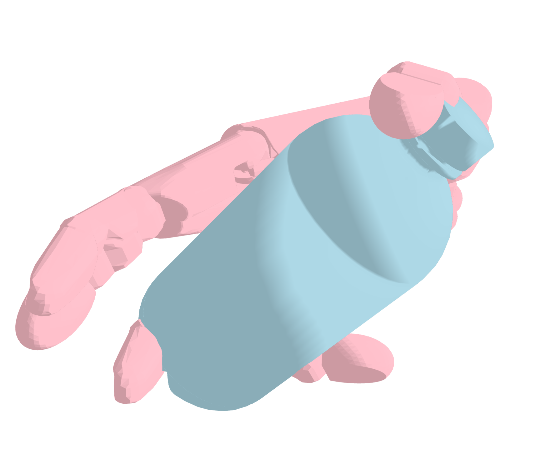}
    \hfill\end{subfigure} \hfill
    \begin{subfigure}{0.11\linewidth}\hfill
    \includegraphics[width=1.0\linewidth]{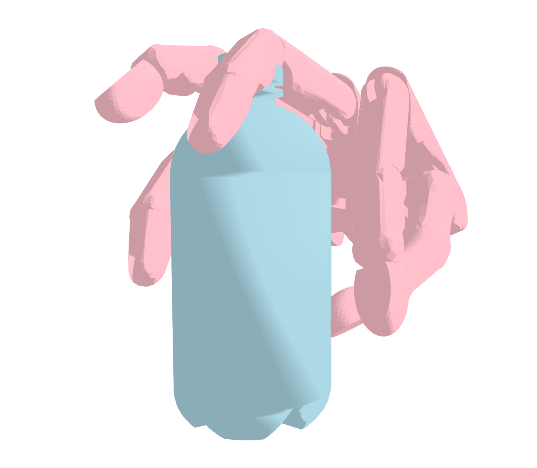}
    \hfill\end{subfigure} \hfill
    \begin{subfigure}{0.11\linewidth}\hfill
    \includegraphics[width=1.0\linewidth]{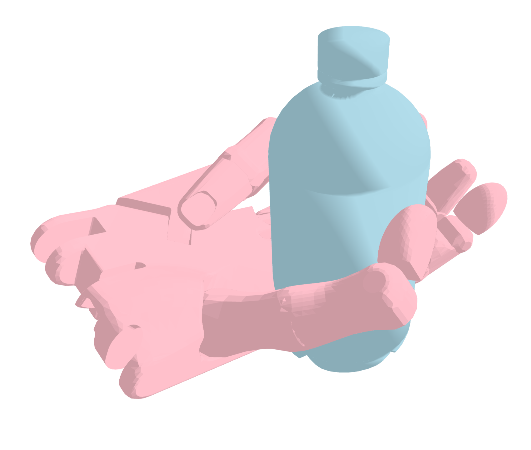}
    \hfill\end{subfigure} 
    \\

    \begin{subfigure}{0.11\linewidth}\hfill
    \includegraphics[width=1.0\linewidth]{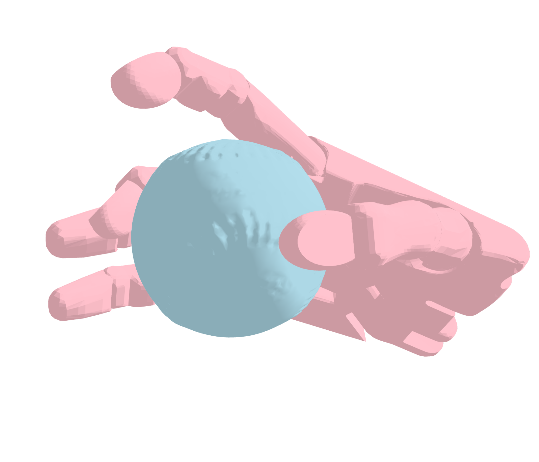}
    \hfill\end{subfigure} \hfill
    \begin{subfigure}{0.11\linewidth}\hfill
    \includegraphics[width=1.0\linewidth]{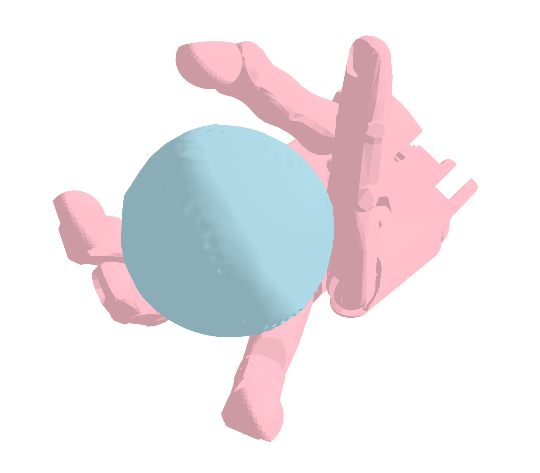}
    \hfill\end{subfigure} \hfill
    \begin{subfigure}{0.11\linewidth}\hfill
    \includegraphics[width=1.0\linewidth]{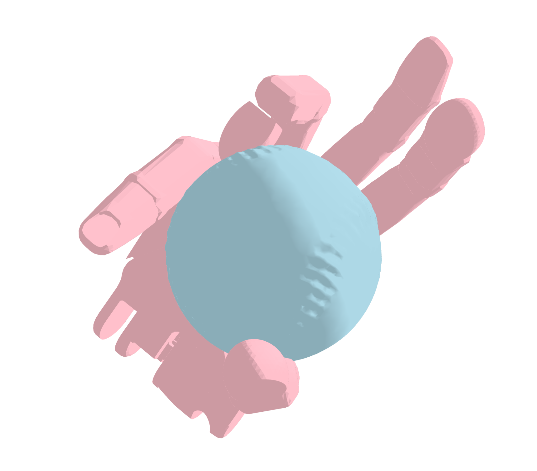}
    \hfill\end{subfigure} \hfill
    \begin{subfigure}{0.11\linewidth}\hfill
    \includegraphics[width=1.0\linewidth]{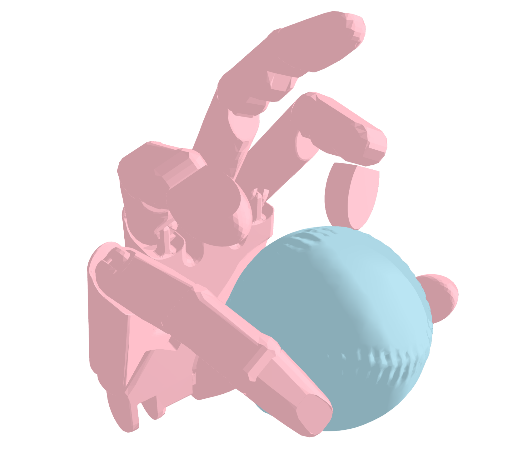}
    \hfill\end{subfigure} \hfill
    \begin{subfigure}{0.11\linewidth}\hfill
    \includegraphics[width=1.0\linewidth]{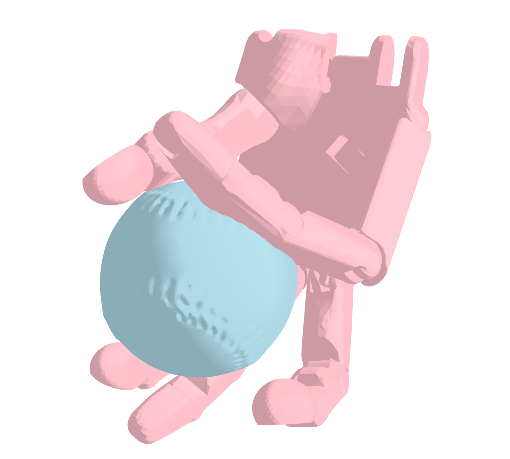}
    \hfill\end{subfigure} \hfill
    \begin{subfigure}{0.11\linewidth}\hfill
    \includegraphics[width=1.0\linewidth]{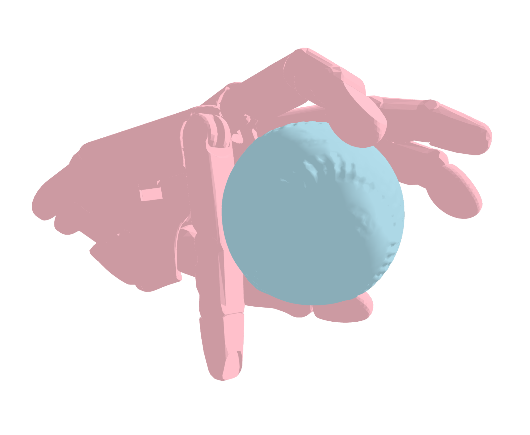}
    \hfill\end{subfigure} \hfill
    \begin{subfigure}{0.11\linewidth}\hfill
    \includegraphics[width=1.0\linewidth]{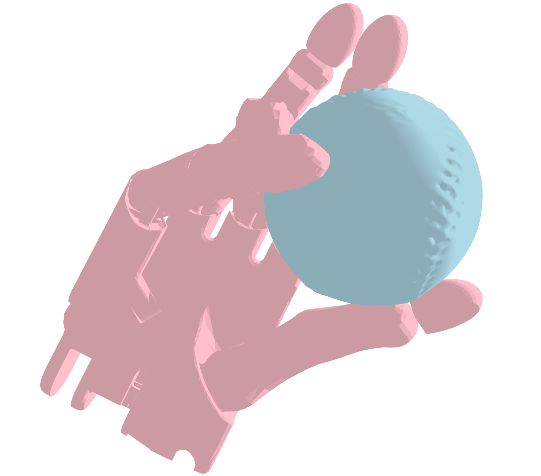}
    \hfill\end{subfigure} \hfill
    \begin{subfigure}{0.11\linewidth}\hfill
    \includegraphics[width=1.0\linewidth]{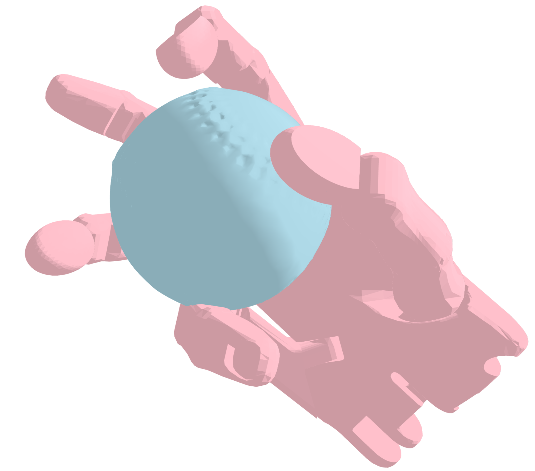}
    \hfill\end{subfigure} 
    \\

    \begin{subfigure}{0.11\linewidth}\hfill
    \includegraphics[width=1.0\linewidth]{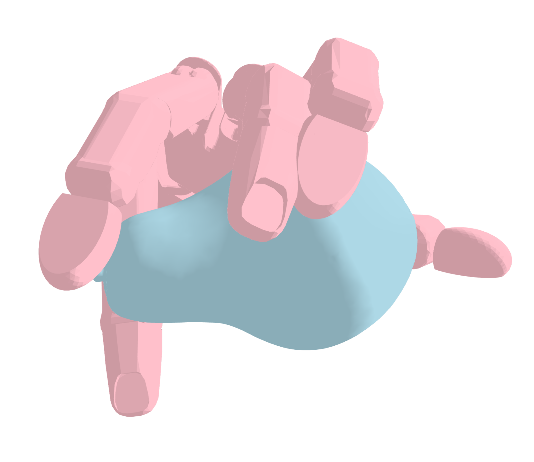}
    \hfill\end{subfigure} \hfill
    \begin{subfigure}{0.11\linewidth}\hfill
    \includegraphics[width=1.0\linewidth]{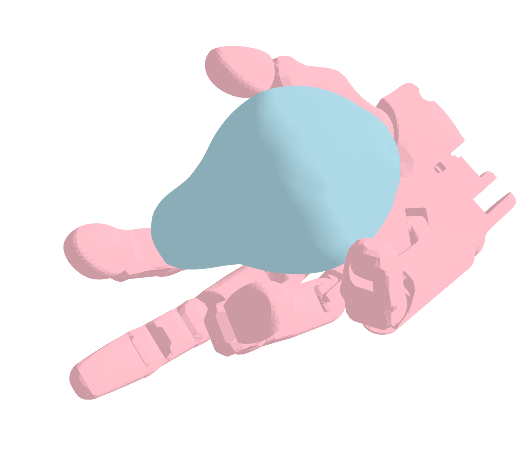}
    \hfill\end{subfigure} \hfill
    \begin{subfigure}{0.11\linewidth}\hfill
    \includegraphics[width=1.0\linewidth]{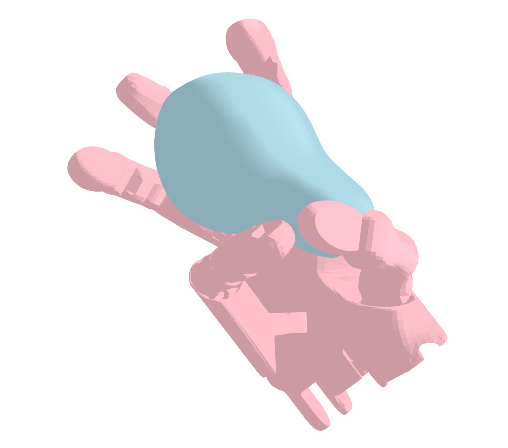}
    \hfill\end{subfigure} \hfill
    \begin{subfigure}{0.11\linewidth}\hfill
    \includegraphics[width=1.0\linewidth]{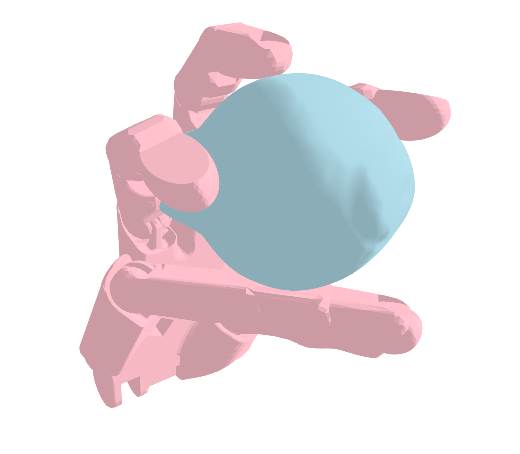}
    \hfill\end{subfigure} \hfill
    \begin{subfigure}{0.11\linewidth}\hfill
    \includegraphics[width=1.0\linewidth]{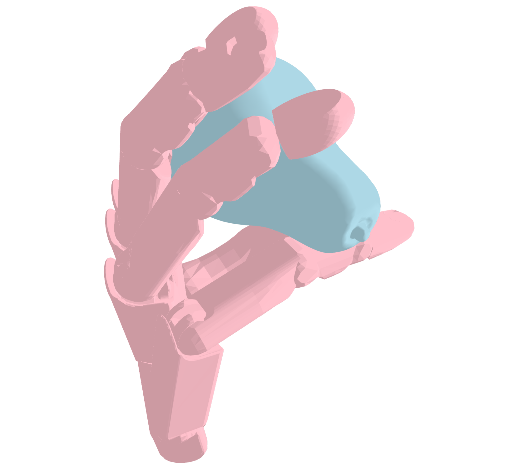}
    \hfill\end{subfigure} \hfill
    \begin{subfigure}{0.11\linewidth}\hfill
    \includegraphics[width=1.0\linewidth]{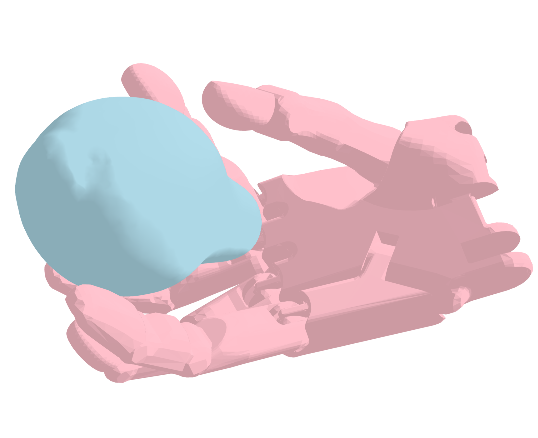}
    \hfill\end{subfigure} \hfill
    \begin{subfigure}{0.11\linewidth}\hfill
    \includegraphics[width=1.0\linewidth]{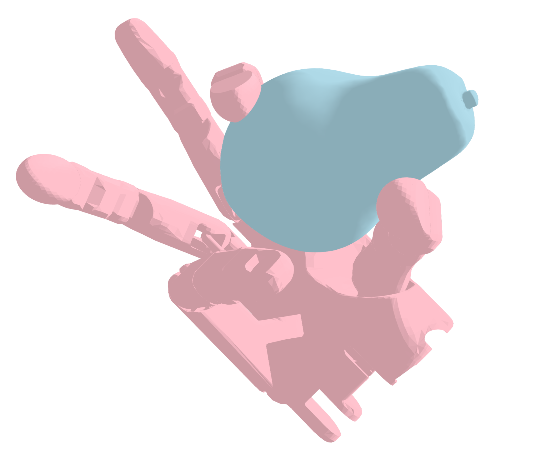}
    \hfill\end{subfigure} \hfill
    \begin{subfigure}{0.11\linewidth}\hfill
    \includegraphics[width=1.0\linewidth]{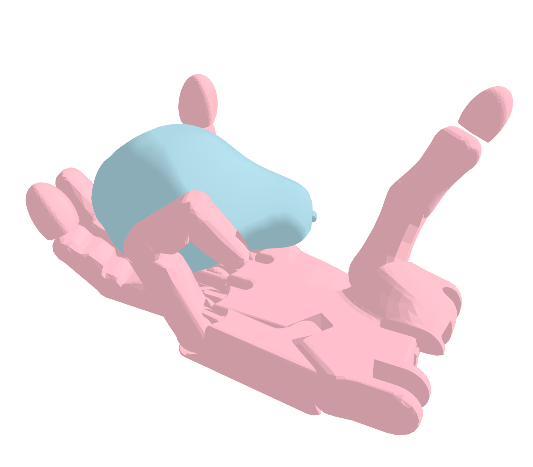}
    \hfill\end{subfigure} 
    \\
    
    \begin{subfigure}{0.11\linewidth}\hfill
    \includegraphics[width=1.0\linewidth]{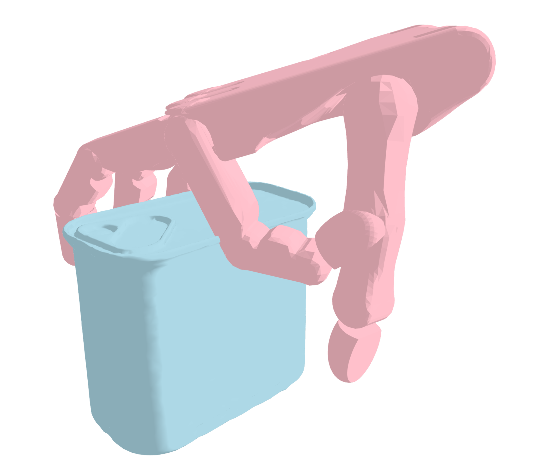}
    \hfill\end{subfigure} \hfill
    \begin{subfigure}{0.11\linewidth}\hfill
    \includegraphics[width=1.0\linewidth]{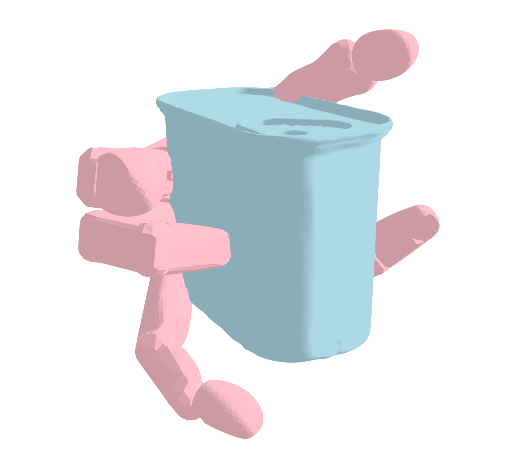}
    \hfill\end{subfigure} \hfill
    \begin{subfigure}{0.11\linewidth}\hfill
    \includegraphics[width=1.0\linewidth]{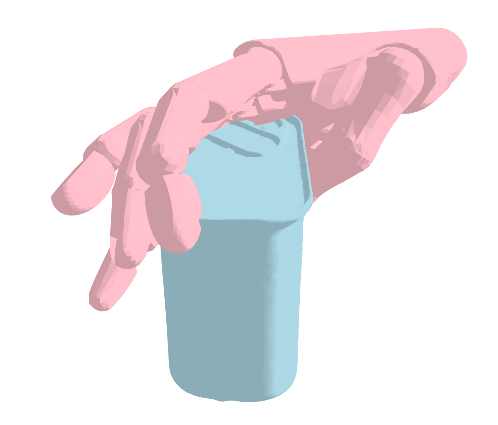}
    \hfill\end{subfigure} \hfill
    \begin{subfigure}{0.11\linewidth}\hfill
    \includegraphics[width=1.0\linewidth]{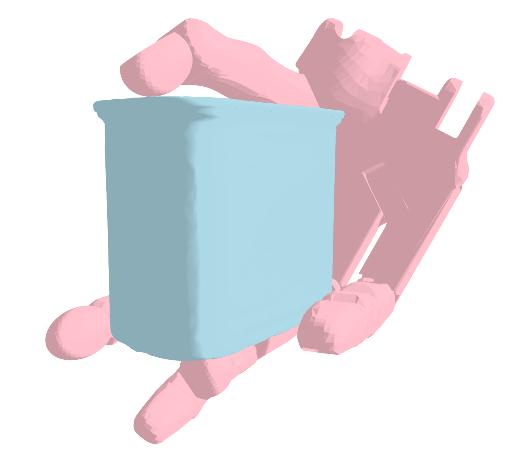}
    \hfill\end{subfigure} \hfill
    \begin{subfigure}{0.11\linewidth}\hfill
    \includegraphics[width=1.0\linewidth]{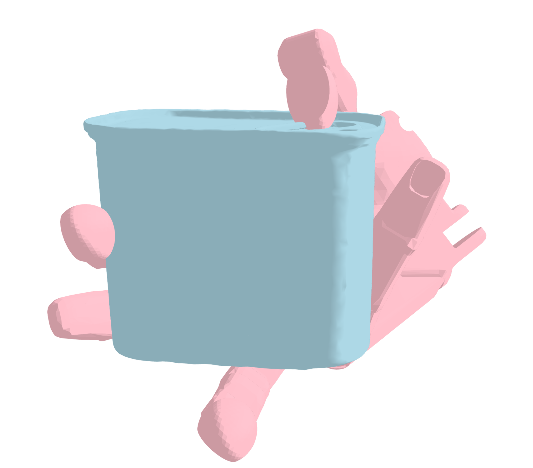}
    \hfill\end{subfigure} \hfill
    \begin{subfigure}{0.11\linewidth}\hfill
    \includegraphics[width=1.0\linewidth]{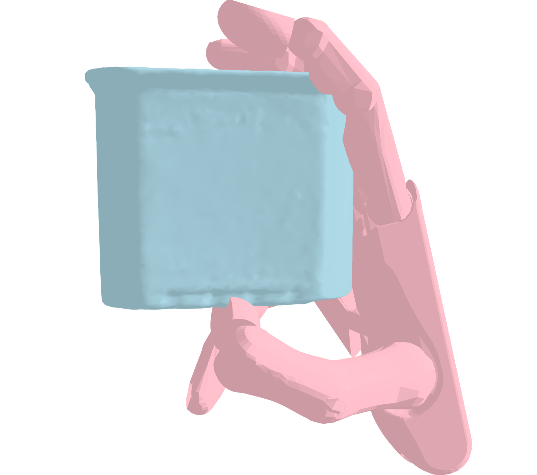}
    \hfill\end{subfigure} \hfill
    \begin{subfigure}{0.11\linewidth}\hfill
    \includegraphics[width=1.0\linewidth]{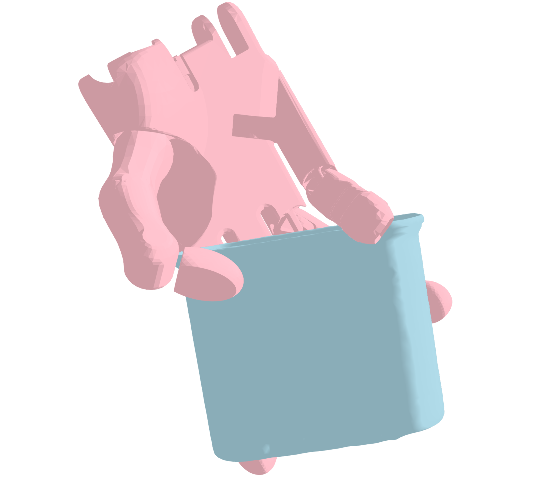}
    \hfill\end{subfigure} \hfill
    \begin{subfigure}{0.11\linewidth}\hfill
    \includegraphics[width=1.0\linewidth]{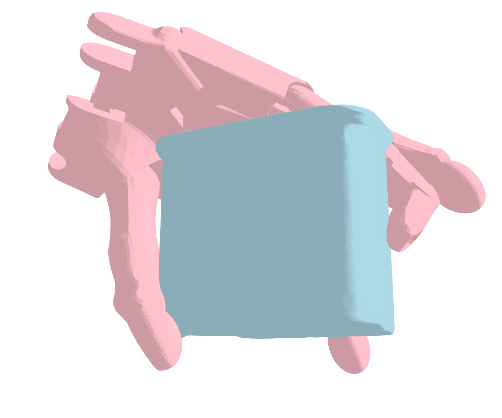}
    \hfill\end{subfigure} 
    \\

    \begin{subfigure}{0.11\linewidth}\hfill
    \includegraphics[width=1.0\linewidth]{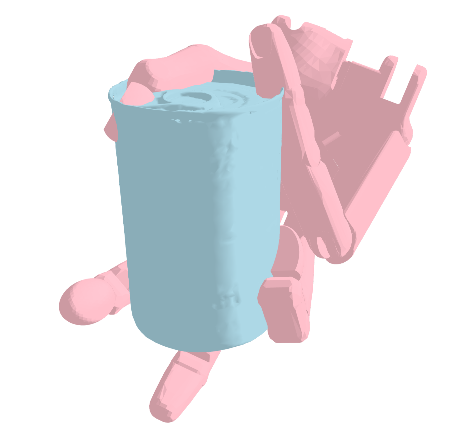}
    \hfill\end{subfigure} \hfill
    \begin{subfigure}{0.11\linewidth}\hfill
    \includegraphics[width=1.0\linewidth]{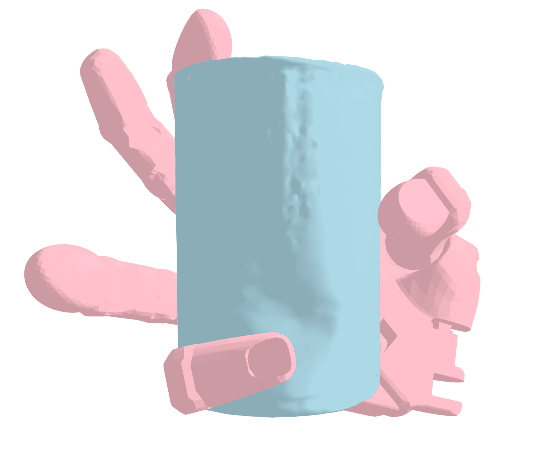}
    \hfill\end{subfigure} \hfill
    \begin{subfigure}{0.11\linewidth}\hfill
    \includegraphics[width=1.0\linewidth]{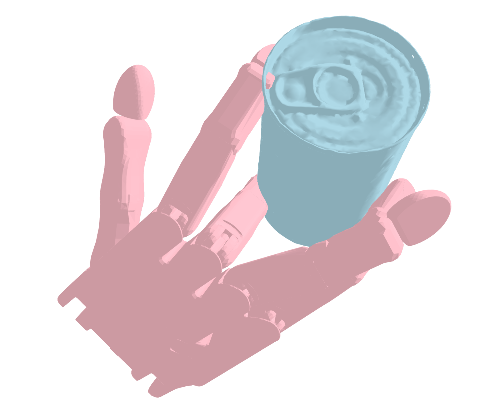}
    \hfill\end{subfigure} \hfill
    \begin{subfigure}{0.11\linewidth}\hfill
    \includegraphics[width=1.0\linewidth]{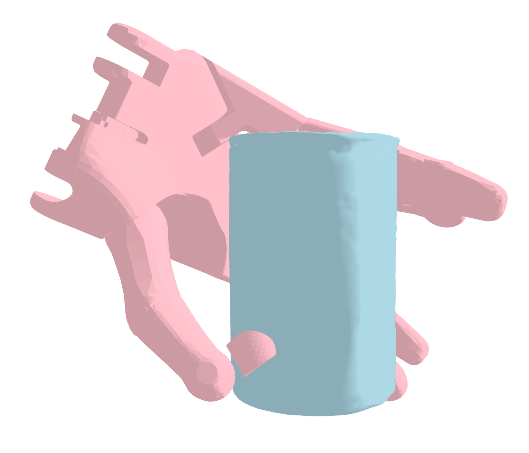}
    \hfill\end{subfigure} \hfill
    \begin{subfigure}{0.11\linewidth}\hfill
    \includegraphics[width=1.0\linewidth]{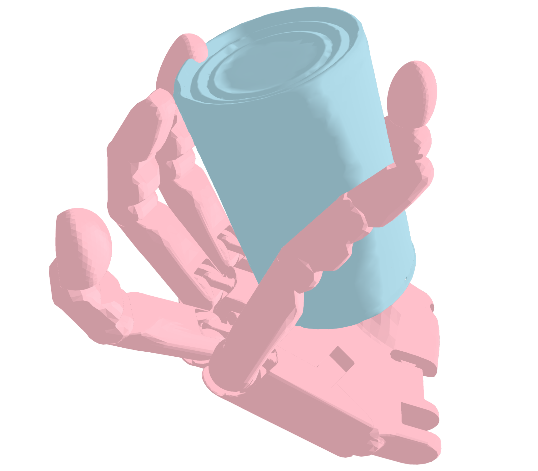}
    \hfill\end{subfigure} \hfill
    \begin{subfigure}{0.11\linewidth}\hfill
    \includegraphics[width=1.0\linewidth]{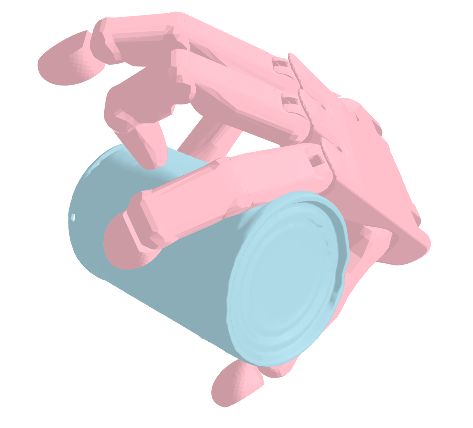}
    \hfill\end{subfigure} \hfill
    \begin{subfigure}{0.11\linewidth}\hfill
    \includegraphics[width=1.0\linewidth]{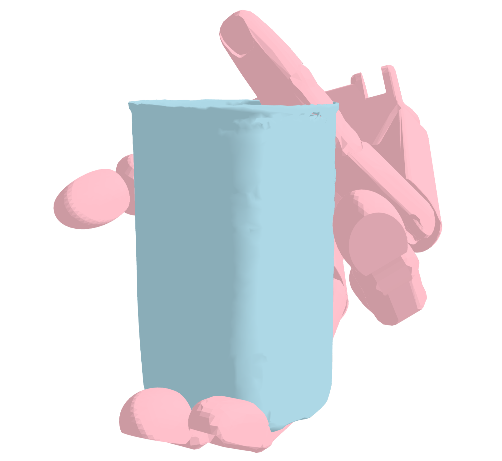}
    \hfill\end{subfigure} \hfill
    \begin{subfigure}{0.11\linewidth}\hfill
    \includegraphics[width=1.0\linewidth]{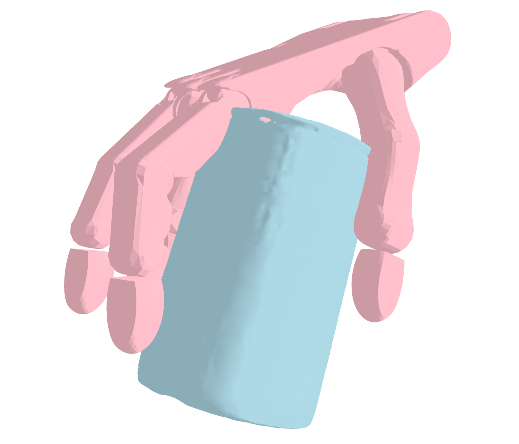}
    \hfill\end{subfigure} 
    \\
    \caption{\textbf{More qualitative results of dexterous grasp pose generation for 3D object.}}
    \label{fig:supp:grasp_gen_qual_supp}
\end{figure*}

We show more qualitative results in \cref{fig:supp:grasp_gen_qual_supp}. Note that the objects are unseen during training time.

\paragraph{Motion Planning for Robot Arm}

We render the planning results into animations for visualization. Please refer to the supplemental demo video for the qualitative results.

\end{document}